\documentclass[10pt,twocolumn,letterpaper]{article}

\usepackage[pagenumbers]{cvpr} %

\usepackage{multirow}
\usepackage{enumitem}
\usepackage{microtype}

\setlist{nosep,leftmargin=1.4em}
\renewcommand{\paragraph}[1]{\vspace{0.35em}\noindent\textbf{#1}}
\newif\ifdraftnotes\draftnotesfalse
\usepackage{CJKutf8}

\definecolor{chgcolor}{rgb}{0,0,0}
\newcommand{\chg}[1]{{\color{chgcolor}#1}}
\definecolor{chgcolorB}{rgb}{0,0,0}
\newcommand{\chgb}[1]{{\color{chgcolorB}#1}}
\usepackage{colortbl}

\definecolor{delfg}{HTML}{B23B2E}
\definecolor{delbg}{HTML}{FBECEB}
\definecolor{addfg}{HTML}{2E7D4F}
\definecolor{addbg}{HTML}{E8F4EC}
\definecolor{boxgray}{HTML}{6B6B6B}
\definecolor{silblue}{HTML}{3D8FD6}
\definecolor{silorange}{HTML}{E65A14}
\newcommand{\rl}[2]{\rotatebox{90}{\makebox[#1]{\footnotesize #2}}}

\definecolor{cvprblue}{rgb}{0.21,0.49,0.74}
\usepackage[pagebackref,breaklinks,colorlinks,allcolors=cvprblue]{hyperref}
\def\confName{CVPR}
\def\confYear{2027}

\title{\chg{RESET: One Demonstration, One Verified World,\\ Self-Improved Skills Across Embodiments}}

\author{Zheng Sun$^{1*}$ \quad Zhuo Li$^{1*}$ \quad Zhihao Li$^{1*}$ \quad Heng Yi Sim$^{1}$ \\
Ge Wang$^{2}$ \quad Ziyue Zhu$^{3}$ \quad Mingkai Jia$^{4}$ \quad Fei Chen$^{1}$\\[0.6em]
{\normalsize Project page: \href{https://zainzh.github.io/RESET/}{\texttt{zainzh.github.io/RESET}}}
}

\begin{document}

\twocolumn[{%
\renewcommand\twocolumn[1][]{#1}%
\maketitle
\vspace{-1em}
\begin{center}

    \centering
    \captionsetup{type=figure}
    \includegraphics[width=0.95\textwidth]{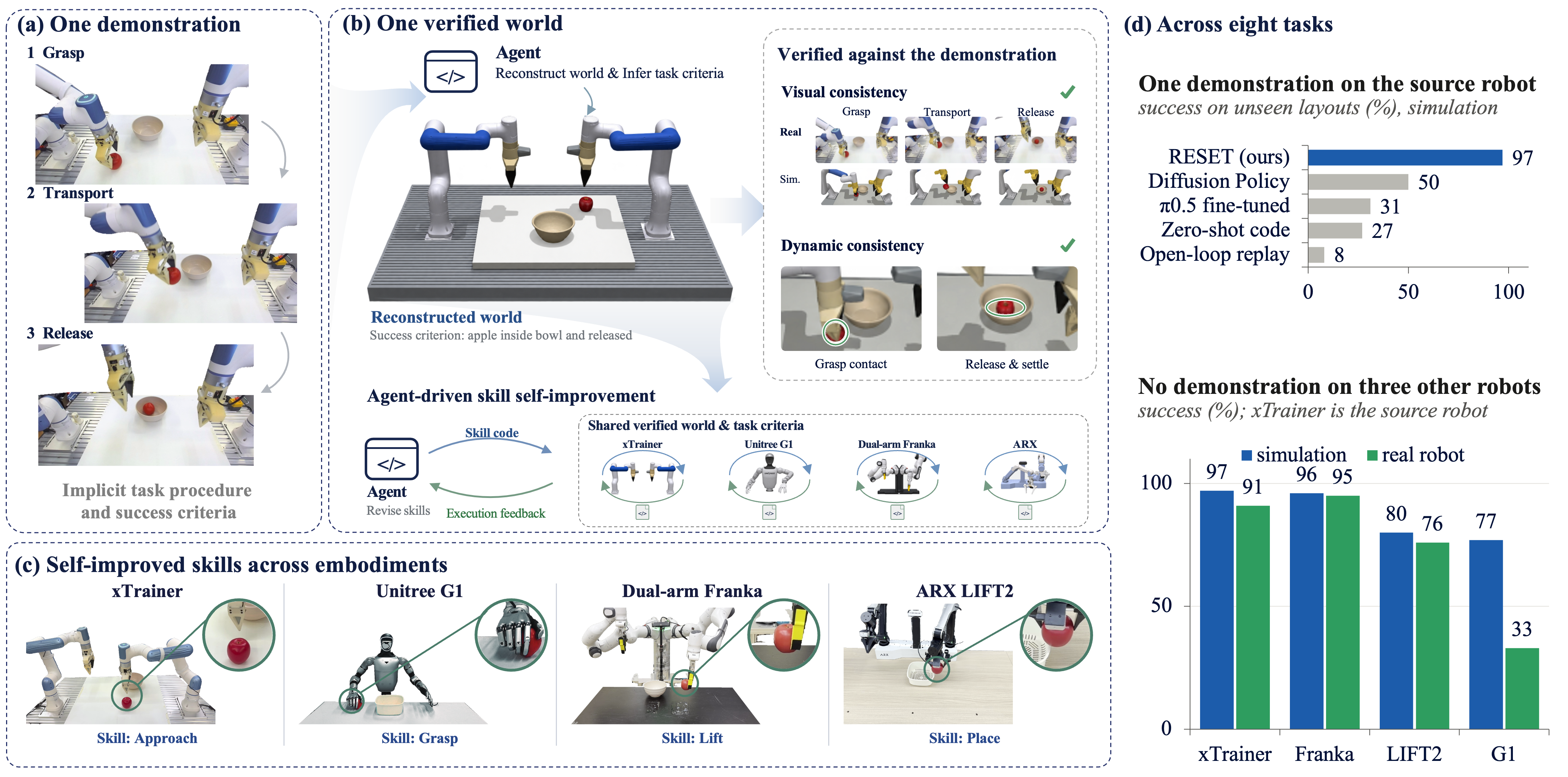}
    \captionof{figure}{\textbf{Overview of RESET.} (a) One real demonstration implicitly specifies the task procedure and terminal success condition. (b) One verified world: RESET reconstructs a contact-rich world and read-only criterion (Rebuild), \chgb{checks it against the demonstration} (Examine), and refines white-box control code from execution feedback (Self-improve). (c) Cross-embodiment transfer: the shared verified world and criterion enable heterogeneous robots to acquire adapted skills without new demonstrations (Transfer) and run them on real robots (Execute). \chg{(d) Results across the eight tasks: the source robot against baselines, and all four robots in simulation and \chgb{on real robots}.}}
    \label{fig:teaser}
\end{center}%
}]

{\renewcommand{\thefootnote}{}\footnotetext{$^{*}$Equal contribution. \quad
$^{1}$The Chinese University of Hong Kong. \quad
$^{2}$The Chinese University of Hong Kong, Shenzhen. \quad
$^{3}$Nankai University. \quad
$^{4}$The Hong Kong University of Science and Technology.}\addtocounter{footnote}{-1}}
\begin{abstract}
Large language model (LLM) coding agents can iteratively self-improve robot manipulation skills in simulation benchmarks, yet real-world tasks provide neither ground-truth success scorers nor resettable practice environments. Our key observation is that a single real demonstration simultaneously specifies scene geometry, defines the terminal success criterion, and serves as a physical reference trajectory for verifying world fidelity---all decoupled from the source robot's kinematics. We present \textbf{RESET}: we reconstruct a contact-rich world and derive a frozen read-only criterion from one demonstration (\textbf{R}ebuild); \chgb{check that world against the demonstration by open-loop replay and silhouette consistency} (\textbf{E}xamine); evolve modular white-box task graphs and skill code from per-node failure evidence (\textbf{S}elf-improve); deploy frozen code on real robots with zero runtime LLM calls (\textbf{E}xecute); and share the verified world and criterion with unseen robots without new demonstrations (\textbf{T}ransfer). \chg{On 8 manipulation tasks, each rebuilt from one real demonstration, RESET achieves 97\% average success on unseen layouts in simulation, whereas zero-shot code reaches 27\%, per-task fine-tuned $\pi_{0.5}$ policies 31\%, \chgb{and Diffusion Policy 50\%}; \chgb{injecting reconstruction defects shows that practice tolerates many reconstruction errors, that geometric errors can still stall or mislead it, and that the check flags most of them while physical randomization covers the rest.} Without any target demonstrations, 3 heterogeneous robots acquire the same task skills in the shared world: a robot with similar arms copies 95\% of its skill functions unchanged from the source robot, while robots with different arms and hands write their own. In real-robot experiments, the resulting programs run on 4 robots and succeed in 80\% of 280 trials.}
\end{abstract}

\section{Introduction}
\label{sec:intro}

LLM-driven coding agents are emerging as a compelling paradigm for acquiring robot manipulation skills. Whereas hand-crafted or one-shot synthesized control programs~\cite{liang2023code} struggle to scale across contact-rich tasks and diverse robots, closed-loop recursive self-improvement (RSI) lets an agent inspect failure evidence, revise control code, and retain verified improvements, steadily advancing performance on simulation benchmarks such as RoboDojo~\cite{zhang2026playful,wang2026reconstruct,li2026embodiedevo,tian2026physevo,hu2026embodied,chen2026robodojo}. Yet this paradigm relies on a strong hidden premise: human engineers have already built the simulation scene, reset mechanism, and ground-truth success scorer. Physical tasks supply none of these. Recent attempts to run self-improvement directly on real robots~\cite{xiao2026enpire} rely on manual or heuristic self-checks and custom reset fixtures, requiring hours to tens of hours of physical trial-and-error per task without generalizing across embodiments. Scaling code self-improvement to the real world thus hinges on a foundational question: \emph{without a hand-engineered benchmark, where do a reliable success criterion and a trustworthy practice world come from?}

Reconstructing real scenes into simulation, practicing inside them, and transferring back to the real world (Real-to-Sim-to-Real) offers a natural route~\cite{torne2024reconciling,han2026re,dan2025x}, but existing pipelines \chg{fall short in three ways}. First, \emph{reconstructed worlds lack pre-training verification}: prior methods proceed straight from scene reconstruction to policy learning, leaving world fidelity to be discovered only post hoc via real-world failure~\cite{nishizawa2026yubi}. Second, \emph{success criteria rely on manual reward design and invite reward hacking}: under multi-round code search, agents readily exploit loose criteria with unintended behaviors~\cite{torne2024reconciling}. Third, \emph{demonstrations remain entangled with source kinematics}: imitating recorded trajectories binds task knowledge to one robot morphology, invalidating the demonstration whenever degrees of freedom (DoF) or end-effectors change~\cite{torne2024reconciling,dan2025x}.

Our core insight is that \textbf{a single successful real-world demonstration simultaneously provides \chg{a blueprint of the world, a success criterion, and a physical reference for verifying that world---all independent of the source robot's kinematics}.} The initial RGB-D frame captures scene geometry, while the terminal state specifies exact geometric completion relations. Crucially, because the demonstration is a known-successful physical trajectory, an accurate world should reproduce object grasps and motion under open-loop contact replay of the recorded joint commands and align rendered object silhouettes with observed masks. Moreover, by distilling the demonstration into an embodiment-agnostic specification (\emph{world + frozen criterion + local contact priors}) rather than imitating its trajectory, an agent can \chg{write modular control code on its own}. This both resolves the scalability bottleneck of manual code-skill authoring and exploits the white-box nature of code skills---explicit geometric guards, stage-level verification, and deterministic execution---for long-horizon stability.

Guided by this insight, we propose \textbf{RESET} (\textbf{R}ebuild, \textbf{E}xamine, \textbf{S}elf-improve, \textbf{E}xecute, \textbf{T}ransfer; \cref{fig:teaser}). \chg{The name reflects what the method restores to a real task: a world that can be reset and practiced in at will.} Rebuild reconstructs an interactive MuJoCo scene from one RGB-D demonstration and extracts a read-only terminal scorer. \chgb{Examine checks the rebuilt world against the demonstration by dynamic replay and silhouette alignment before practice, leaving unidentifiable mass and friction to randomized practice.} Self-improve evolves an object-centric task graph and white-box skill library from per-node failure evidence under diagnostic/hold-out gating. Execute runs frozen code online with zero LLM calls, low cost, and node-level debuggability. Transfer hands the verified world and criterion unchanged to target robots that adapt only a thin hardware layer.

\chg{We evaluate RESET across 4 diverse embodiments and 8 real manipulation tasks. From one demonstration on the source robot, RESET achieves 97\% success on unseen layouts in simulation, where zero-shot code and per-task fine-tuned $\pi_{0.5}$ policies stay near 30\% \chgb{and Diffusion Policy reaches 50\%}. \chgb{Controlled defect injection shows that practice tolerates many reconstruction errors. A geometric error, however, can make a task unsolvable or hide a failure during practice, and it changes what a simulated success rate means; Examine flags most such errors, and physical randomization covers mass and friction.} Without any target demonstration, Dual-arm Franka, ARX LIFT2, and Unitree G1 acquire their own skills in the shared world. The final programs succeed in 80\% of 280 real-robot trials on 4 physical platforms.}

Our main contributions are threefold:
\begin{itemize}
\item \chg{\textbf{A framework from one demonstration to skills across embodiments.} RESET turns a single real demonstration into frozen, deployable code skills, both for the demonstrated robot and for robots that never received one.}
\item \chg{\textbf{The demonstration replaces the benchmark.} One demonstration supplies what self-improvement otherwise takes from a hand-built benchmark: a practice world, a success criterion, and a reference for checking that world. Agent self-improvement can therefore run on real tasks without hand-built scenes, rewards, or resets.}
\item \chg{\textbf{Evidence on eight real tasks and four embodiments.} RESET reaches 97\% success on unseen layouts in simulation where zero-shot code reaches 27\%; three target robots learn the tasks with no new demonstration\chgb{, by practicing in the shared world rather than by copying the source robot's code; practice tolerates many reconstruction errors;} and frozen skills run on 4 physical platforms (80\% success over 280 real-robot trials).}
\end{itemize}
\begin{figure*}[t]
\centering
\includegraphics[width=0.92\textwidth]{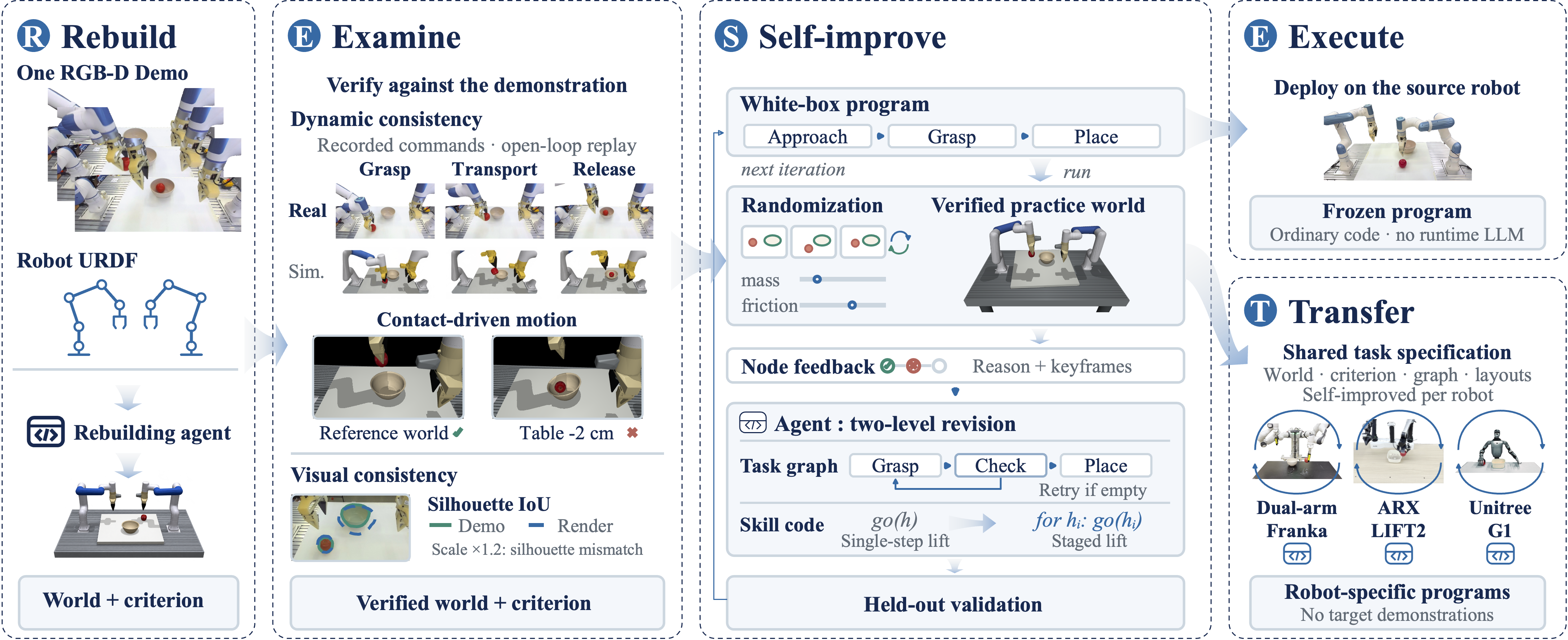}
\caption{\chg{\textbf{The five steps of RESET, illustrated on the \emph{Apple} task.} Rebuild turns one RGB-D demonstration and the robot's URDF into an interactive world and a success criterion. Examine checks that world against the demonstration by open-loop replay and by rendered silhouettes. Self-improve revises a white-box program from node-level failures and keeps a revision only if it does better on held-out layouts. Execute runs the final program on the real robot with no LLM call. Transfer gives other robots the same world and criterion to practice in, without a demonstration of their own.}}
\label{fig:method}
\end{figure*}

\section{Related Work}
\label{sec:related}
\paragraph{Code-skill self-improvement in simulation benchmarks.}
Code as Policies~\cite{liang2023code} and Voyager~\cite{wang2023voyager} pioneered synthesizing control programs and skill libraries with LLMs. Recent closed-loop RSI frameworks---including RATs~\cite{zhang2026playful}, \chgb{the concurrent} RPG~\cite{wang2026reconstruct}, EmbodiedEvo~\cite{li2026embodiedevo}, PhysEvo~\cite{tian2026physevo}, Embodied Turing Machines~\cite{hu2026embodied}, PyRUA-Lean~\cite{si2026fewer}, \chg{and ReSkill~\cite{sun2026reskill}}---iteratively refine task graphs and skill code from execution feedback on benchmarks such as RoboDojo~\cite{chen2026robodojo}. However, all assume human-engineered simulation scenes, resets, and ground-truth scorers. RESET removes this benchmark dependency by running self-improvement inside a world reconstructed from and verified against a single real demonstration, while \chgb{also enabling transfer across robot embodiments}.

\paragraph{Self-improvement on physical robots.}
To bypass simulation benchmarks, ENPIRE~\cite{xiao2026enpire} explores and revises policies directly on real robots. While demonstrating real-world RSI feasibility, physical exploration demands hours of robot wear, custom reset mechanisms, and robot-specific verification logic. RESET confines trial-and-error to a verified simulation world, reserving the real robot for executing the final code and reusing the world across embodiments.

\paragraph{Real-to-Sim-to-Real and scene verification.}
RialTo~\cite{torne2024reconciling}, Re3Sim~\cite{han2026re}, and RoboSplat~\cite{yang2025novel} reconstruct scenes to train neural policies or synthesize data; X-Sim~\cite{dan2025x} derives cross-embodiment dense rewards from simulated object motion; EmbodiRSI~\cite{lang2026embodirsi} and D-World~\cite{wen2026d} update digital twins from single images or online rollouts. Yet prior pipelines treat reconstruction as an open-loop preprocessing step \chgb{without checking the rebuilt world before training}~\cite{nishizawa2026yubi}. RESET uses the demonstration itself as a dynamic and visual verification standard, characterizes how reconstruction errors affect code policies, and outputs inspectable white-box programs.

\paragraph{Learning from a single demonstration.}
Demo2Code~\cite{wang2023demo2code} summarizes demonstrations into code specifications, RoboSplat~\cite{yang2025novel} augments single demonstrations via novel-view synthesis, and Human2Sim2Robot~\cite{lum2025crossing} bridges human-robot morphology gaps via RL from one video. Unlike methods that imitate demonstrated trajectories, RESET uses the demonstration solely to build and verify the environment and terminal criterion; runtime controllers never replay the demonstration, unlocking zero-demonstration transfer.

\section{Method}
\label{sec:method}

\subsection{Motivation}
\label{sec:motivation}

\chg{Three questions shape the design of RESET. Each starts from an observation about what happens without the corresponding component (\cref{sec:exp}).}

\chg{\paragraph{Why does code need practice, and where can it practice?}
Code written once is not enough. Given the same interfaces and contact hints that RESET uses, an agent's first program succeeds on 27\% of unseen simulated layouts on average, varies widely between samples (70\%, 0\%, and 36\% on one task), and never succeeds on the four hardest tasks (\cref{tab:main}). Such a program improves only by being run, failing, and being revised, which takes an environment that resets at will and a score that needs no human. A real task offers neither, and practicing on a real robot costs hours of supervised trials per task~\cite{xiao2026enpire}. One successful demonstration, however, already contains both: its first frame shows the scene and its final state defines success, which motivates \emph{Rebuild}.}

\chgb{\paragraph{Why check the rebuilt world against the demonstration?}
No reconstruction is perfect, and a success rate measured in a wrong world says little about the real one. When object scale is off by 10\% or the table is $2\,\mathrm{cm}$ too low, a skill that is perfect in the correct world falls to 0\% on several tasks (\cref{tab:sensitivity}); and under a loosely worded criterion, a program that knocks the carton flat is scored 100\%. The demonstration gives a check for both. It is a physical execution known to succeed, so a usable world must reproduce it; and its final state, not a text description, defines success. What one demonstration cannot determine, mass and friction, we do not claim to verify and instead randomize during practice. This motivates \emph{Examine}.}

\chg{\paragraph{Why code, and why share the task rather than the motion?}
The demonstrated motion itself transfers poorly: replayed open-loop it succeeds on 8\% of new layouts, and a vision-language-action policy fine-tuned on 100 trajectories per task reaches 31\% \chgb{and a Diffusion Policy trained on the same trajectories 50\%}. A program differs in three ways. It exposes which step failed, which gives an agent something specific to revise (\cref{sec:exp-ablation}). It runs on the real robot with no model in the loop. And it keeps what the task requires, the world and the criterion, apart from how one robot meets that requirement, so a robot with different arms or hands can practice in the same world (\cref{sec:exp-transfer}). This motivates \emph{Self-improve}, \emph{Execute}, and \emph{Transfer}.}

\subsection{Problem Formulation and Overview}
\label{sec:formulation}

\chg{Let $\mathcal{D} = \{(I_t, D_t, q_t, g_t)\}_{t=0}^{T}$ be one successful real demonstration on a source robot $\mathcal{E}_s$, with RGB image $I_t$, depth map $D_t$, joint angles $q_t$, and gripper command $g_t$. The scene has observable geometry $\theta_{\mathrm{obs}}$ (object scale, table height, camera pose), which $\mathcal{D}$ constrains, and latent physics $\theta_{\mathrm{lat}}$ (mass, friction), which one successful execution cannot identify. Given target robots $\{\mathcal{E}_k\}$ with no demonstration, we synthesize control programs $P_s$ and $\{P_k\}$ that complete the task from unseen object layouts, without hand-written rewards. RESET does so in five stages (\cref{fig:method}): Rebuild, Examine, Self-improve, Execute, and Transfer.}

\begin{table*}[t]
\centering\small

\caption{\textbf{Main blind-test success rates (\%) on source embodiment xTrainer across 50 unseen layouts per task (TEST 0--49), \chg{in simulation}.} Zero-shot Code reports the mean and best of 3 independent samples. $\pi_{0.5}$ \chgb{and Diffusion Policy each train} one dedicated model per task on \chgb{the same} 100 successful RESET rollouts \chgb{($\pi_{0.5}$ is fine-tuned; Diffusion Policy is trained from scratch)}.}
\label{tab:main}
\setlength{\tabcolsep}{4.0pt}
\begin{tabular}{@{}lccccccccc@{}}
\toprule
Method & Apple & Cubes & Stack Cups & Build Tower & Mug-Box & Open Drawer & Sweep & Lift-Box & Mean \\\midrule
Open-loop Demo Replay & 4\% & 0\% & 12\% & 0\% & 38\% & 0\% & 0\% & 10\% & 8\% \\
Zero-shot Code (3-run mean) & 85\% & 35\% & 55\% & 39\% & 0\% & 0\% & 0\% & 0\% & 27\% \\
Zero-shot Code (3-run best) & \textbf{100\%} & 70\% & \textbf{96\%} & 76\% & 0\% & 0\% & 0\% & 0\% & 43\% \\
$\pi_{0.5}$ fine-tuned (per-task)~\cite{intelligence2025pi_} & 4\% & 0\% & 76\% & 6\% & 72\% & 4\% & 32\% & 52\% & 31\% \\
\chgb{Diffusion Policy (per-task)} & \chgb{20\%} & \chgb{0\%} & \chgb{94\%} & \chgb{6\%} & \chgb{94\%} & \chgb{26\%} & \chgb{70\%} & \chgb{88\%} & \chgb{50\%} \\
\textbf{RESET (Ours)} & \textbf{100\%} & \textbf{98\%} & \textbf{96\%} & \textbf{100\%} & \textbf{100\%} & \textbf{100\%} & \textbf{88\%} & \textbf{94\%} & \textbf{97\%} \\\bottomrule
\end{tabular}
\end{table*}

\subsection{Rebuild: World and Read-Only Criterion}
\label{sec:rebuild}

\chg{\paragraph{Contact-rich world reconstruction.}
We recover the camera pose by aligning the robot's kinematics with the images across the demonstration, and segment the task objects in the first frame $(I_0, D_0)$ with a pretrained promptable segmentation model. A multimodal coding agent iteratively models each object as a mesh, keeping the cavities and articulated parts that stacking, insertion, and drawer pulling depend on. The meshes are registered and scaled to the first-frame depth point cloud, the table plane is fitted, and the scene is exported to a physics simulator together with the robot's URDF. $\theta_{\mathrm{lat}}$ receives nominal values.}

\chg{\paragraph{Terminal criterion and read-only scorer.}
The success criterion is read from the demonstrated final state as a conjunction of geometric relations (containment, support, insertion, hanging, upright orientation) plus a stillness condition after release. \chgb{The criterion is confirmed once by a person, after the world has been rebuilt and has passed Examine.} The scorer is then read-only and hidden from the controller, so practice cannot game it.}

\subsection{Examine: Dual-Channel World Verification}
\label{sec:examine}

\chgb{Sensor noise, camera error, and scale error make every rebuilt world deviate from the real scene. Because $\mathcal{D}$ is known to succeed, Examine checks whether the rebuilt world can reproduce it, through two complementary channels. \emph{The dynamic channel} places objects at their first-frame poses and replays the recorded joint and gripper commands open-loop; objects are never attached to the gripper, so grasping and placement must arise from contact. It measures how many of the demonstrated grasps are reproduced and how far the objects deviate from their tracked motion, which constrains table height and contact support. \emph{The visual channel} renders the world at the recorded joint angles before contact and measures the silhouette IoU of each task object against the demonstration masks, which constrains camera pose and object scale. These measurements are used to compare reconstructions of the same scene and to discard those that reproduce the demonstration clearly worse; the margins are fixed on development tasks before held-out evaluation. Examine is a check that the world is usable, not a guarantee of fidelity: mass and friction leave both channels almost unchanged and are covered by randomized practice.}

\subsection{Self-Improve: Two-Level Code Evolution}
\label{sec:improve}

\chg{The verified world, randomized collision-free layouts, and the read-only scorer form a task-specific benchmark that resets automatically. In it the controller is a white-box program $P = (G, \Pi, \Lambda)$: an object-centric task graph $G$ whose nodes call skills with entry and exit checks and retry budgets, a skill library $\Pi$ (pose estimation, inverse kinematics, and grasp, place, sweep, and handover primitives), and the bindings $\Lambda$ between them. $P$ may read only what a physical robot provides: camera streams, calibration, joint encoders, and the gripper-to-object contact offsets observed in the demonstration. It never reads simulator state or the scorer, so the same program runs unchanged \chgb{on the real robot}.}

\chg{In each round, a coding agent runs $P$ on diagnostic layouts, reads the status and failure reason of every node together with keyframes, locates the earliest node that went wrong, and revises the program at two levels. Task-level revisions change the graph (e.g., a retry after an empty grasp, a handover between arms); skill-level revisions change the geometry inside a skill (e.g., where to grasp). To avoid fitting to particular failures, a revised program replaces the current one only if it succeeds more often on a separate set of hold-out layouts. When practice converges or the round budget is spent, $P$ is fixed and evaluated once on sealed layouts.}

\subsection{\chgb{Execute and Transfer}}
\label{sec:execute}\label{sec:transfer}

\chg{The final program runs on the physical robot as ordinary code, with no LLM call and no human intervention. Before each node it estimates object poses from segmentation and depth, then executes the skill. All reasoning happens offline, \chgb{and because the program is frozen after practice, the code that runs on the real robot is exactly the code evaluated in simulation;} a failure can be traced to a specific node.}

\chg{A target robot $\mathcal{E}_k$ receives everything that is independent of the robot unchanged: the verified world, the criterion, the layout distribution, and the source robot's task graph with its notes on physical pitfalls. It supplies its own URDF, camera calibration, and contact offsets for its end-effector, and runs Self-improve in the shared world to obtain $P_k$. \chgb{No trajectory or skill code of the source robot is required: a target robot reuses a source function when it suits its own arms or hands and otherwise writes its own (\cref{sec:exp-transfer}). The demonstration is thus used once, to build and check the world, and never as a motion to imitate.}}

\section{Experiments}
\label{sec:exp}

\begin{figure*}[t]
\centering
\includegraphics[width=0.9\textwidth]{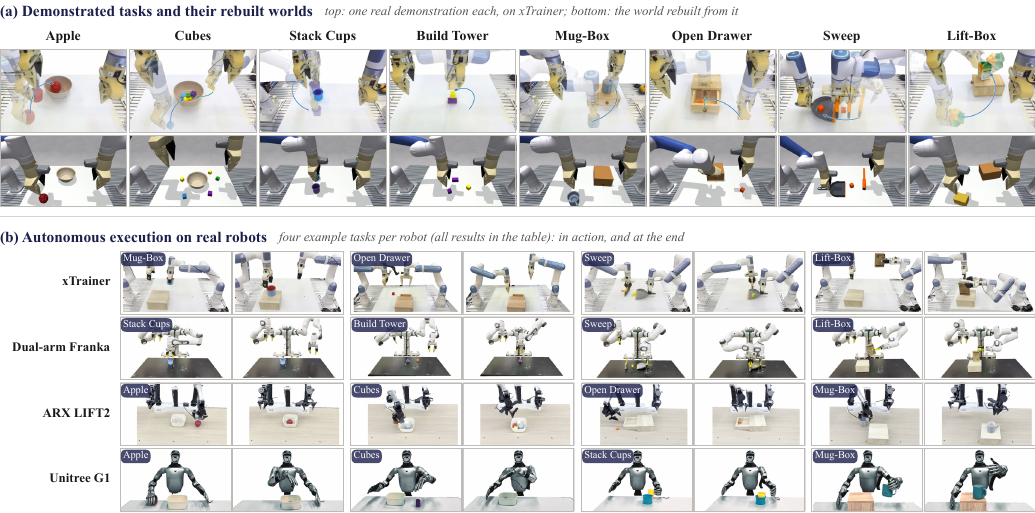}
\caption{\chg{\textbf{Demonstrated tasks, their rebuilt worlds, and autonomous real-robot execution.} (a) Top: the eight tasks, each demonstrated once on the source robot xTrainer. Bottom: the world rebuilt from each demonstration, rendered from the demonstration camera. (b) The final programs running on four robots without human intervention. For each robot we show four of the tasks it ran (all results are in \cref{tab:real}), each in action and at the end. Franka, LIFT2, and G1 received no demonstration.}}
\label{fig:real_rollouts}
\end{figure*}

We evaluate five core questions---Q1: layout generalization vs.\ baselines; \chgb{Q2: tolerance to reconstruction errors;} Q3: Self-improve component ablations; Q4: zero-demonstration cross-embodiment transfer; and Q5: physical deployment and episode-level Sim-to-Real agreement.

\subsection{Experimental Setup}
\label{sec:exp-setup}

\paragraph{Embodiments and tasks.}
We evaluate 4 platforms with different kinematics, cameras, and end-effectors (\cref{fig:teaser}c). The dual-arm 6-DoF xTrainer (parallel grippers) is the only \emph{source embodiment} and provides one demonstration per task. The three \emph{target embodiments} receive none: ARX LIFT2 (6-DoF arms on a lift), Unitree G1 (five-finger hands; lower body fixed in simulation), and a 7-DoF Dual-arm Franka. We test 8 tabletop tasks (\cref{fig:real_rollouts}): \emph{Apple}, \emph{Cubes}, \emph{Stack Cups}, \emph{Build Tower}, \emph{Mug-Box} (bimanual handover), \emph{Open Drawer} (articulated pull + place), \emph{Sweep} (tool sweeping), and \emph{Lift-Box} (upright orientation).

\chg{\paragraph{Implementation.}
Rebuild uses GPT-6 Astra (object assets, in Blender), SAM3~\chgb{\cite{carion2026sam}} (segmentation), and MuJoCo~\cite{todorov2012mujoco}; Self-improve uses Claude Code with Claude Opus 5.5.} \chgb{A person confirms the success criterion once per task; this changed the criterion only for \emph{Lift-Box}.}

\paragraph{Protocol and sandbox isolation.}
\chg{Q1--Q4 are evaluated in simulation, in the rebuilt worlds.} Layouts use disjoint seeds: Diagnostic (DEV 0--19), Hold-out Adoption (DEV 20--39), and sealed Blind Test (TEST 0--49, \chg{evaluated once, after development ends,} under the final read-only scorer). \chgb{Each layout moves every task object by up to $\pm 3$--$6\,\mathrm{cm}$ along each table axis and rotates it by up to $\pm 10^\circ$--$180^\circ$ about the vertical, around its demonstrated pose; layouts with collisions or objects out of reach are discarded.} Ablations and reruns use isolated agents that see only the initial code, the task text, and their own logs.

\subsection{Main Results on Source Embodiment (Q1)}
\label{sec:exp-main}

\paragraph{Baselines and quantitative comparison.}
\Cref{tab:main} compares RESET against Open-loop Demo Replay, Zero-shot Code (3 independent samples per task with the same interfaces and contact priors), and per-task $\pi_{0.5}$ policies~\cite{intelligence2025pi_} (fine-tuned for 6,000 steps on 100 RESET trajectories from disjoint layouts). Open-loop replay averages 8\%: the layouts cannot be solved by memorizing a trajectory. Zero-shot code averages 26.8\% with extreme variance (\emph{Cubes}: 70/0/36\%; \emph{Stack Cups}: 96/18/50\%) and scores 0\% across all 3 samples on \emph{Mug-Box}, \emph{Open Drawer}, \emph{Sweep}, and \emph{Lift-Box}. Per-task $\pi_{0.5}$ averages 30.8\%, and \chgb{a Diffusion Policy~\cite{chi2025diffusion} trained from scratch on the same trajectories 49.8\%: it approaches RESET on \emph{Stack Cups}, \emph{Mug-Box}, and \emph{Lift-Box}, but stays at or below 26\% on \emph{Apple}, \emph{Cubes}, \emph{Build Tower}, and \emph{Open Drawer}, which need precise grasps or placements. Both policies use their default settings; 100 trajectories are little data for fine-tuning a model as large as $\pi_{0.5}$, which may explain its lower score.} RESET reaches 97.0\% overall and 88\%--100\% on the four tasks where zero-shot code scores 0\%.

\paragraph{Repeatability and criterion precision.}
Independent reruns from zero-shot starts reach 100\% on \emph{Build Tower}, 98\% on \emph{Open Drawer}, and 86\% on \emph{Sweep}. On \emph{Lift-Box}, a loose text-translated criterion (``carton on box'') scores a zero-shot program 100\% for knocking the carton flat; the demonstration-derived upright criterion scores it 0\%.

\begin{figure*}[t]
\centering
\includegraphics[width=0.86\textwidth]{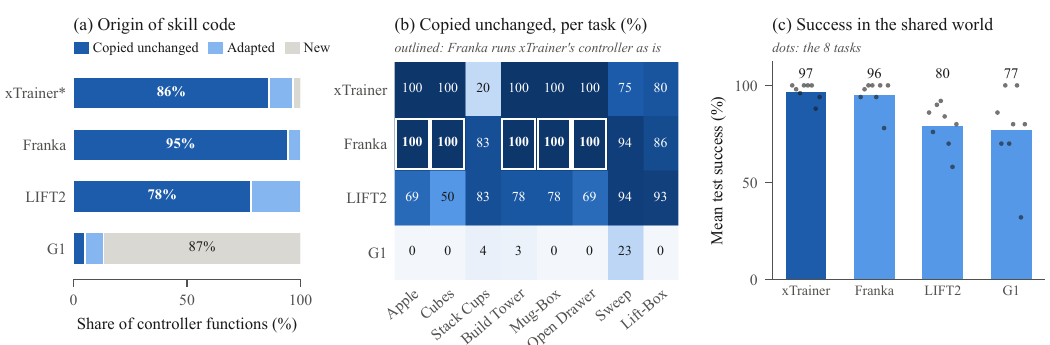}
\caption{\textbf{Cross-embodiment skill reuse (AST analysis) and task performance.} (a) Proportion of skills reused unchanged from another embodiment, adapted, or written from scratch. (b) Per-task unchanged skill reuse, highlighting 5 tasks where Dual-arm Franka passes directly with xTrainer's controller. (c) Blind-test success distributions across embodiments. \chgb{$^*$For the source robot, the share of its functions that also appear unchanged in another robot's code.}}
\label{fig:reuse}
\end{figure*}

\subsection{\chgb{How Much Reconstruction Error Can RESET Tolerate? (Q2)}}
\label{sec:exp-recon}

\chgb{No reconstruction is perfect, so we inject one defect at a time into the rebuilt worlds: geometric errors (object scale $\times 0.8$--$\times 1.2$, a table $2\,\mathrm{cm}$ too low, a shifted camera) and latent physical errors (friction $\times 0.3$, mass $\times 3$). We ask whether practice in such a world still yields a skill that works in the correct one, whether a success rate measured there can be trusted, and whether Examine flags the world.}

\begin{table}[t]
\centering\footnotesize
\caption{\chg{\textbf{Wrong worlds: how often they break a skill, and how often Examine \chgb{flags} them.} A skill learned in the correct world is run in worlds with one injected defect; entries count task--defect cases over six tasks. Breaks: success of at most 50\%. Last column: breaking cases that Examine flags.}}
\label{tab:sensitivity}
\setlength{\tabcolsep}{4pt}
\begin{tabular}{@{}lccc@{}}
\toprule
Injected defect & Skill breaks & \chgb{Examine flags} & \chgb{Breaking, flagged} \\\midrule
\chgb{None (correct world)} & \chgb{0/6} & \chgb{0/6} & \chgb{--} \\
Scale $\times 0.8$, $\times 0.9$ & 1/12 & 11/12 & 1/1 \\
Scale $\times 1.1$--$\times 1.2$ & 12/18 & 11/18 & 9/12 \\
Table $-2\,\mathrm{cm}$ & 3/6 & 6/6 & 3/3 \\
Camera shift & 0/6 & 6/6 & -- \\\midrule
Friction $\times 0.3$ & 0/6 & 0/6 & -- \\
Mass $\times 3$ & 2/6 & 0/6 & 0/2 \\
\bottomrule
\end{tabular}
\end{table}

\chgb{\paragraph{Finding 1: Practice tolerates most reconstruction errors; a wrong world hurts when it hides a failure.}
We run the full Self-improve inside defective worlds and test the result in the correct world (\cref{tab:practice}). In most cases the outcome is the same as after practice in the correct world: \emph{Mug-Box} reaches 100\% from all three defective worlds, \emph{Build Tower} 98\%--100\%, \emph{Apple} 94\% with the table $2\,\mathrm{cm}$ too low, and \emph{Stack Cups} 82\%--100\% in four of six runs with enlarged objects. The revisions an agent makes re-anchor the skill on what it perceives (e.g., a grasp height relative to the perceived rim instead of the demonstrated absolute height), and such revisions carry over. Two cases fail. A geometric error can make the task unsolvable in the practice world, so the program stays where it started (\emph{Apple} with enlarged objects). More subtly, a geometric error can hide a failure: with the table $2\,\mathrm{cm}$ too low, the zero-shot \emph{Stack Cups} program happens to release at the right height and succeeds in 98\% of practice episodes, so practice finds nothing to fix. In three independent runs the resulting program reaches only 18\%--20\% in the correct world, against 96\% after practice there.}

\begin{table}[t]
\centering\footnotesize
\caption{\chgb{\textbf{Practicing in a defective world, then testing in the correct one.} Success in the correct world after Self-improve was run in a world with the given defect. Start: the zero-shot program. Several values: independent runs. \emph{Stack Cups} at $\times 1.1$ and $\times 1.2$: criterion scaled with the world.}}
\label{tab:practice}
\setlength{\tabcolsep}{3.2pt}
\begin{tabular}{@{}lccccc@{}}
\toprule
 & & \multicolumn{4}{c}{Practice world} \\
\cmidrule(lr){3-6}
Task & Start & Correct & Scale $\times 1.1$ & Scale $\times 1.2$ & Table $-2\,\mathrm{cm}$ \\\midrule
Apple & 78\% & 100\% & 78\% & 78\% & 94\% \\
Build Tower & 42\% & 100\% & 100\% & 98\% & 100\% \\
Mug-Box & 0\% & 100\% & 100\% & 100\% & 100\% \\
Stack Cups & 18\% & 96\% & 18/98/98\% & \chgb{8/82/100\%} & 18/20/18\% \\
\bottomrule
\end{tabular}
\end{table}

\chgb{\paragraph{Finding 2: A success rate measured in a wrong world cannot be trusted.}
The opposite direction is not tolerant. A skill that is perfect in the correct world breaks in 12 of 18 worlds with objects enlarged by 10\%--20\% and in 3 of 6 with the table $2\,\mathrm{cm}$ lower, whereas camera shifts and friction changes are absorbed by online localization. A geometric error therefore changes what a simulated success rate means, even when it does not prevent learning, which is why we check the world against the demonstration and verify that it predicts real-robot outcomes (\cref{sec:exp-real}).}

\chgb{\paragraph{Finding 3: Examine flags most geometric errors; randomization covers physics.}
Examine compares versions of the same scene and flags those that reproduce the demonstration clearly worse than the best one. It flags \chgb{no correct world and} 13 of the 16 geometric defects that break a skill, including every lowered table (\cref{tab:sensitivity}), \chgb{and every world in \cref{tab:practice} where practice always failed}. \chgb{It misses three enlarged worlds, so it is a check, not a guarantee.} Mass and friction are invisible to Examine and are covered by practicing across 4 worlds with randomized mass and friction: on \emph{Mug-Box} this lifts success under an unseen mass $\times 3$ from 10\% to 100\% (\cref{tab:rand}); the $3\times$ heavy \emph{Open Drawer} stays unsolved.}

\begin{table}[t]
\centering\footnotesize
\caption{\chg{\textbf{Practicing across randomized physics.} Success (\%) in the correct world and under unseen physical defects\chgb{, on the only two tasks that a physical defect breaks (\cref{tab:sensitivity})}.}}
\label{tab:rand}
\setlength{\tabcolsep}{2.6pt}
\begin{tabular}{@{}llcccc@{}}
\toprule Task & Practice & Correct & Mass $\times 3$ & Fric.\ $\times 0.3$ & Fric.\ $\times 3$ \\\midrule
\multirow{2}{*}{Mug-Box} & Nominal only & 100\% & 10\% & 100\% & 100\% \\
 & Randomized & 100\% & \textbf{100\%} & 100\% & 100\% \\\midrule
\multirow{2}{*}{Open Drawer} & Nominal only & 98\% & 0\% & 100\% & 95--100\% \\
 & Randomized & 98\% & 0\% & 100\% & 100\% \\\bottomrule
\end{tabular}
\end{table}

\begin{table*}[t]
\centering\small
\caption{\chg{\textbf{Simulation and real-robot results across embodiments.} Sim.: success rate (\%) over 50 unseen test layouts per task (TEST 0--49). Real: successes/trials of the same frozen program on the real robot, without code edits or human intervention (--: not evaluated). The three target robots received no demonstration.}}
\label{tab:transfer}\label{tab:real}
\setlength{\tabcolsep}{5.5pt}
\begin{tabular}{@{}llccccccccc@{}}
\toprule
Robot & & Apple & Cubes & Stack Cups & Build Tower & Mug-Box & Open Drawer & Sweep & Lift-Box & Mean \\\midrule
\multirow{2}{*}{xTrainer (source)} & Sim. & 100\% & 98\% & 96\% & 100\% & 100\% & 100\% & 88\% & 94\% & 97\% \\
 & Real & 10/10 & 10/10 & 9/10 & 8/10 & 10/10 & 10/10 & 6/10 & 10/10 & 73/80 \\\midrule
\multirow{2}{*}{Dual-arm Franka} & Sim. & 94\% & 98\% & 100\% & 100\% & 94\% & 100\% & 78\% & 100\% & 96\% \\
 & Real & 10/10 & 10/10 & 10/10 & 10/10 & 10/10 & 8/10 & 8/10 & 10/10 & 76/80 \\\midrule
\multirow{2}{*}{ARX LIFT2} & Sim. & 86\% & 76\% & 90\% & 92\% & 84\% & 70\% & 58\% & 80\% & 80\% \\
 & Real & 7/10 & 8/10 & 9/10 & 7/10 & 9/10 & 9/10 & 6/10 & 6/10 & 61/80 \\\midrule
\multirow{2}{*}{Unitree G1} & Sim. & 86\% & 70\% & 100\% & 70\% & 80\% & 100\% & 32\% & 80\% & 77\% \\
 & Real & 5/10 & 2/10 & 3/10 & -- & 3/10 & -- & -- & -- & 13/40 \\\bottomrule
\end{tabular}
\end{table*}

\subsection{Self-Improvement Ablations (Q3)}
\label{sec:exp-ablation}

On \emph{Mug-Box} and \emph{Open Drawer}, starting from zero-shot programs that score 0\% (\cref{tab:ablation}), removing node-level failure logs and keyframes (outcome-only feedback) leaves performance at 12\% and 0\% after 8 rounds, whereas full RESET converges in 3--5 rounds to 100\% and 98\%: node-level failure evidence is essential. \chg{Removing hold-out gating, the rule that adopts a revision only if it does better on the hold-out layouts,} still solves \emph{Mug-Box} (100\%) but stays at 0\% on \emph{Open Drawer} \chgb{after all 8 rounds}.

\begin{table}[t]
\centering\footnotesize
\caption{\chg{\textbf{Ablations of Self-improve}, each started from the same zero-shot program that scores 0. Success: test success (\%). Rounds: self-improvement rounds used, out of a budget of 8.}}
\label{tab:ablation}
\setlength{\tabcolsep}{4.5pt}
\begin{tabular}{@{}lcccc@{}}
\toprule
 & \multicolumn{2}{c}{Mug-Box} & \multicolumn{2}{c}{Open Drawer} \\
\cmidrule(lr){2-3}\cmidrule(lr){4-5}
Variant & Success & Rounds & Success & Rounds \\\midrule
\textbf{Full RESET} & \textbf{100\%} & 3 & \textbf{98\%} & 5 \\
w/o hold-out gating & \textbf{100\%} & 3 & 0\% & 8 \\
Outcome-only feedback & 12\% & 8 & 0\% & 8 \\\bottomrule
\end{tabular}
\end{table}

\subsection{Cross-Embodiment Transfer (Q4)}
\label{sec:exp-transfer}

\paragraph{Transfer performance across morphologies.}
We hand xTrainer's 8 verified worlds and criteria to three target robots (\cref{tab:transfer}). \chg{Dual-arm Franka and ARX LIFT2 reach 95.5\% and 79.5\%, and Unitree G1, with five-finger hands instead of parallel jaws, reaches 77.3\% (32\% on \emph{Sweep}, where its fingers lack lateral stiffness on the flat brush handle).}

\chgb{\paragraph{Is it the shared world, or just copied code?}
To separate the two, we run the source robot's final programs unchanged on each target robot, with no practice. On Dual-arm Franka, whose arms are close to xTrainer's, five programs work as they are (94\%--100\%); the other three reach 10\%, 24\%, and 0\% (\emph{Stack Cups}, \emph{Sweep}, \emph{Lift-Box}) and rise to 100\%, 78\%, and 100\% after practice in the shared world. On ARX LIFT2 the source programs succeed on no task, and on Unitree G1 only on \emph{Apple} (18\%); after practice the two robots average 80\% and 77\%. The source robot's task graph and notes are not required either: given only the shared world and criterion and starting from scratch, LIFT2 reaches 94\% and 84\% on \emph{Mug-Box} and \emph{Open Drawer} and G1 reaches 74\% and 92\%, comparable to the 84\%, 70\%, 80\%, and 100\% obtained when they are inherited.}

\paragraph{Skill-level reuse analysis.}
Comparing function-level syntax trees across the final skill libraries (\cref{fig:reuse}), 94.5\% of Dual-arm Franka's skill functions are identical to another robot's and LIFT2 reuses 78.2\%, whereas Unitree G1 writes 86.7\% of its skills from scratch \chg{yet still reaches 77\% on average}. What carries over across embodiments is the verified world and criterion, not the code.

\subsection{Real-World Deployment and Sim-to-Real Agreement (Q5)}
\label{sec:exp-real}

\chg{We deploy the frozen programs on four physical robots with no code edits and no human intervention (10 trials per task, objects placed by hand at varied positions). The source xTrainer and the zero-demonstration Dual-arm Franka and ARX LIFT2 run all 8 tasks, and Unitree G1 runs 4 (\cref{tab:real}, \cref{fig:real_rollouts}). In total, 223 of 280 real-robot trials succeed (80\%)\chgb{, and 210 of 240 (88\%) on the three fixed-base robots}. xTrainer succeeds in 73/80 trials (91\%) and Dual-arm Franka in 76/80 (95\%), close to their simulation rates (97\% and 96\%); both arms have a large workspace and localize objects accurately in our scenes. ARX LIFT2 reaches 61/80 (76\%, against 80\% in simulation): its arms have a smaller reachable workspace, so some layouts place objects near the limit of its reach. Unitree G1 succeeds in 13/40 trials on its four tasks (33\%, against 84\% in simulation on the same tasks). Its whole-body controller shifts the base whenever the arms move, and the open-loop program does not compensate for this shift (\cref{sec:conclusion}).}

\chg{\paragraph{Does the verified world predict real outcomes?}
For every real trial we register the first camera frame to recover the actual object layout, reproduce that layout in the verified world, and run the same frozen program there. Simulation and the real robot give the same outcome in 95\% of the 80 trials on xTrainer and in 90\% on ARX LIFT2, a robot that took no part in building or examining the world.}
\paragraph{Cost.}
\label{sec:exp-cost}
Self-improve converges in 2--8 rounds, using 100--320 simulation episodes \chgb{and about 30M LLM tokens per task and robot. Deployment uses no LLM tokens.}

\section{Limitations and Conclusion}
\label{sec:conclusion}

\paragraph{Limitations.}
RESET currently targets rigid, articulated, and quasi-static sweeping tasks, not deformable cloth or dynamic throwing. Extreme latent parameter jumps that demand a wholly different contact topology (e.g., a $3\times$ heavy drawer) remain beyond parameter randomization. \chg{Finally, practice does not yet model the effect of a humanoid's whole-body controller: on Unitree G1 the controller shifts the base whenever the arms move, so real-robot success falls well below that of fixed-base arms (\cref{tab:real}); adapting RESET to real humanoids is left to future work.}

\paragraph{Conclusion.}
\chg{We presented RESET, which brings agent-driven code self-improvement from hand-built benchmarks to real tasks. One real demonstration provides the practice world, the success criterion, and the reference for checking that world, so what a task requires is separated from how a particular robot performs it. \chgb{The rebuilt world need not be perfect: practice tolerates most reconstruction errors, and checking the world against the demonstration flags those that would mislead it.} More broadly, RESET offers a one-shot, cross-embodiment route to robot skills that is carried by an agent's general knowledge of the world and of code: a user demonstrates a task once, and different robots obtain inspectable, ready-to-run skills.}

{
    \small
    \bibliographystyle{ieeenat_fullname}
    \bibliography{main}
}

\clearpage
\appendix
\setcounter{figure}{0}\renewcommand{\thefigure}{S\arabic{figure}}
\renewcommand{\dbltopfraction}{0.92}\renewcommand{\textfraction}{0.05}\setcounter{dbltopnumber}{2}

\twocolumn[{%
  \centering
  {\Large\bfseries Appendix\par}
  \vspace{1.2em}
  \captionsetup{type=figure}
\setlength{\tabcolsep}{1.5pt}
\newcommand{\tw}{0.237\textwidth}
\newcommand{\goal}[1]{\parbox[t]{\tw}{\centering\footnotesize #1}}
\begin{tabular}{@{}c@{\hspace{3pt}}cccc@{}}
 & Apple & Cubes & Stack Cups & Build Tower \\
\rl{1.06in}{Demonstration} & \includegraphics[width=\tw]{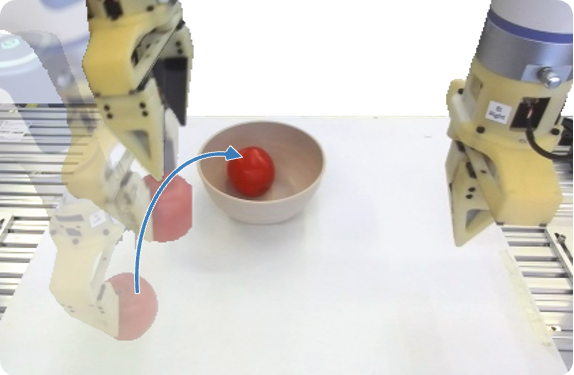} & \includegraphics[width=\tw]{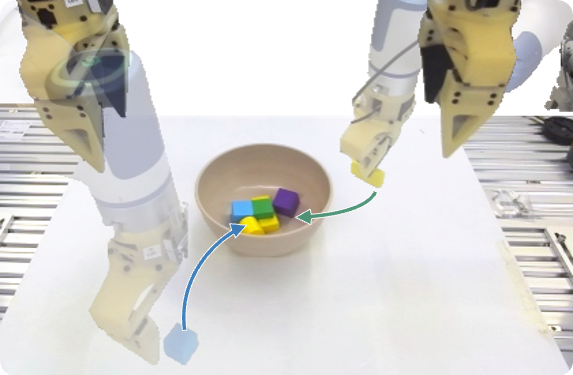} & \includegraphics[width=\tw]{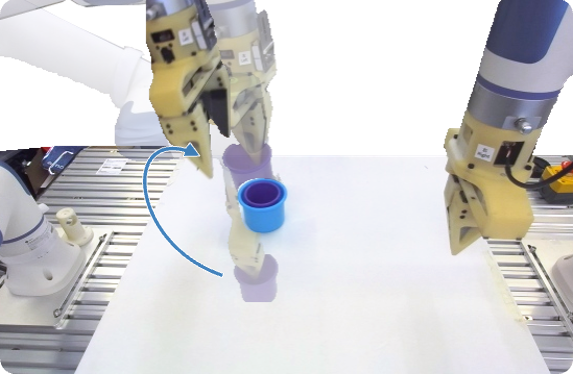} & \includegraphics[width=\tw]{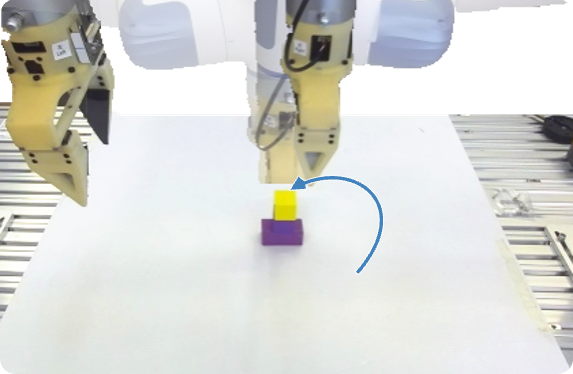} \\
\rl{0.92in}{Rebuilt world} & \includegraphics[width=\tw]{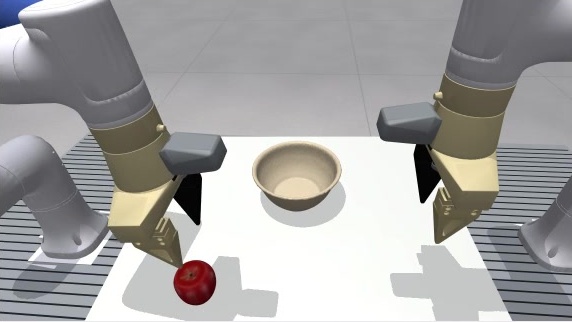} & \includegraphics[width=\tw]{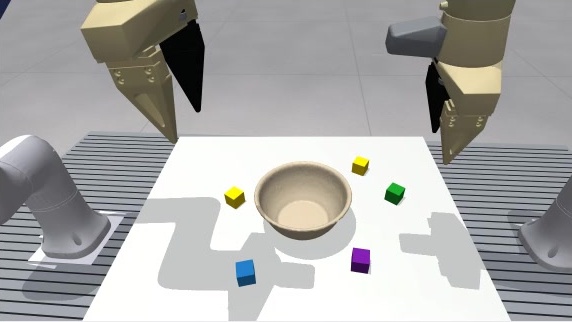} & \includegraphics[width=\tw]{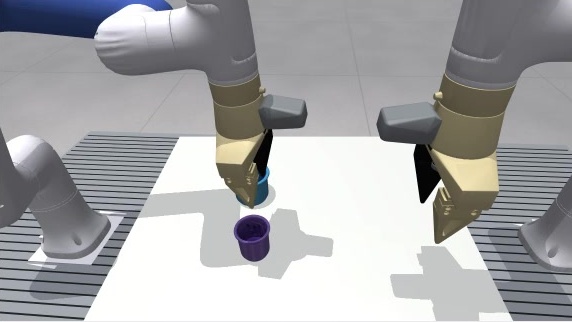} & \includegraphics[width=\tw]{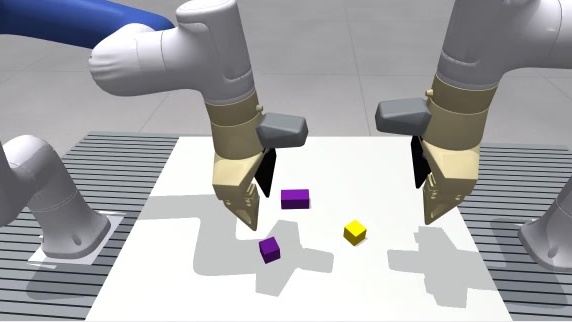} \\
\rl{0.28in}{Goal} & \goal{Apple inside the bowl.} & \goal{All five blocks inside the bowl.} & \goal{Purple cup nested in\\the blue cup, both upright.} & \goal{Purple cube on the purple bar,\\yellow cube on top.} \\[12pt]
 & Mug-Box & Open Drawer & Sweep & Lift-Box \\
\rl{1.06in}{Demonstration} & \includegraphics[width=\tw]{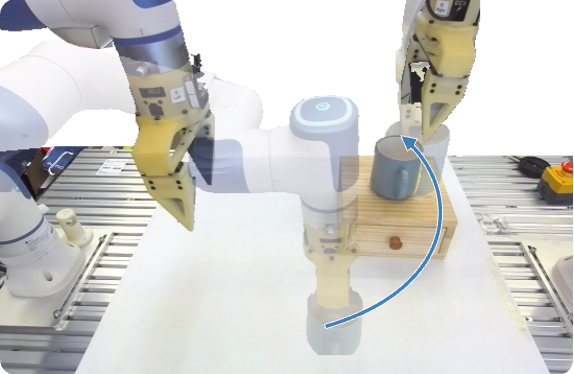} & \includegraphics[width=\tw]{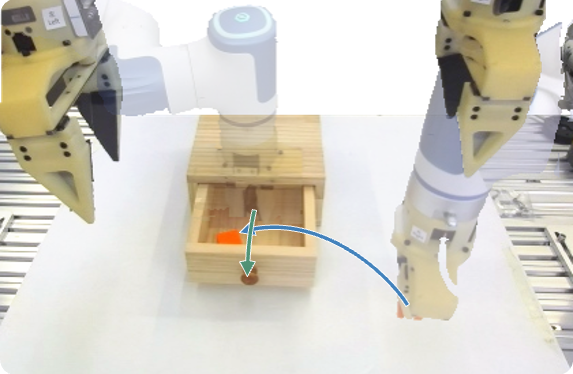} & \includegraphics[width=\tw]{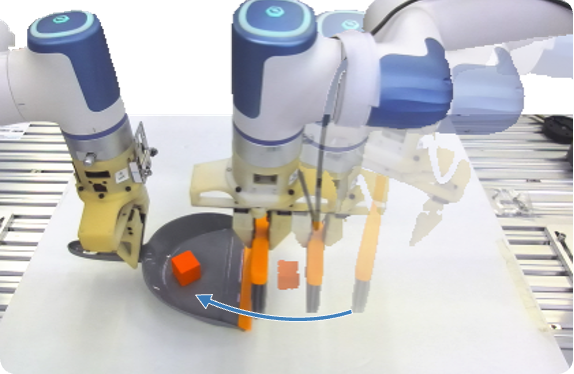} & \includegraphics[width=\tw]{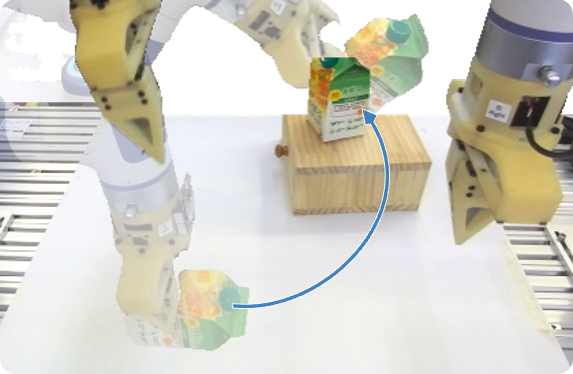} \\
\rl{0.92in}{Rebuilt world} & \includegraphics[width=\tw]{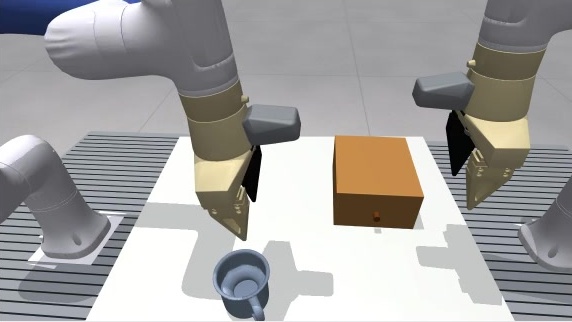} & \includegraphics[width=\tw]{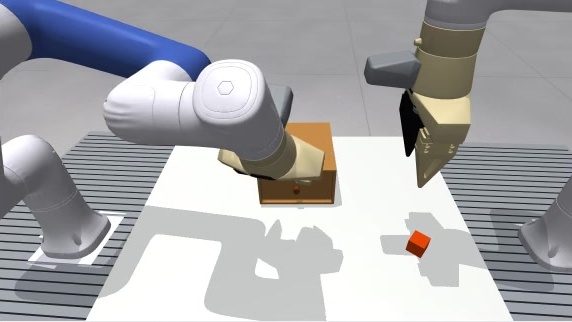} & \includegraphics[width=\tw]{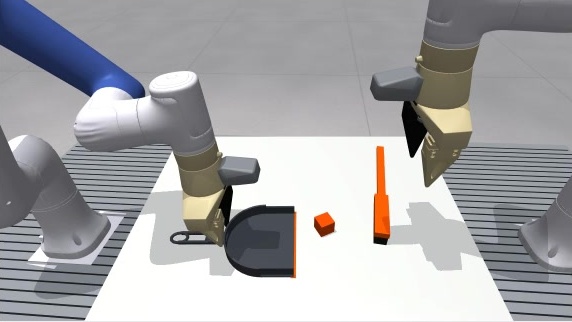} & \includegraphics[width=\tw]{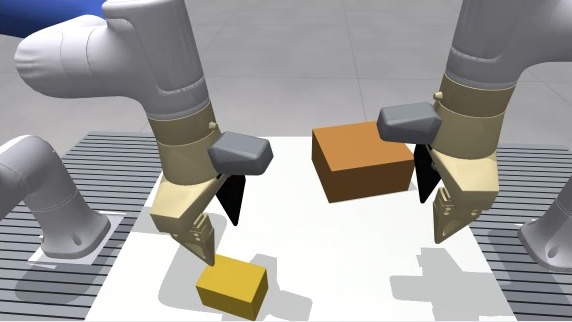} \\
\rl{0.28in}{Goal} & \goal{Mug upright on the wooden box.} & \goal{Drawer opened and the\\orange block inside.} & \goal{Block swept into the dustpan\\with the brush.} & \goal{Juice carton on top of the\\wooden box, standing upright.} \\
\end{tabular}
\captionof{figure}{\textbf{The eight tasks and their rebuilt worlds.} Demonstration: the single demonstration on the source robot xTrainer, with the demonstrated object motion drawn as arrows. Rebuilt world: the interactive world rebuilt from that demonstration, rendered from the demonstration camera (a larger view of Fig.~3(a)). Goal: what the success criterion requires, in plain words.}
\label{apx:fig:tasks}
  \vspace{1.2em}
}]

\section{Tasks and rebuilt worlds}
\label{apx:sec:tasks}

Each of the eight tasks is demonstrated once on the source robot xTrainer (Sec.~4.1). \cref{apx:fig:tasks} shows each demonstration together with the world rebuilt from it (Sec.~3.3) and the goal that the success criterion reads from the demonstrated final state. Every criterion also requires the objects to be released and still at the end of the episode.

\section{Examine examples}
\label{apx:sec:examine}

\begin{figure*}[t]
\centering
\setlength{\tabcolsep}{2pt}
\newcommand{\tw}{0.40\textwidth}
\begin{tabular}{@{}cc@{}}
\includegraphics[width=\tw]{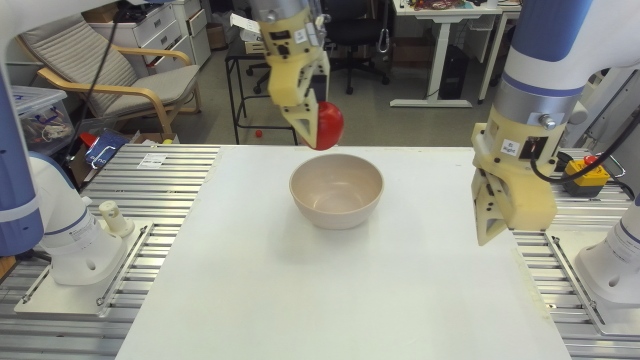} &
\includegraphics[width=\tw]{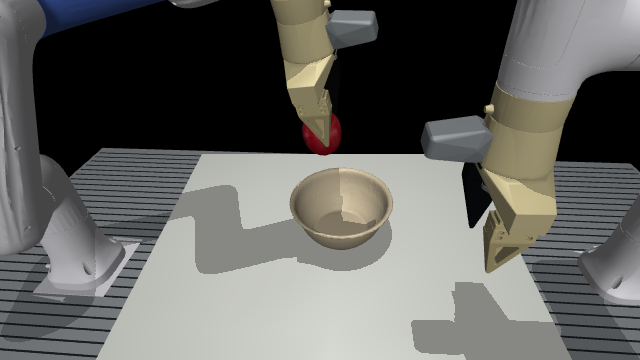} \\
(a) Real demonstration & (b) Correct world \\[5pt]
\includegraphics[width=\tw]{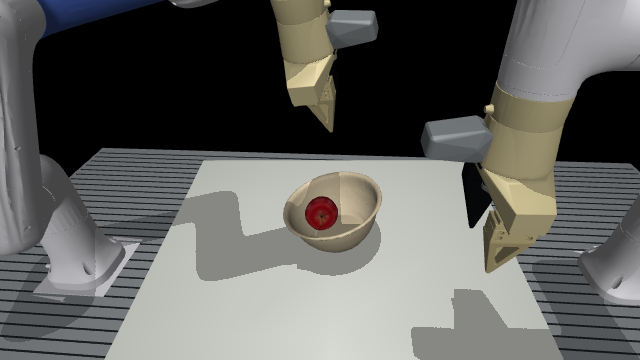} &
\includegraphics[width=\tw]{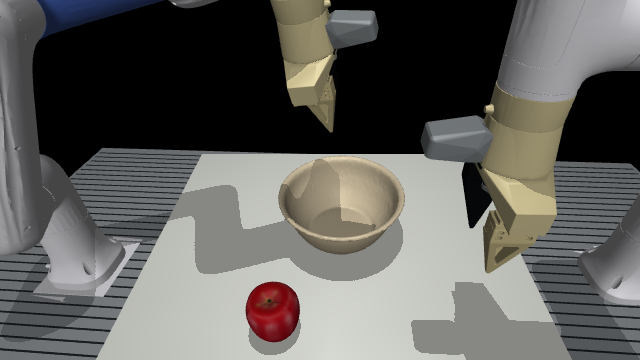} \\
(c) Table $2\,\mathrm{cm}$ too low & (d) Objects scaled $\times 1.2$ \\
\end{tabular}
\caption{\textbf{The dynamic channel of Examine on the Apple task.} All four images show the same moment of the demonstration, shortly after the apple is lifted. (b)--(d) replay the recorded joint and gripper commands open-loop in three rebuilt worlds; the apple is never attached to the gripper, so it moves only through contact. (b) The apple is held in the gripper, as in the demonstration. (c) The gripper is empty and the apple lies in the bowl. (d) The gripper is empty and the apple is still on the table.}
\label{apx:fig:dynamic}
\vspace{12pt}
\setlength{\tabcolsep}{2pt}
\renewcommand{\tw}{0.40\textwidth}
\begin{tabular}{@{}cc@{}}
\includegraphics[width=\tw]{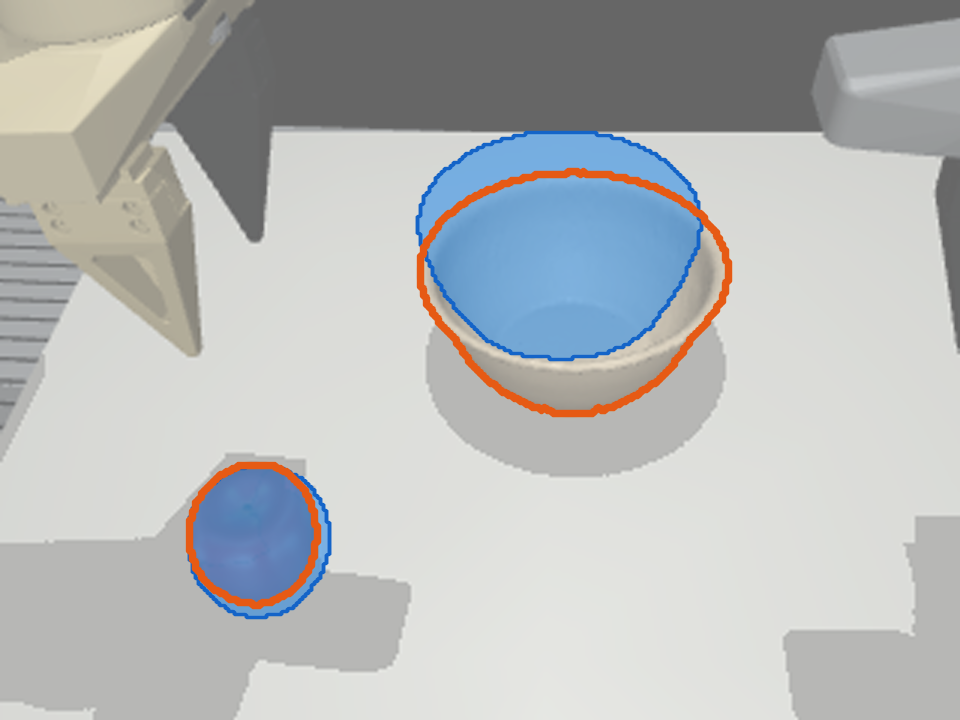} &
\includegraphics[width=\tw]{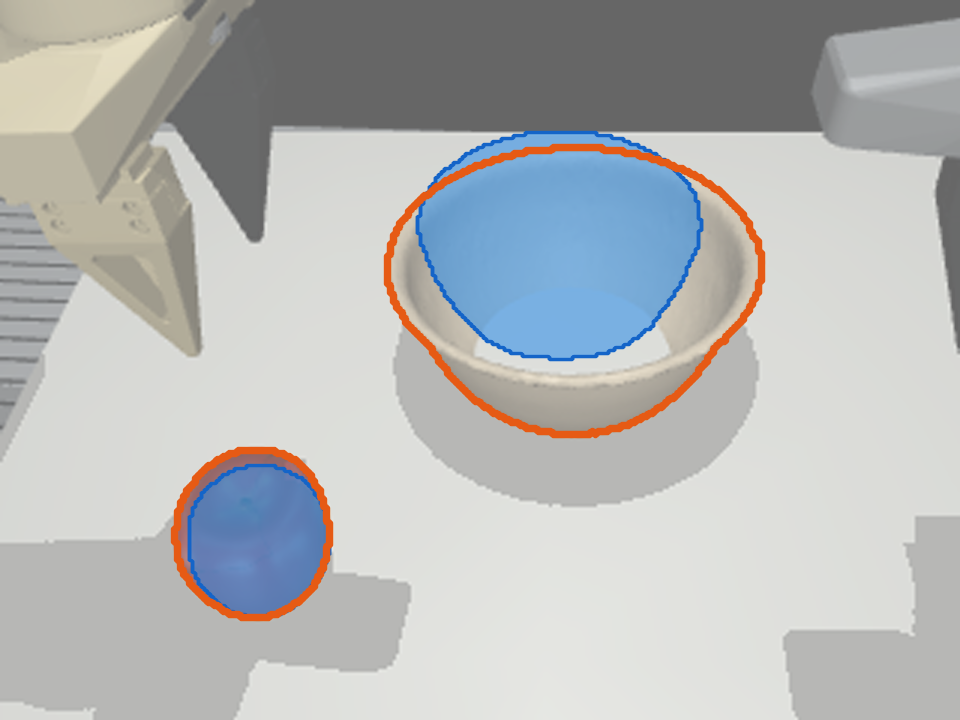} \\
(a) Correct world & (b) Objects scaled $\times 1.2$ \\
\end{tabular}
\caption{\textbf{The visual channel of Examine on the Apple task.} Each rebuilt world is rendered at the recorded joint angles of the first frame, before contact (render shown lightened and cropped to the objects). \textcolor{silblue}{Blue}: the demonstration masks of the apple and the bowl. \textcolor{silorange}{Orange}: outlines of the same two objects rendered in the rebuilt world. The channel measures the silhouette IoU of each object against its demonstration mask. In (b), both rendered outlines are larger than in (a).}
\label{apx:fig:visual}
\end{figure*}

Examine (Sec.~3.4) checks a rebuilt world against the demonstration through two channels. \cref{apx:fig:dynamic,apx:fig:visual} show both channels on the Apple task, in the correct world and in worlds with one of the geometric defects injected in Sec.~4.3: the table $2\,\mathrm{cm}$ too low, and all objects scaled by $\times 1.2$.

\section{One round of Self-improve}
\label{apx:sec:selfimprove}

\begin{figure*}[t]
\centering
\newlength{\iw}\setlength{\iw}{0.29\textwidth}
\newlength{\mw}\setlength{\mw}{0.345\textwidth}
\newcommand{\codeline}[3]{\colorbox{#2}{\parbox{\dimexpr\mw-2\fboxsep\relax}{\color{#1}\ttfamily\scriptsize #3}}}
\setlength{\fboxsep}{2pt}
\begin{tabular}{@{}c@{\hspace{3pt}}c@{\hspace{3pt}}c@{\hspace{3pt}}c@{\hspace{3pt}}c@{}}
\raisebox{-.5\height}{\includegraphics[width=\iw]{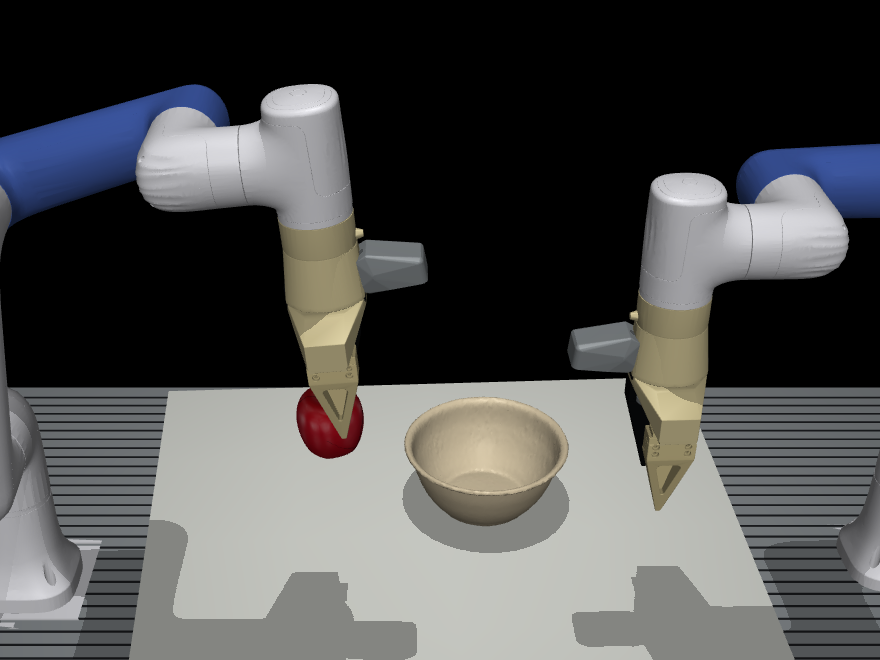}} & $\rightarrow$ &
\begin{minipage}[c]{\mw}\centering
{\footnotesize (b) Failure record}\\[2pt]
\fbox{\parbox{\dimexpr\mw-2\fboxsep-0.8pt\relax}{\centering\ttfamily\scriptsize
\textcolor{delfg}{move\_fail\ \ arm=left}\\[1pt]
\textcolor{boxgray}{IK did not converge}}}\\[1pt]
$\downarrow$\\[1pt]
{\footnotesize (c) Code revision}\\[2pt]
\codeline{delfg}{delbg}{-\ self.go(arm, np.r\_[cur[:2], h], R, \dots}\\[-1.5pt]
\codeline{addfg}{addbg}{+\ for hh in (h, h - .04, h - .07, \dots):}\\[-1.5pt]
\codeline{addfg}{addbg}{+\ \ \ self.go(arm, np.r\_[cur[:2], hh], \dots}
\end{minipage} & $\rightarrow$ &
\raisebox{-.5\height}{\includegraphics[width=\iw]{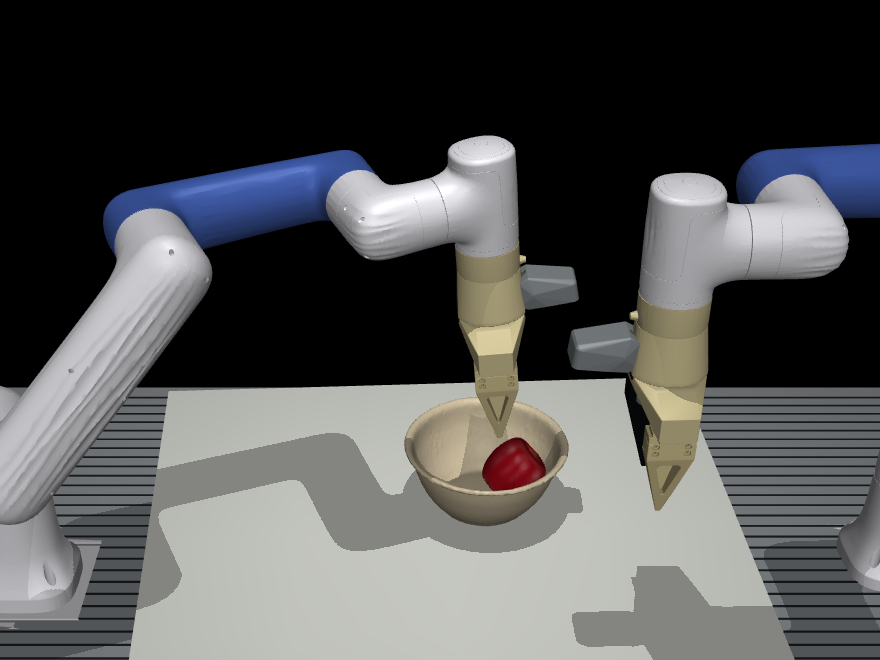}} \\[2pt]
\footnotesize (a) Failed practice episode & & & & \footnotesize (d) After the revision \\
\end{tabular}
\caption{\textbf{One round of Self-improve on the Apple task.} (a) A practice episode of the round-0 program in the rebuilt world: the apple is held beside the bowl. (b) The failure record of that episode. (c) The revision the coding agent made in round 1 (lines shortened with an ellipsis): the single move of the arm to height \texttt{h} becomes a loop over decreasing heights. (d) A practice episode of the revised program: the apple is in the bowl.}
\label{apx:fig:selfimprove}
\end{figure*}

In each round, Self-improve (Sec.~3.5) runs the program on diagnostic layouts, reads the status and failure reason of every node, and revises the code. \cref{apx:fig:selfimprove} follows one such revision on the Apple task on the source robot xTrainer, from a failed practice episode of the round-0 program to the program after the round-1 revision.

\section{Simulation execution on four robots}
\label{apx:sec:sim}

In Transfer (Sec.~3.6), each target robot practices in the same rebuilt worlds with the same criteria, without a demonstration of its own. \cref{apx:fig:sim} shows one key frame of the final program of each of the four robots on each of the eight tasks. Success rates are reported in Tab.~5.

\begin{figure*}[p]
\centering
\setlength{\tabcolsep}{1.5pt}
\newcommand{\tw}{0.236\textwidth}

\begin{tabular}{@{}c@{\hspace{3pt}}cccc@{}}
 & Apple & Cubes & Stack Cups & Build Tower \\
\rl{0.91in}{xTrainer (source)} & \includegraphics[width=\tw]{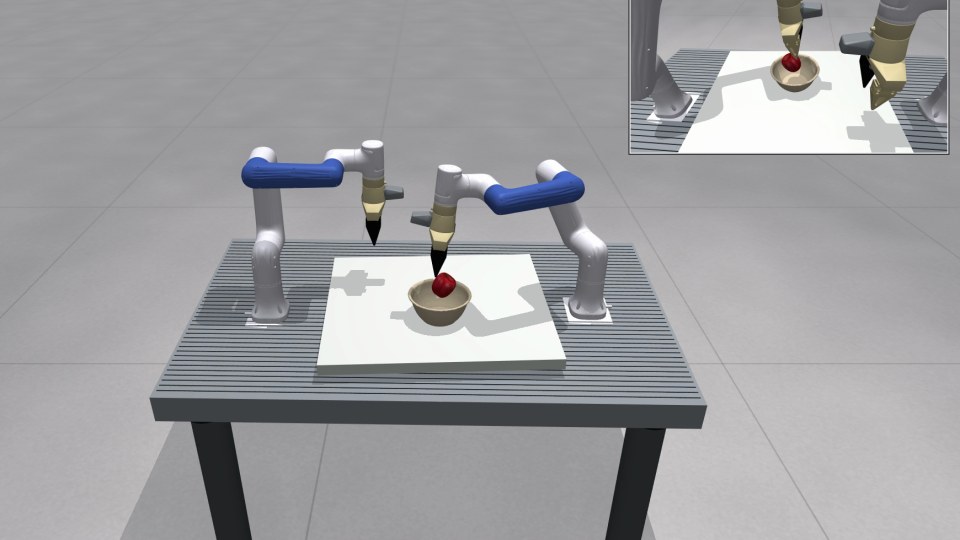} & \includegraphics[width=\tw]{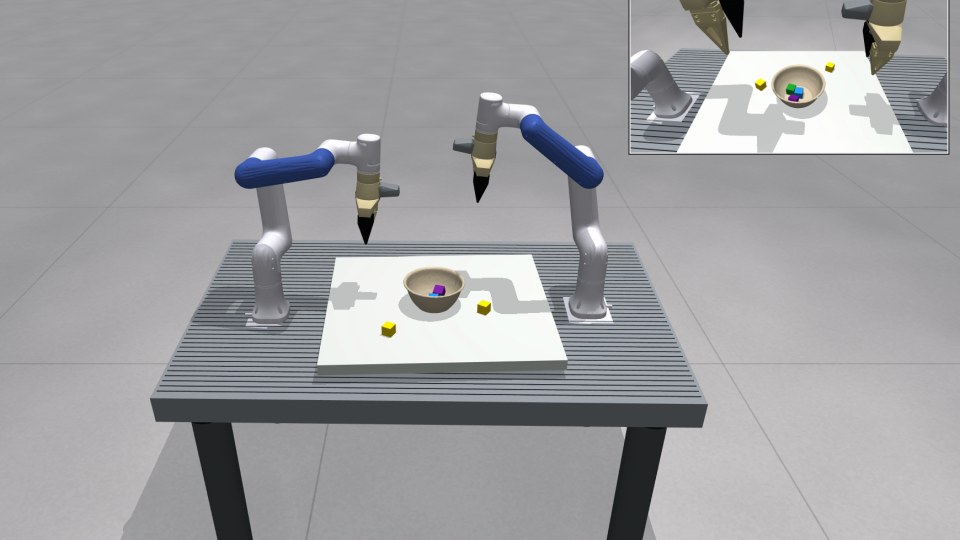} & \includegraphics[width=\tw]{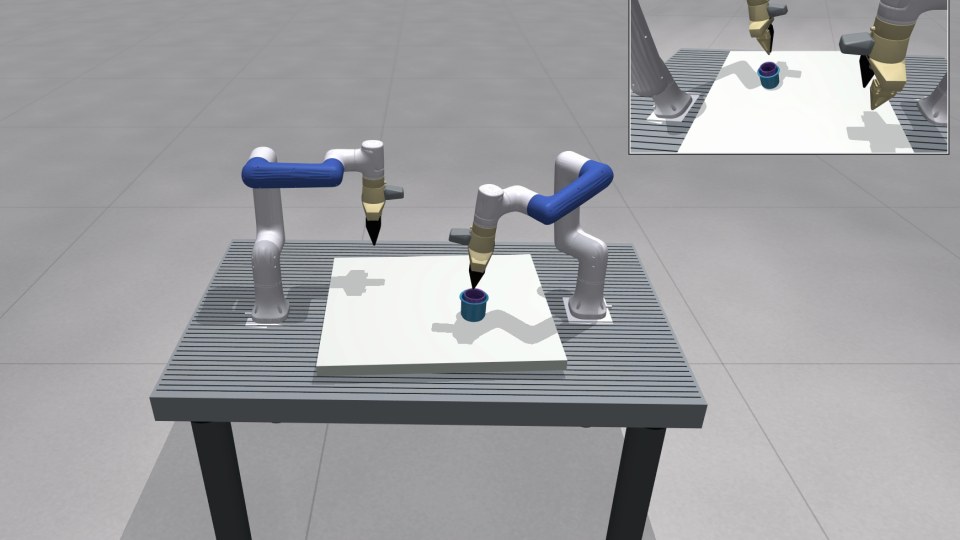} & \includegraphics[width=\tw]{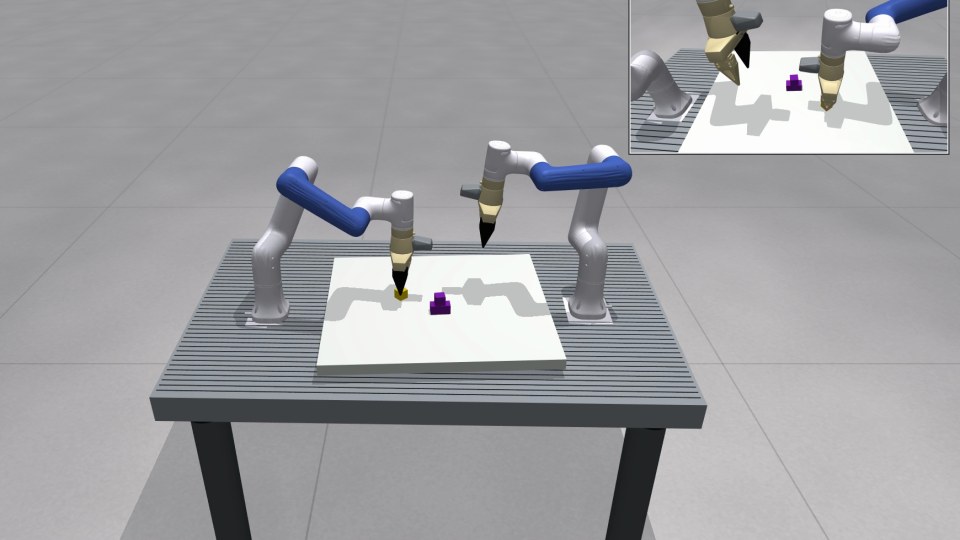} \\
\rl{0.91in}{Dual-arm Franka} & \includegraphics[width=\tw]{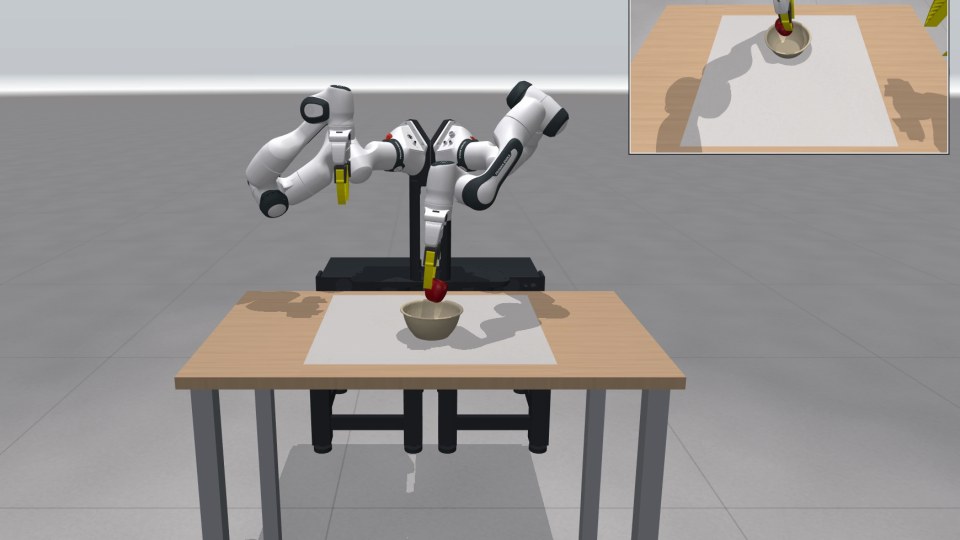} & \includegraphics[width=\tw]{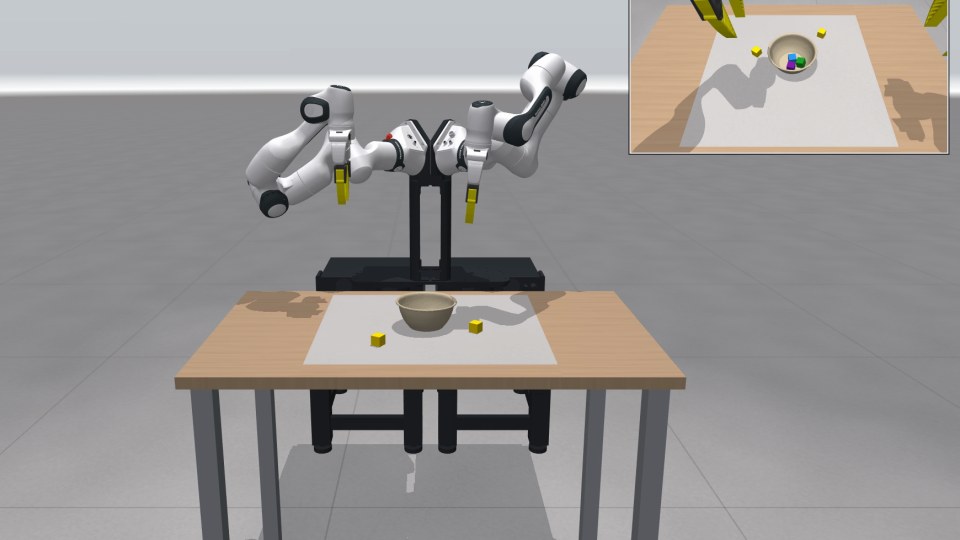} & \includegraphics[width=\tw]{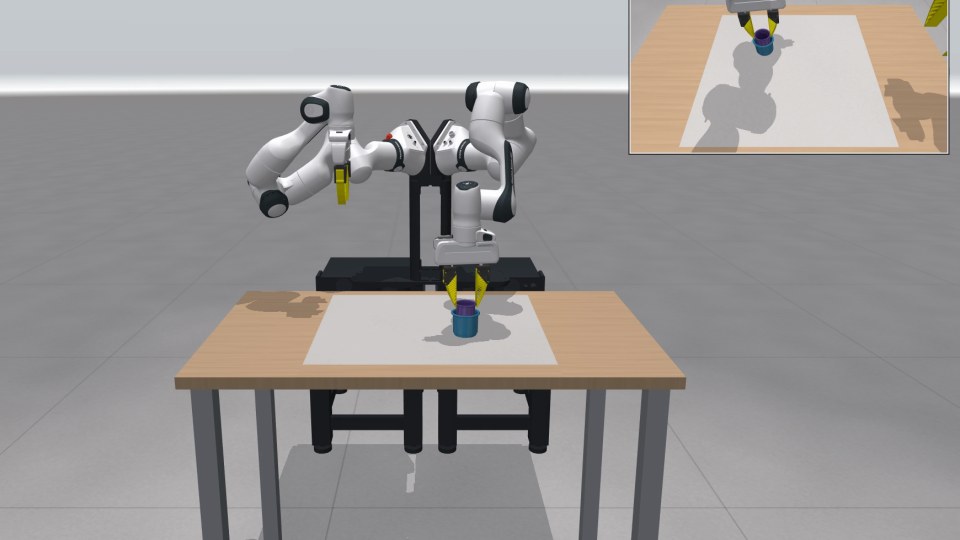} & \includegraphics[width=\tw]{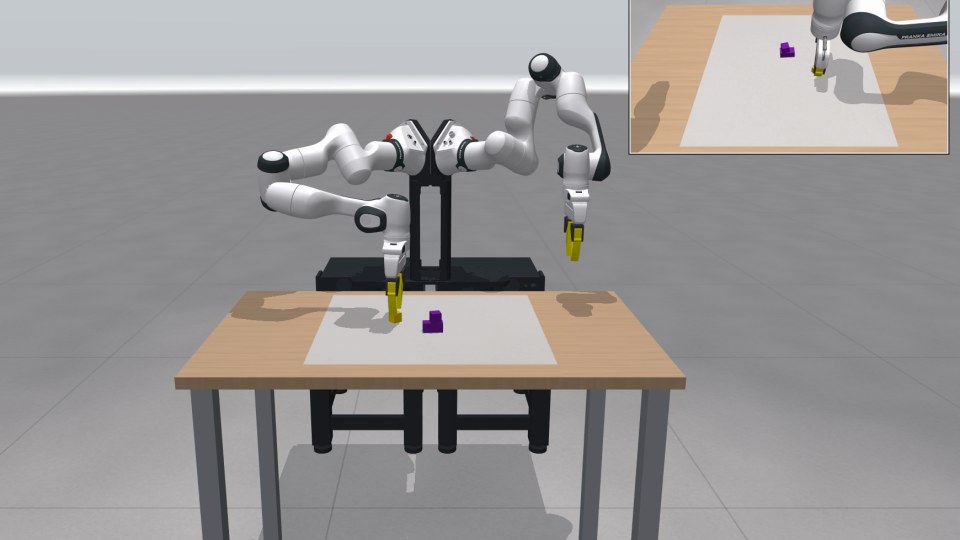} \\
\rl{0.91in}{ARX LIFT2} & \includegraphics[width=\tw]{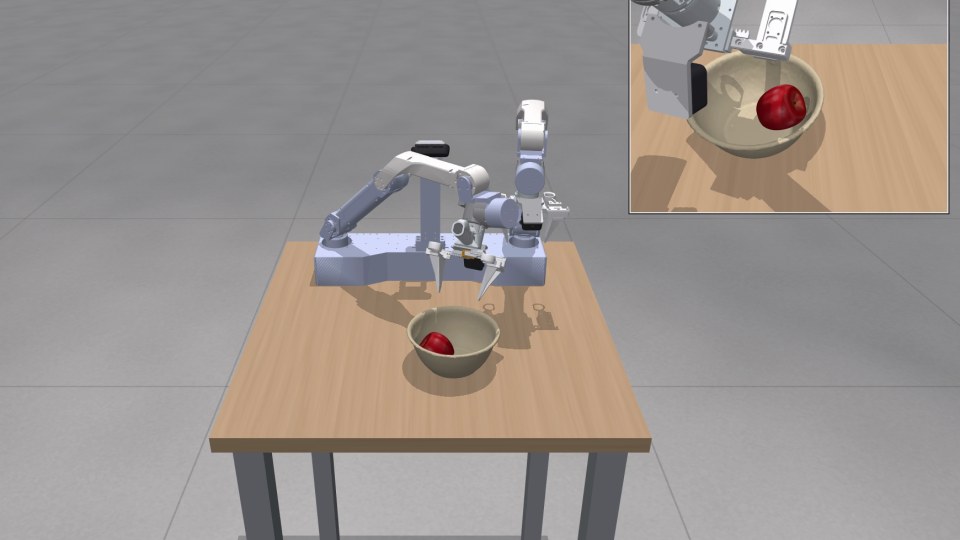} & \includegraphics[width=\tw]{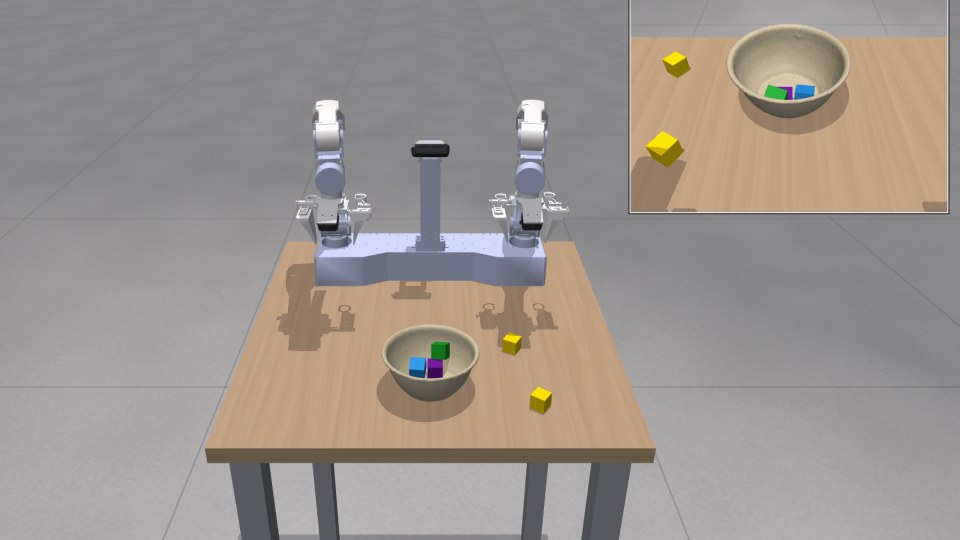} & \includegraphics[width=\tw]{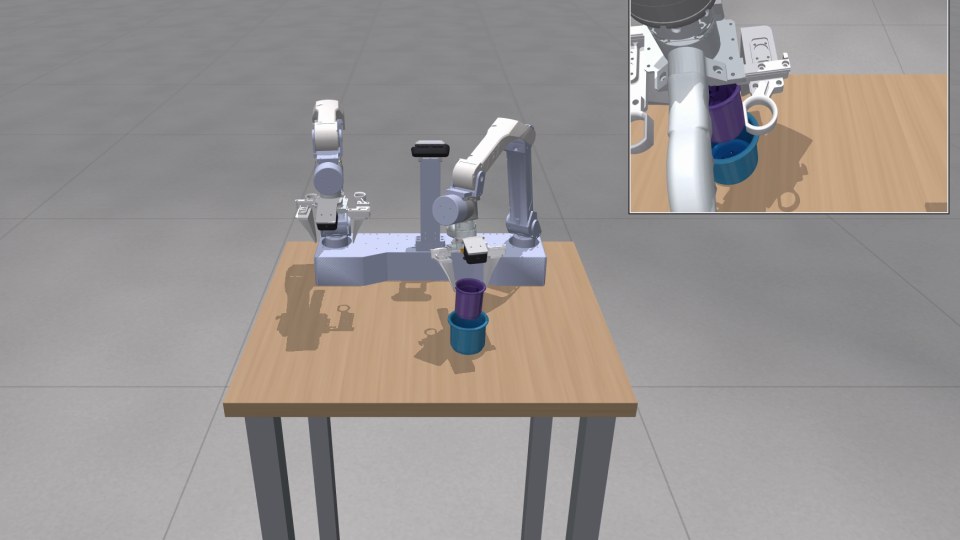} & \includegraphics[width=\tw]{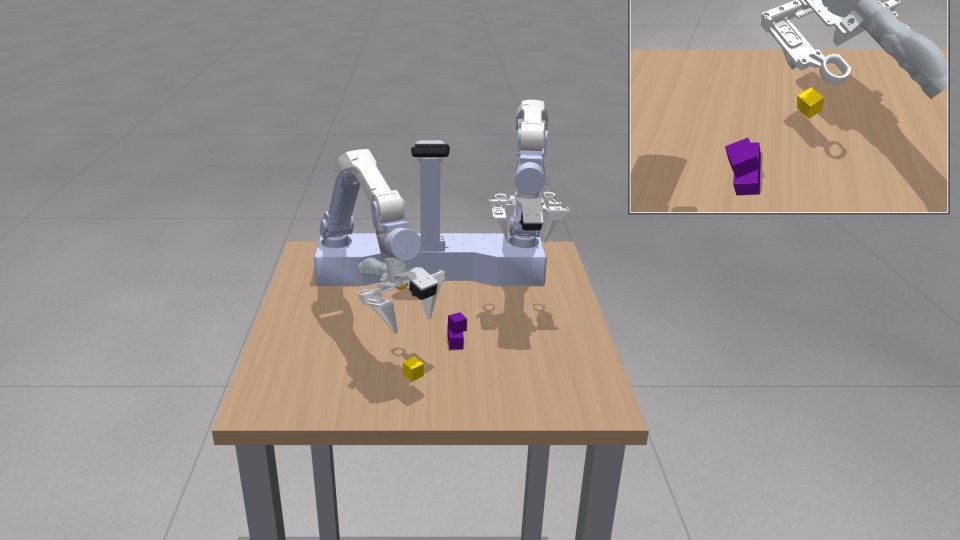} \\
\rl{0.91in}{Unitree G1} & \includegraphics[width=\tw]{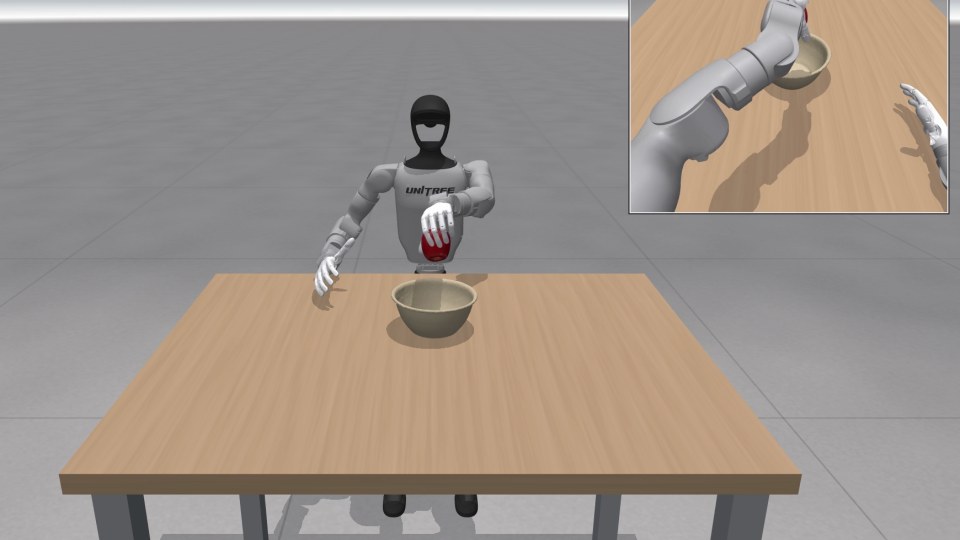} & \includegraphics[width=\tw]{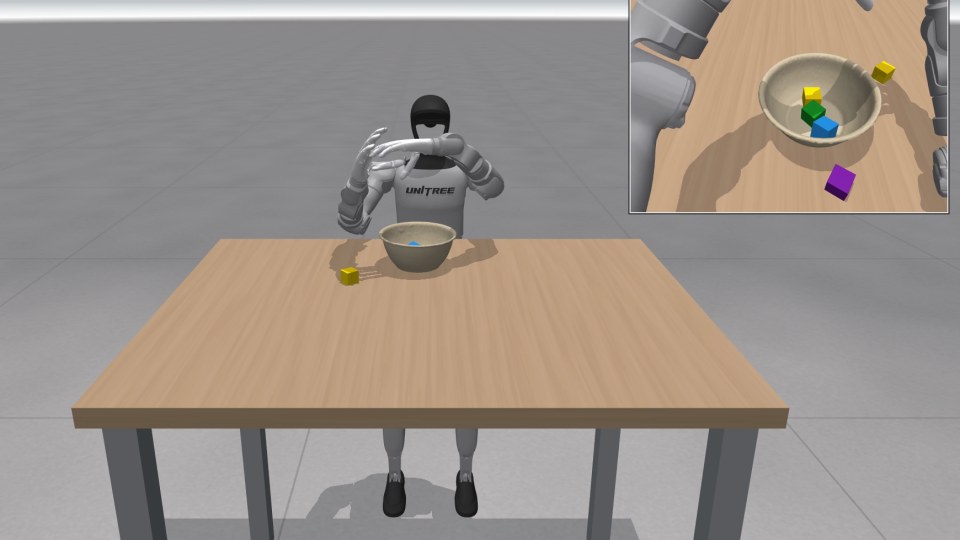} & \includegraphics[width=\tw]{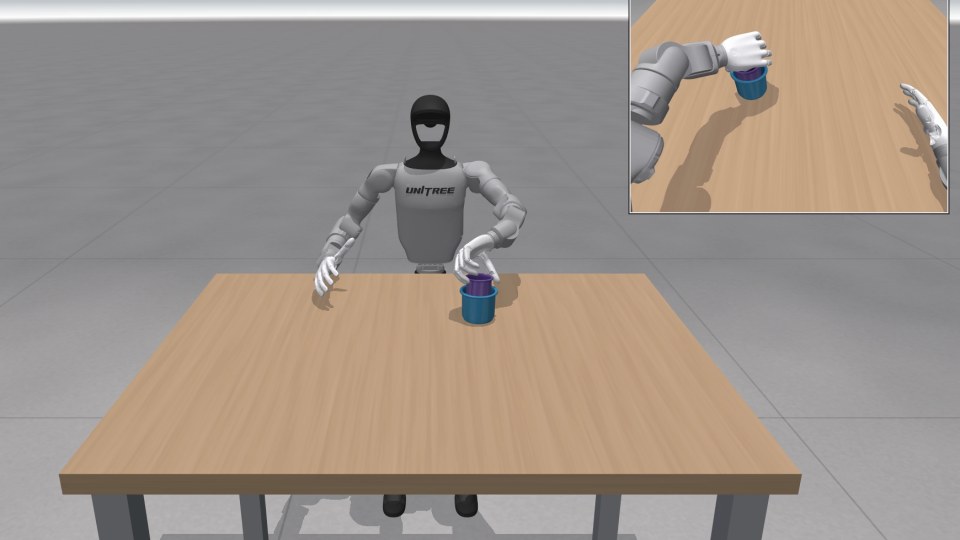} & \includegraphics[width=\tw]{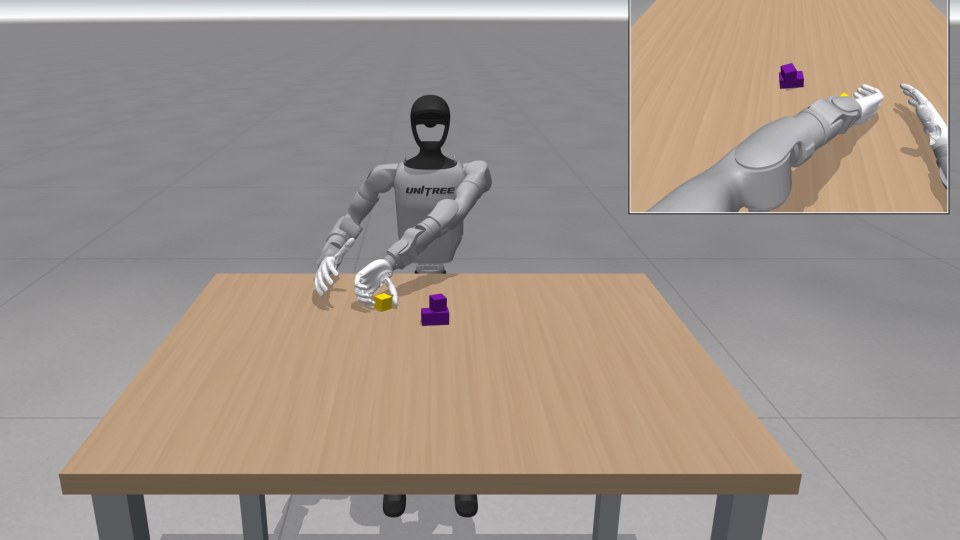} \\
\noalign{\vspace{6pt}}
 & Mug-Box & Open Drawer & Sweep & Lift-Box \\
\rl{0.91in}{xTrainer (source)} & \includegraphics[width=\tw]{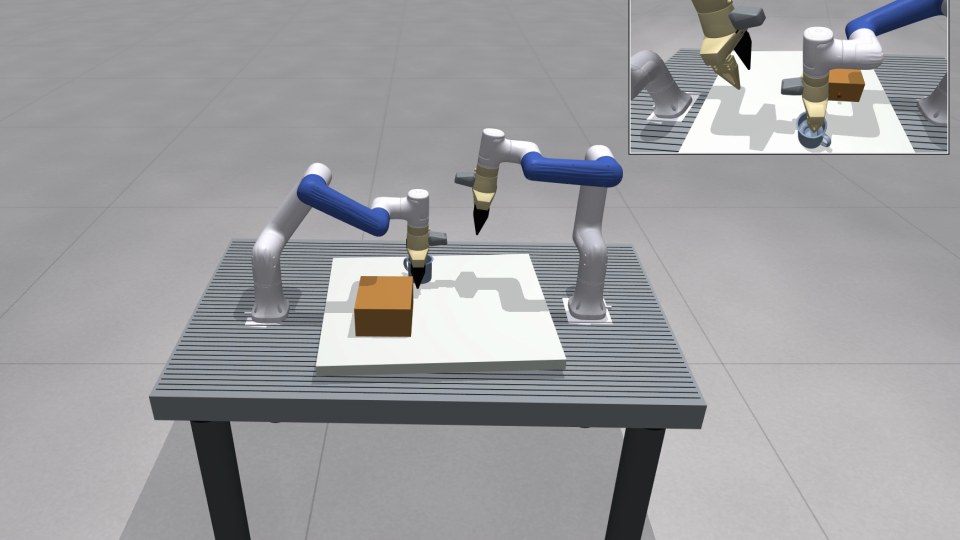} & \includegraphics[width=\tw]{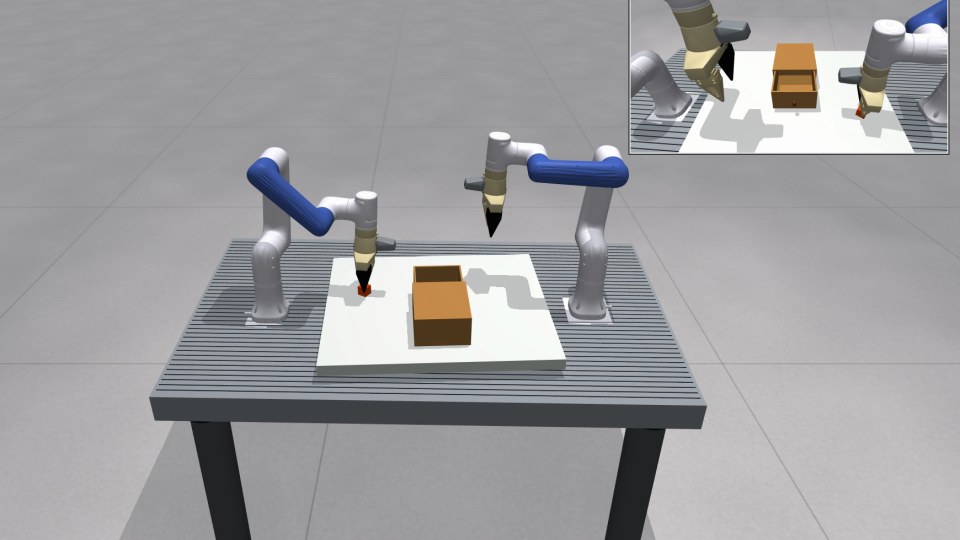} & \includegraphics[width=\tw]{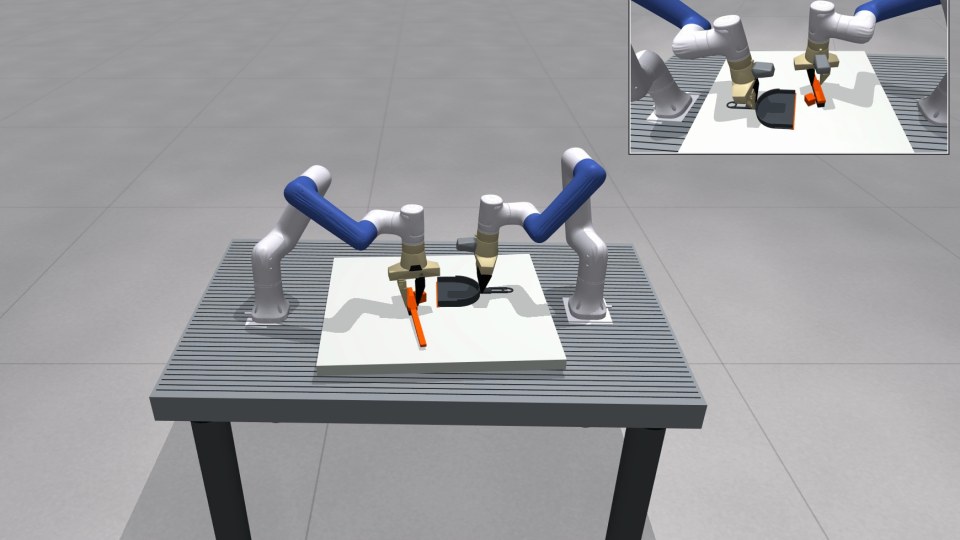} & \includegraphics[width=\tw]{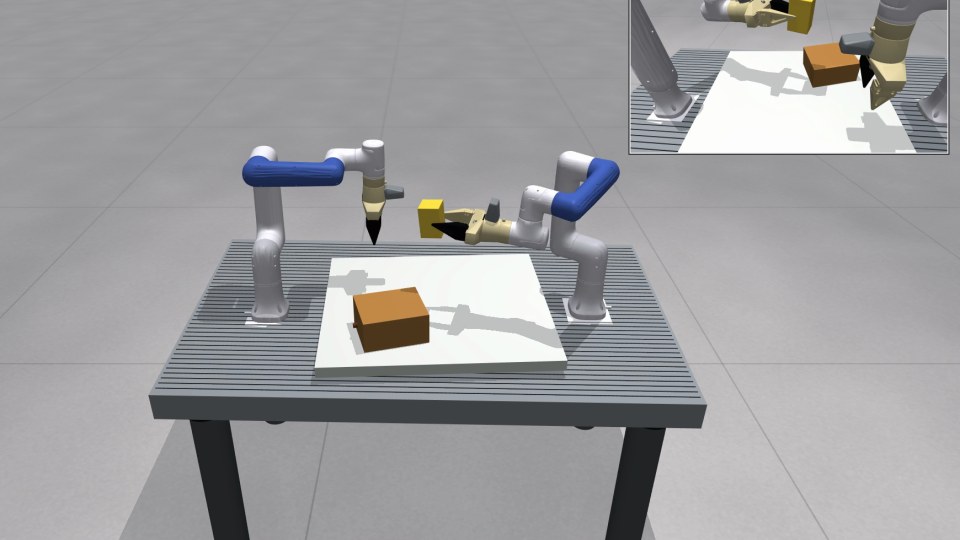} \\
\rl{0.91in}{Dual-arm Franka} & \includegraphics[width=\tw]{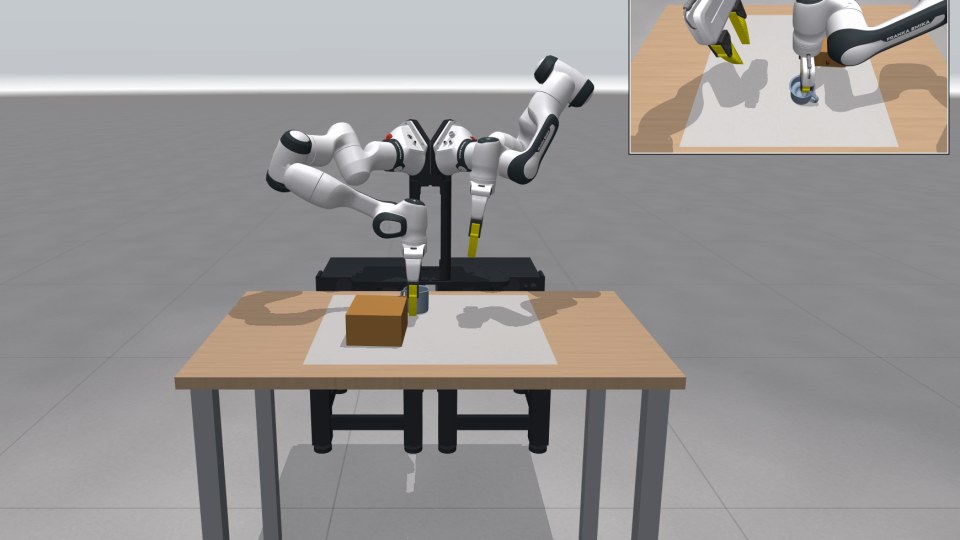} & \includegraphics[width=\tw]{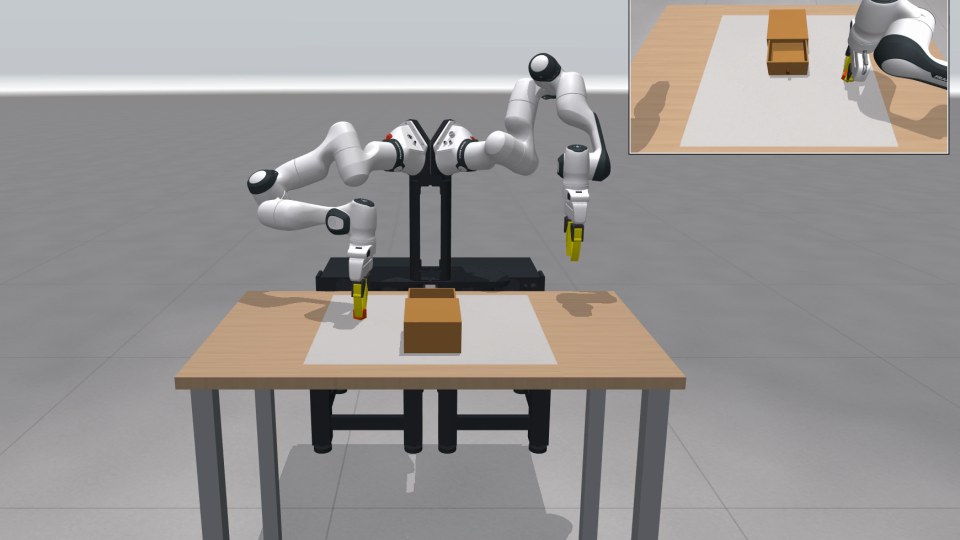} & \includegraphics[width=\tw]{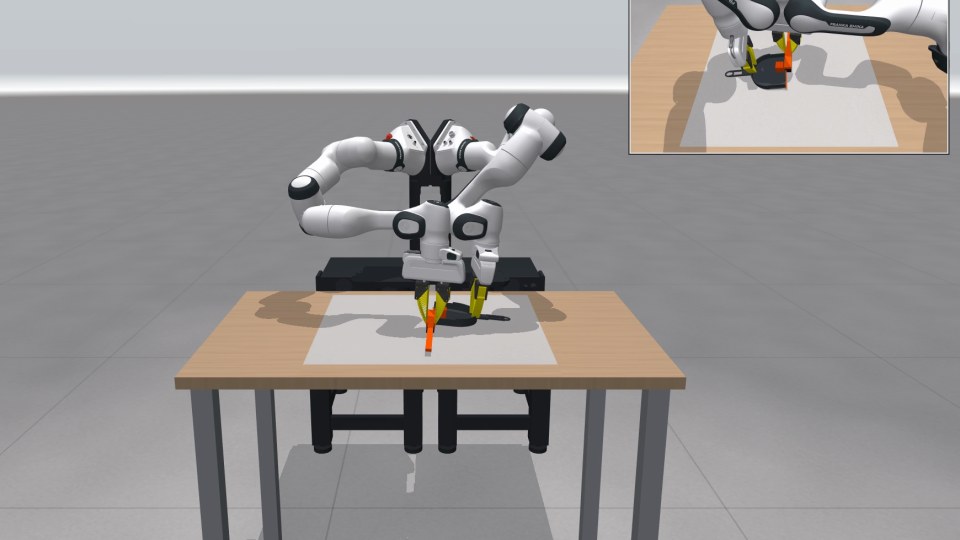} & \includegraphics[width=\tw]{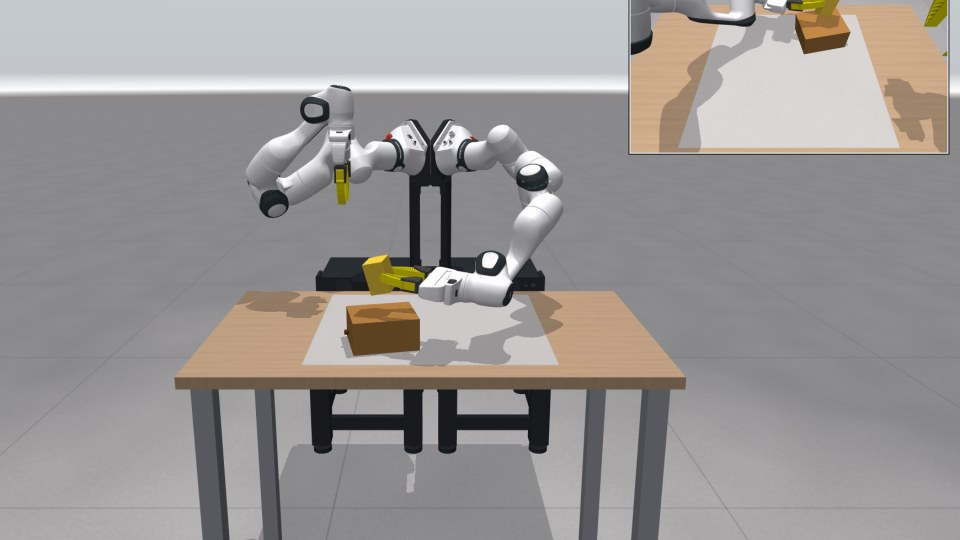} \\
\rl{0.91in}{ARX LIFT2} & \includegraphics[width=\tw]{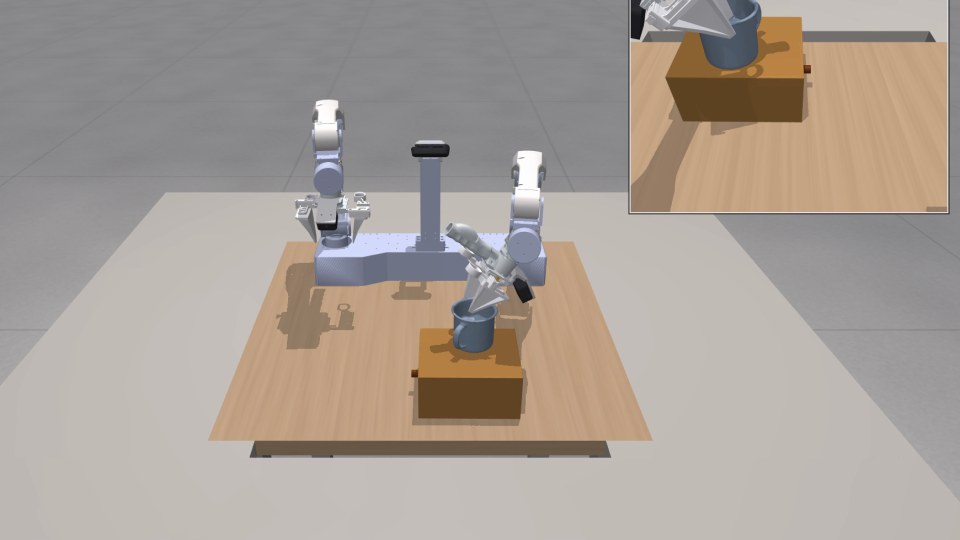} & \includegraphics[width=\tw]{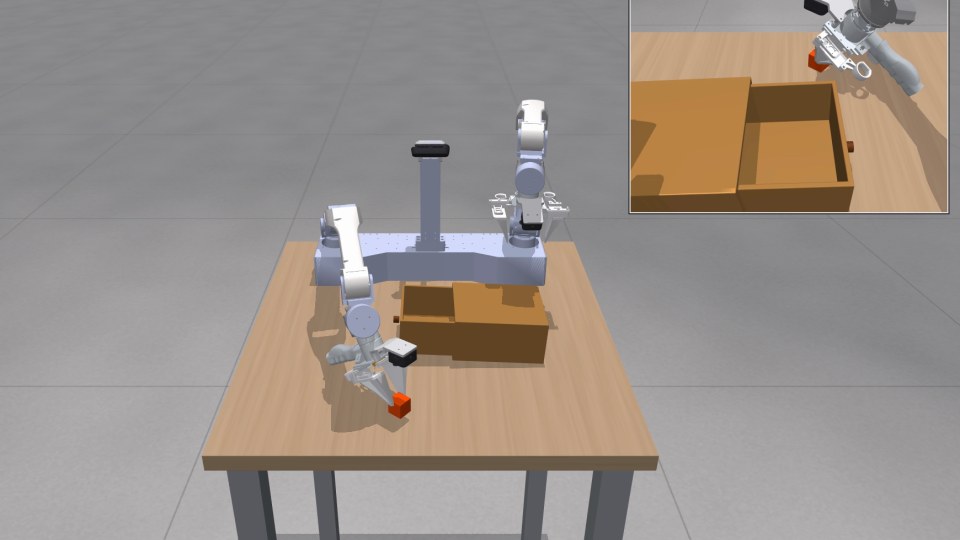} & \includegraphics[width=\tw]{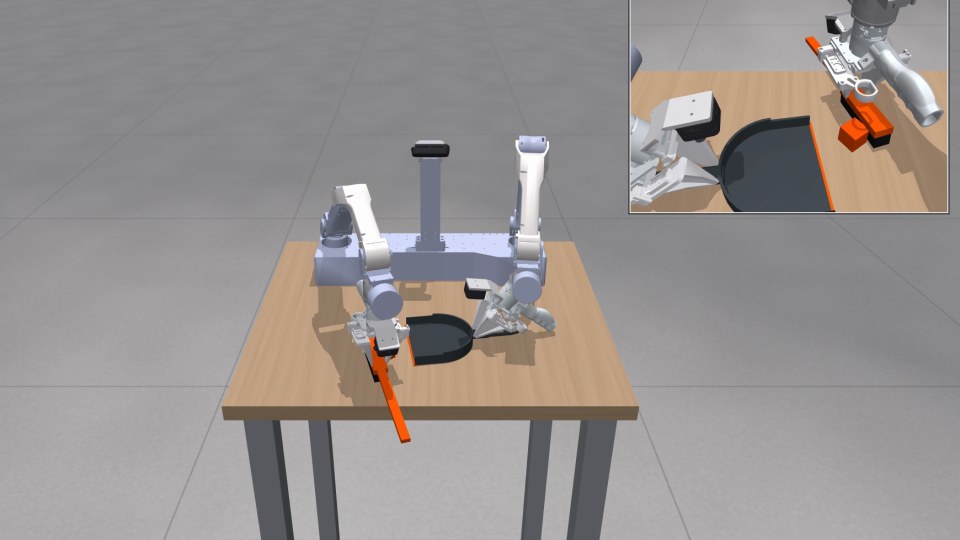} & \includegraphics[width=\tw]{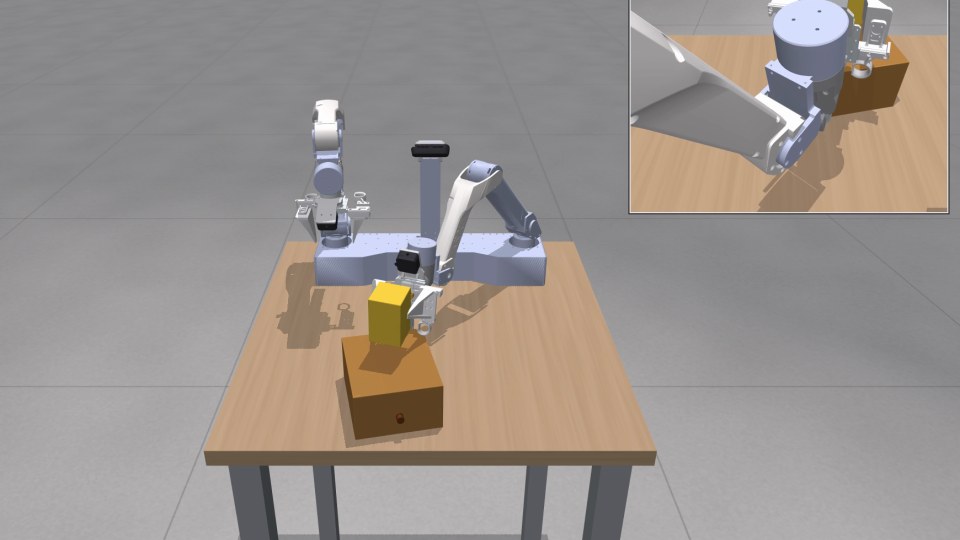} \\
\rl{0.91in}{Unitree G1} & \includegraphics[width=\tw]{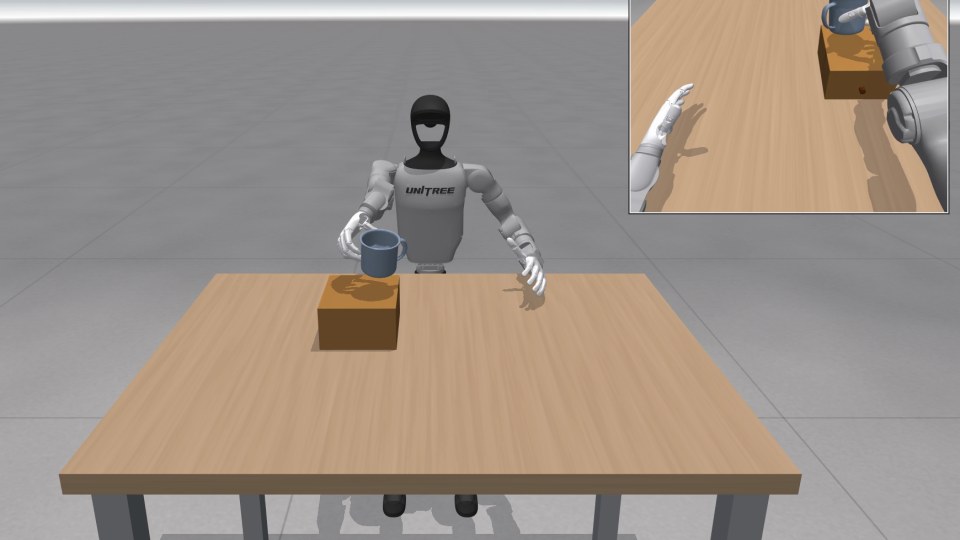} & \includegraphics[width=\tw]{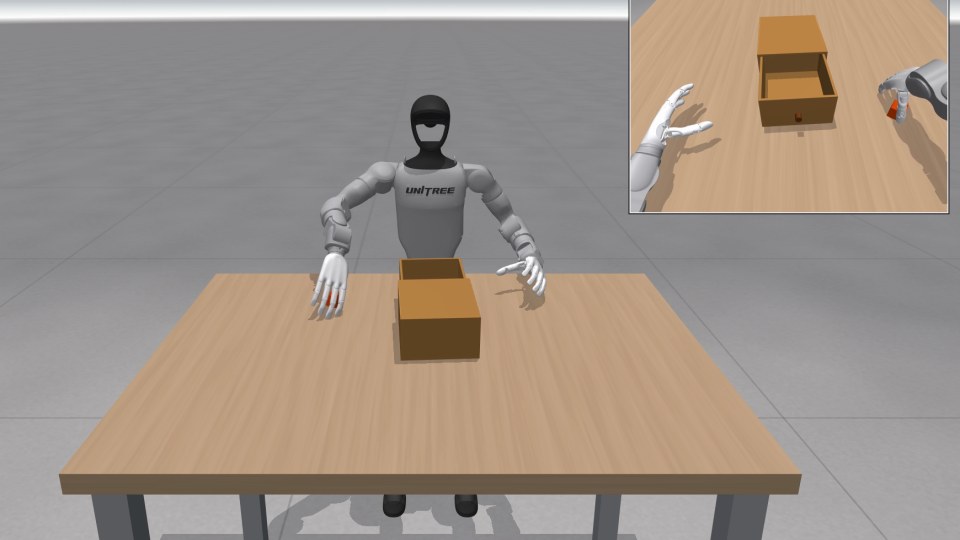} & \includegraphics[width=\tw]{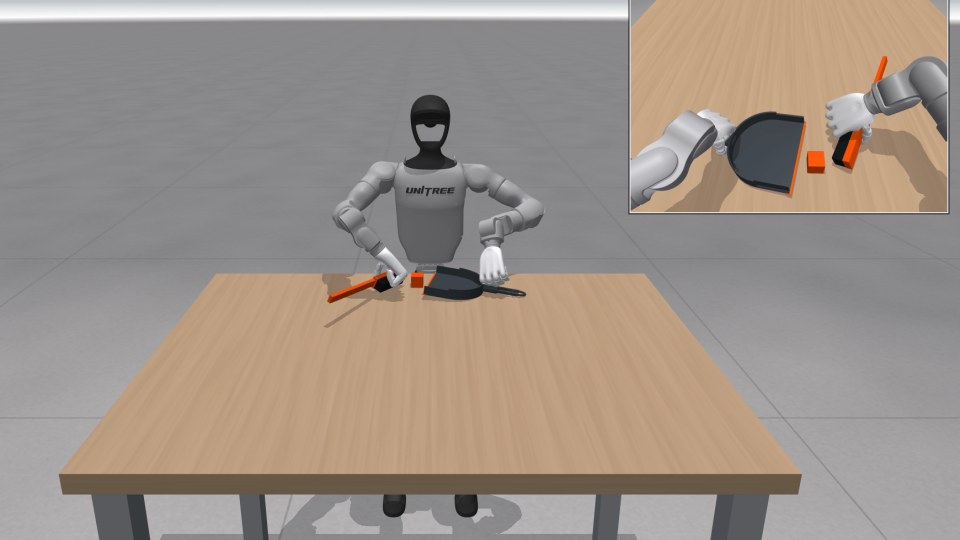} & \includegraphics[width=\tw]{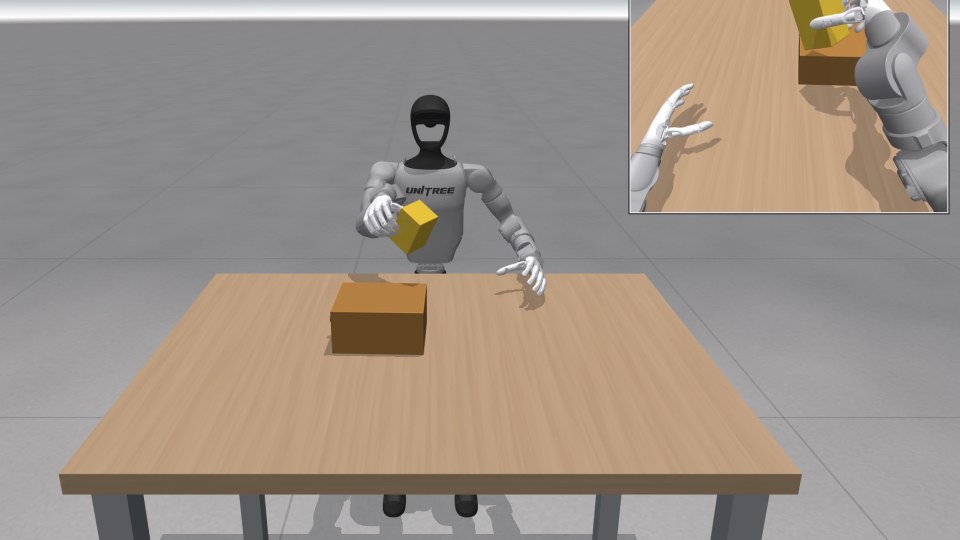} \\
\end{tabular}
\caption{\textbf{Simulation execution, four robots on eight tasks.} One key frame of each robot's final program in the rebuilt world of each task; the inset at the top right shows the robot's own camera view. Dual-arm Franka, ARX LIFT2, and Unitree G1 received no demonstration.}
\label{apx:fig:sim}
\end{figure*}

\section{Real-robot execution}
\label{apx:sec:real-exec}

The final programs run on the real robots as ordinary code, with no code edits, no human intervention, and no LLM call (Sec.~3.6). Figs.~\ref{apx:fig:real-xtrainer}--\ref{apx:fig:real-g1} show, for each robot and task, three frames of one trial: the start, a key moment, and the final state. xTrainer, Dual-arm Franka, and ARX LIFT2 ran all eight tasks on the real robot, and Unitree G1 ran four of them (Apple, Cubes, Stack Cups, and Mug-Box), so \cref{apx:fig:real-g1} has two rows. \cref{apx:fig:layouts} shows one program of Dual-arm Franka run from three different object layouts on each of three tasks.

\newcommand{\rw}{0.150\textwidth}
\newcommand{\rlab}[1]{\rotatebox{90}{\makebox[0.58in]{\scriptsize #1}}}

\newcommand{\rhead}{ & \scriptsize Start & \scriptsize Key moment & \scriptsize Final state & & & \scriptsize Start & \scriptsize Key moment & \scriptsize Final state \\[1pt]}
\newcommand{\rtab}{\setlength{\tabcolsep}{1pt}}

\begin{figure*}[p]
\centering
\setcounter{figure}{5}\renewcommand{\thefigure}{S6a}\renewcommand{\theHfigure}{S6a}
\rtab
\begin{tabular}{@{}c@{\hspace{2pt}}ccc@{\hspace{6pt}}c@{\hspace{0pt}}c@{\hspace{2pt}}ccc@{}}
\rhead
\rlab{Apple} & \includegraphics[width=\rw]{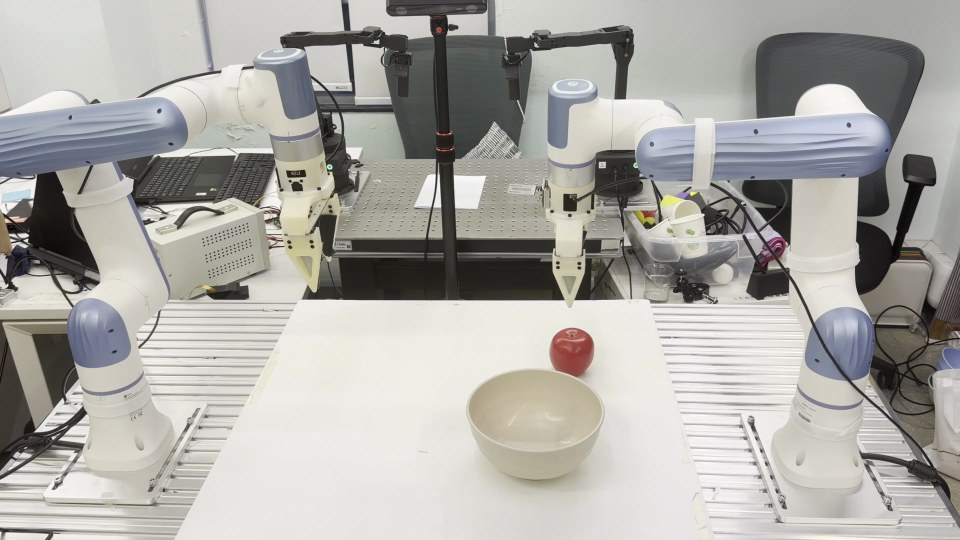} & \includegraphics[width=\rw]{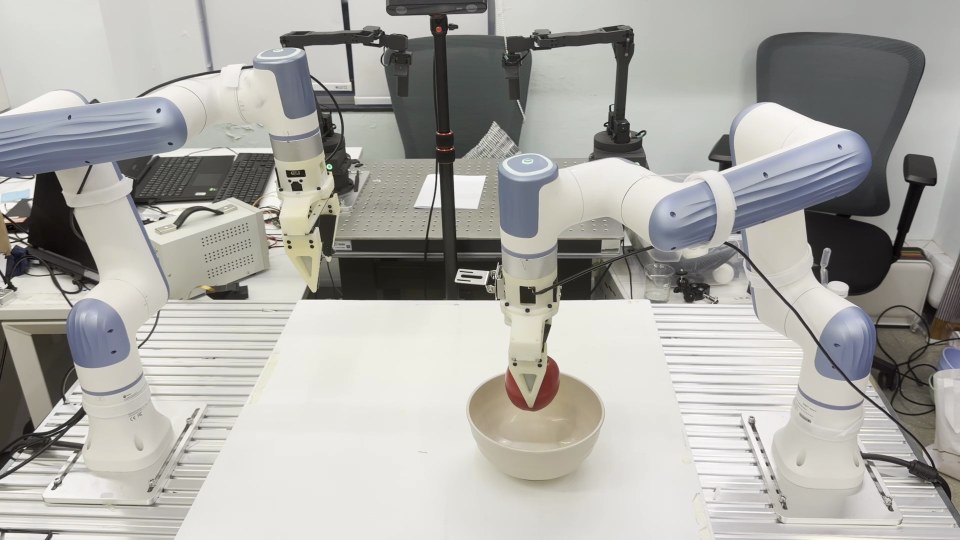} & \includegraphics[width=\rw]{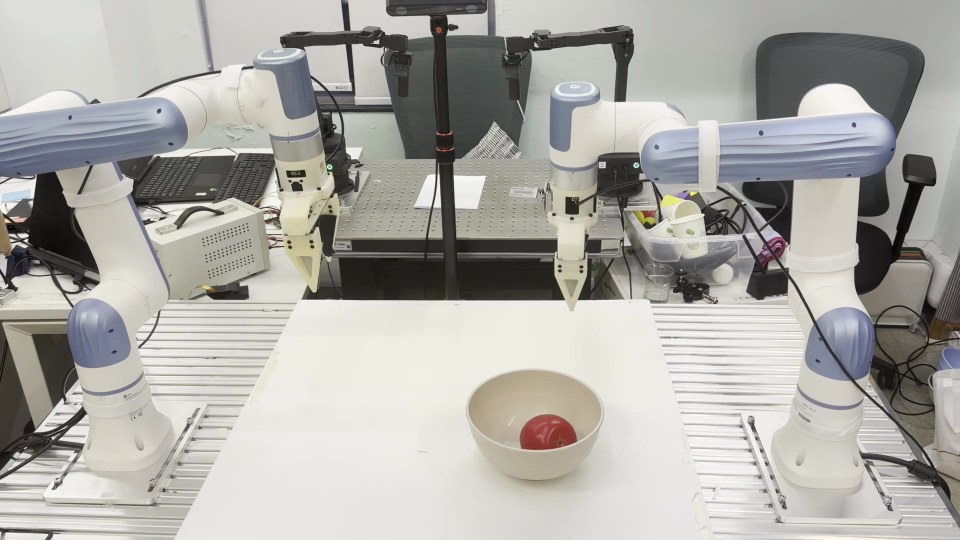} & & \rlab{Cubes} & \includegraphics[width=\rw]{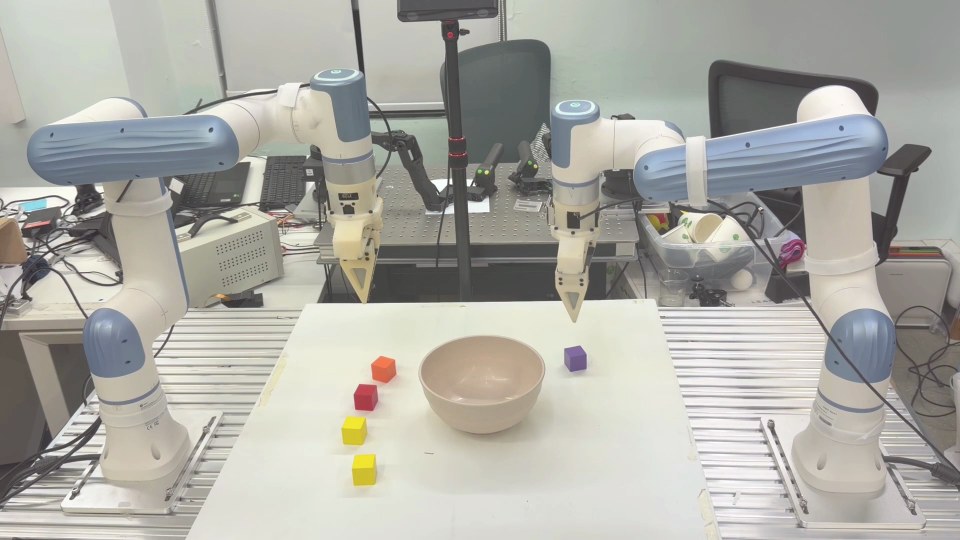} & \includegraphics[width=\rw]{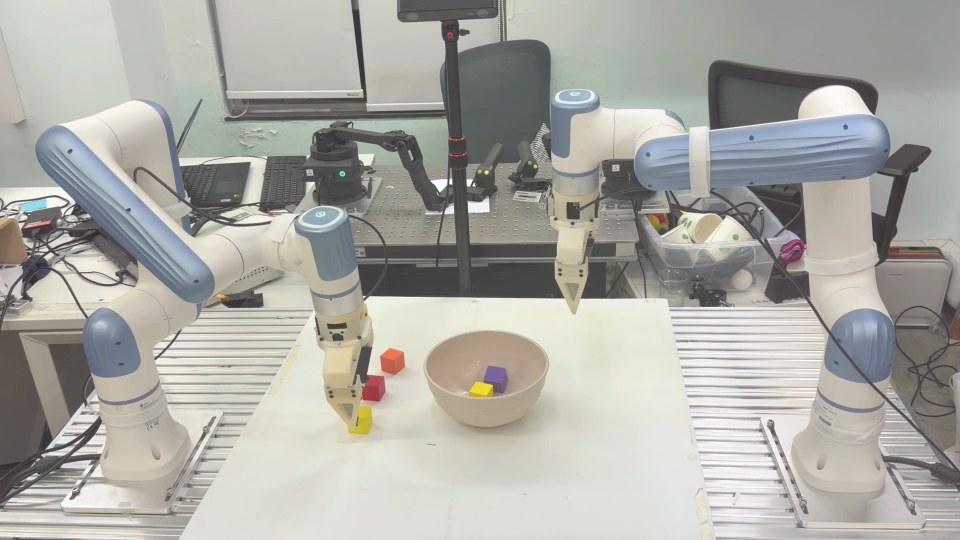} & \includegraphics[width=\rw]{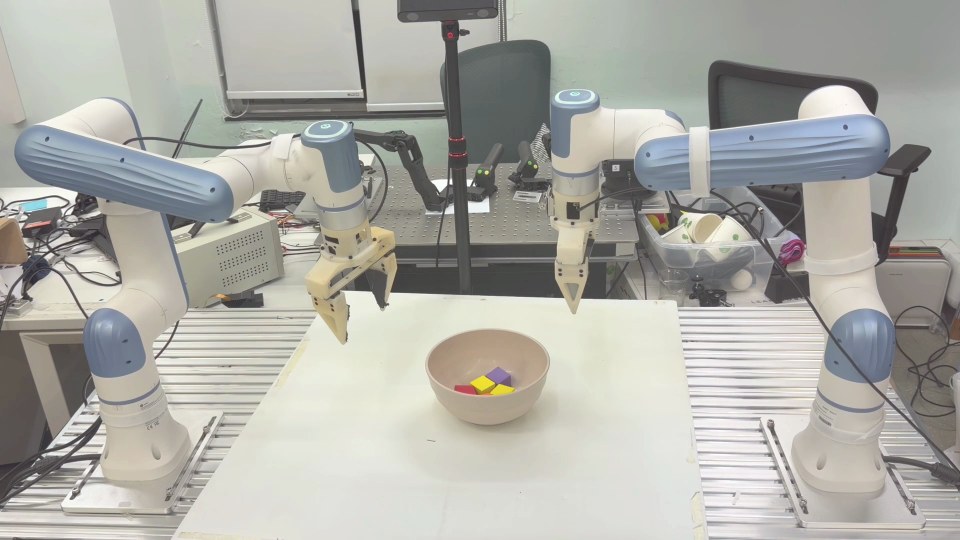} \\[1pt]
\rlab{Stack Cups} & \includegraphics[width=\rw]{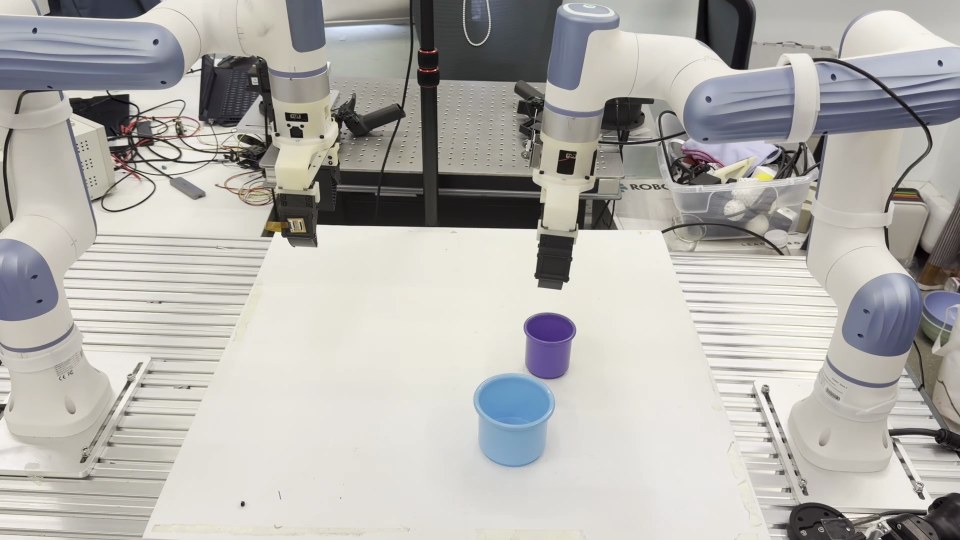} & \includegraphics[width=\rw]{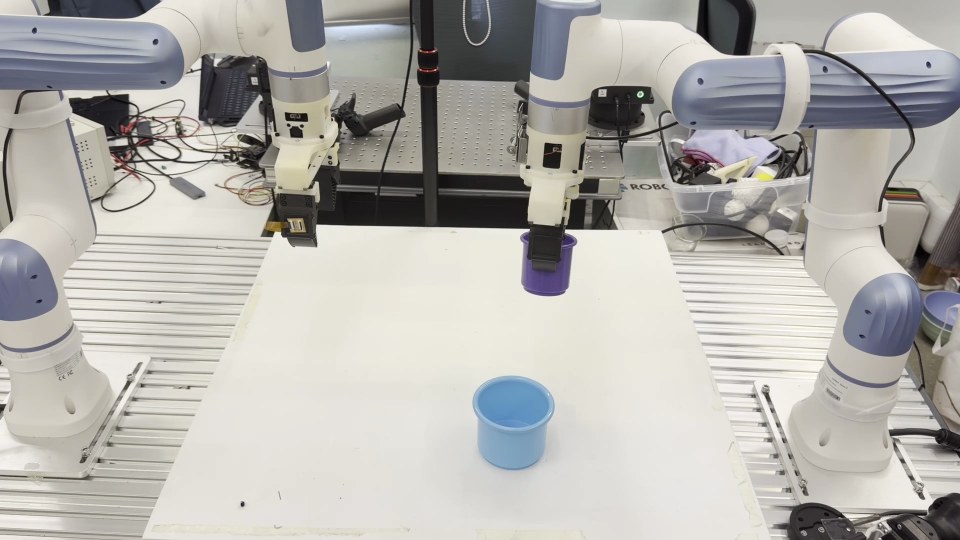} & \includegraphics[width=\rw]{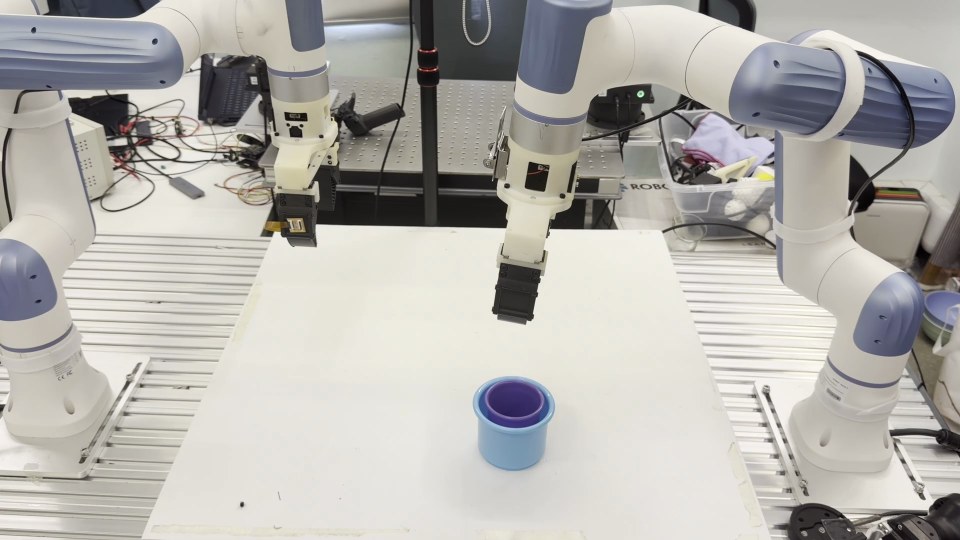} & & \rlab{Build Tower} & \includegraphics[width=\rw]{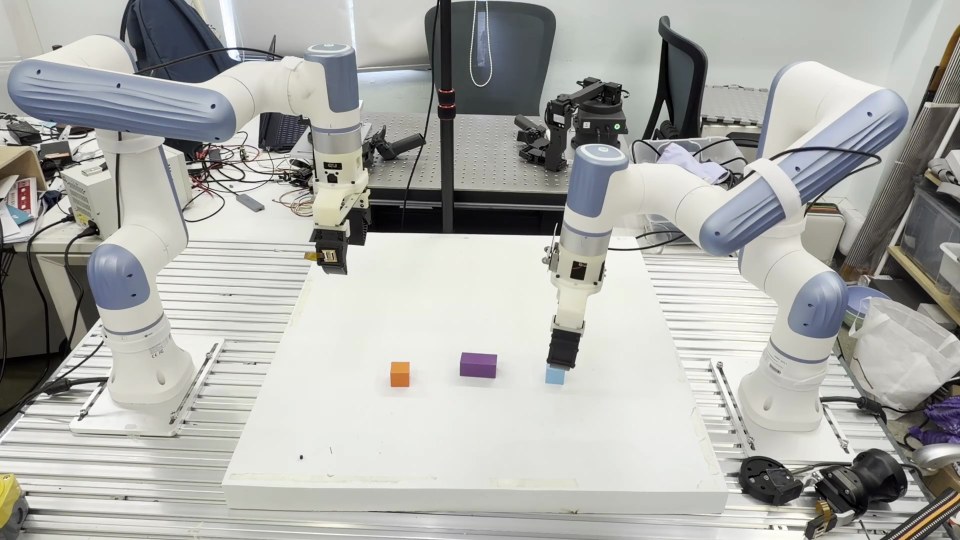} & \includegraphics[width=\rw]{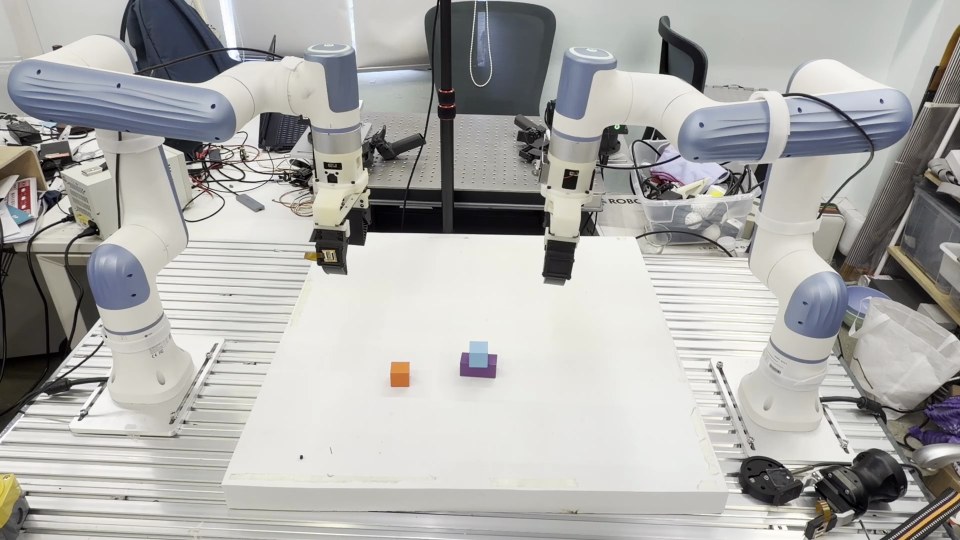} & \includegraphics[width=\rw]{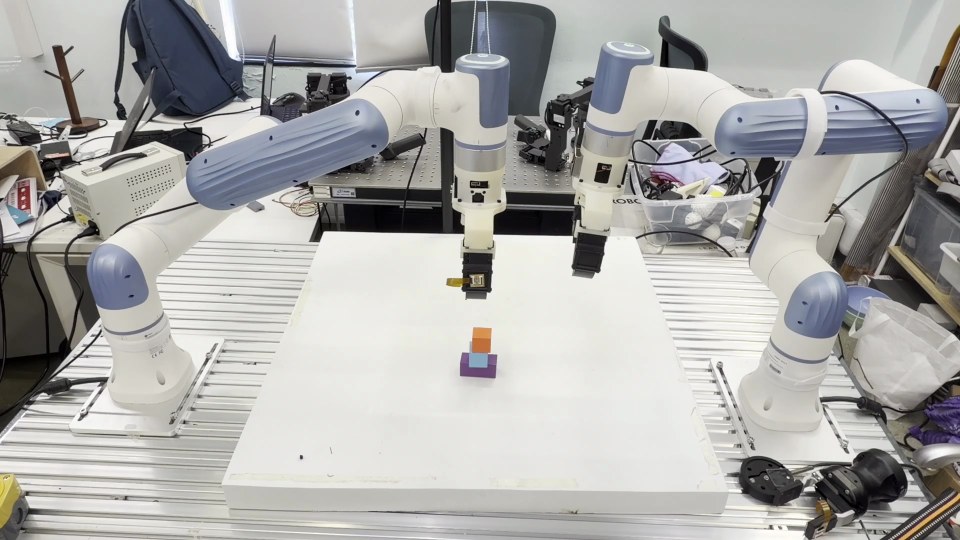} \\[1pt]
\rlab{Mug-Box} & \includegraphics[width=\rw]{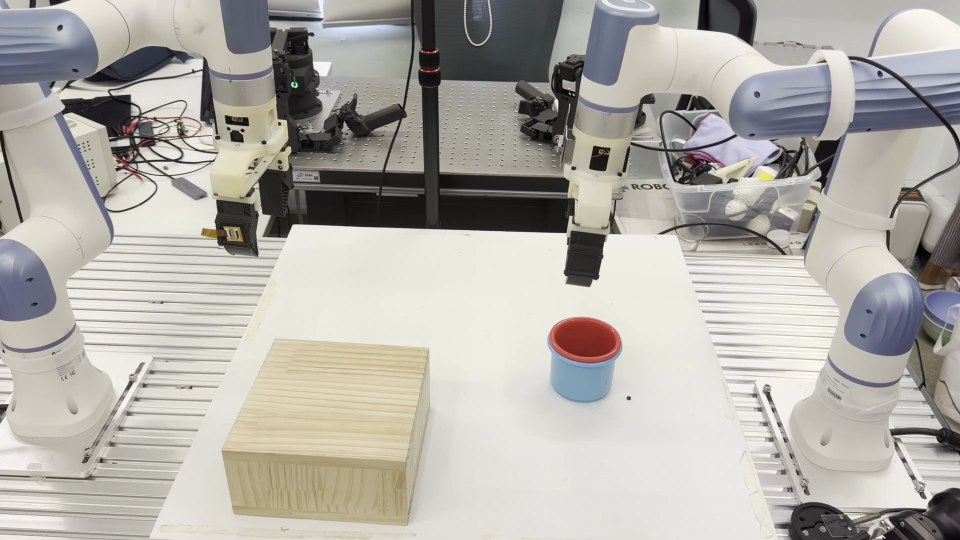} & \includegraphics[width=\rw]{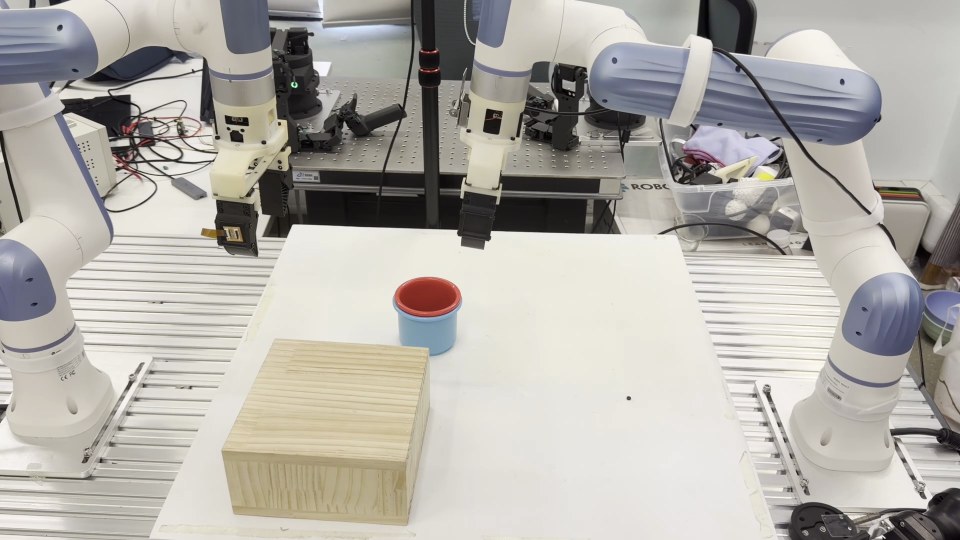} & \includegraphics[width=\rw]{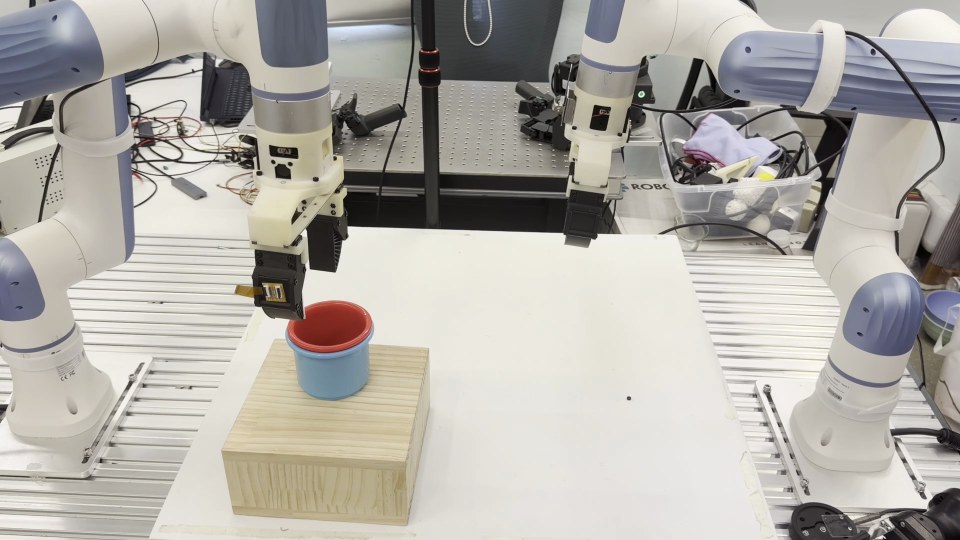} & & \rlab{Open Drawer} & \includegraphics[width=\rw]{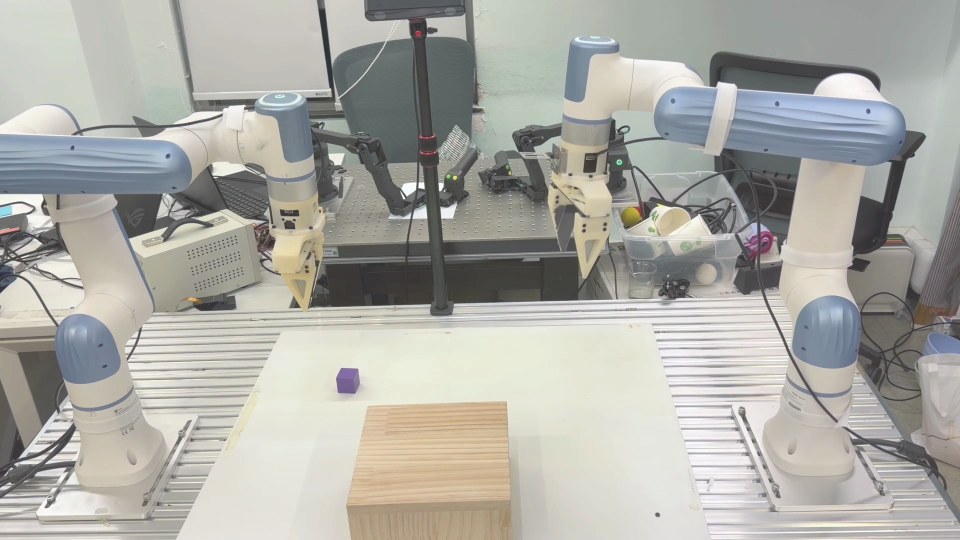} & \includegraphics[width=\rw]{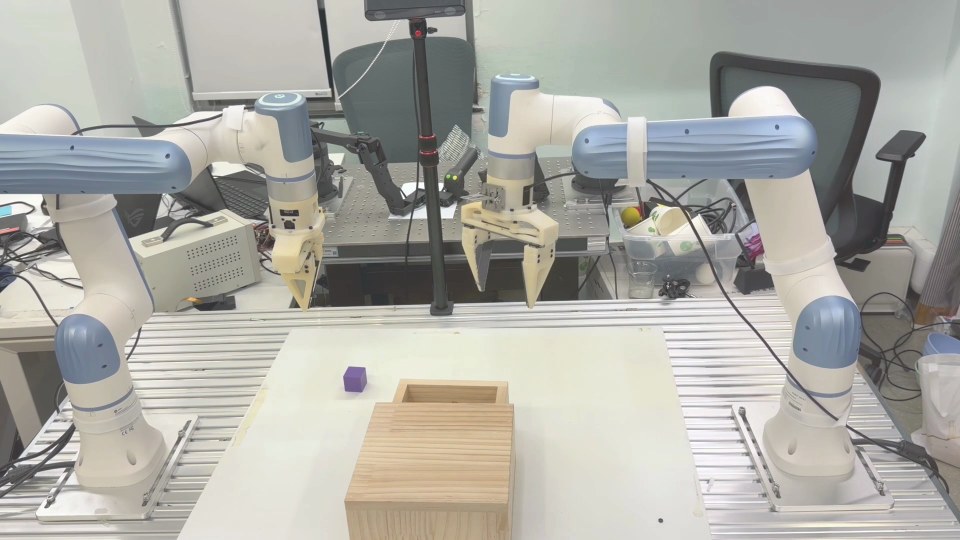} & \includegraphics[width=\rw]{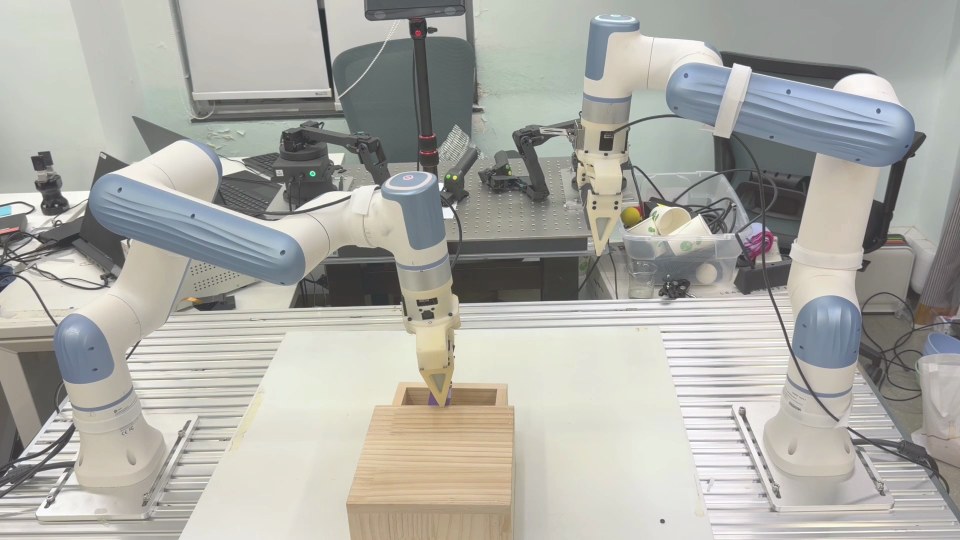} \\[1pt]
\rlab{Sweep} & \includegraphics[width=\rw]{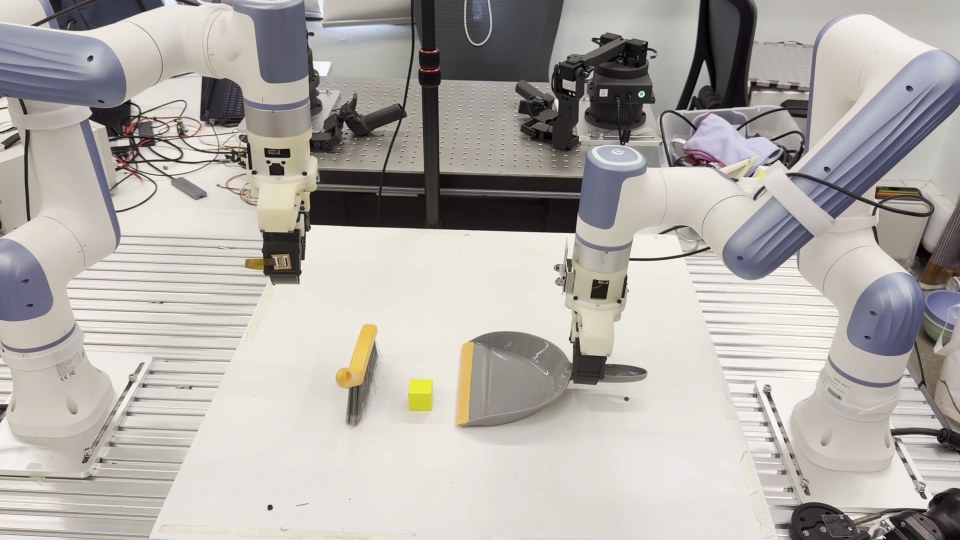} & \includegraphics[width=\rw]{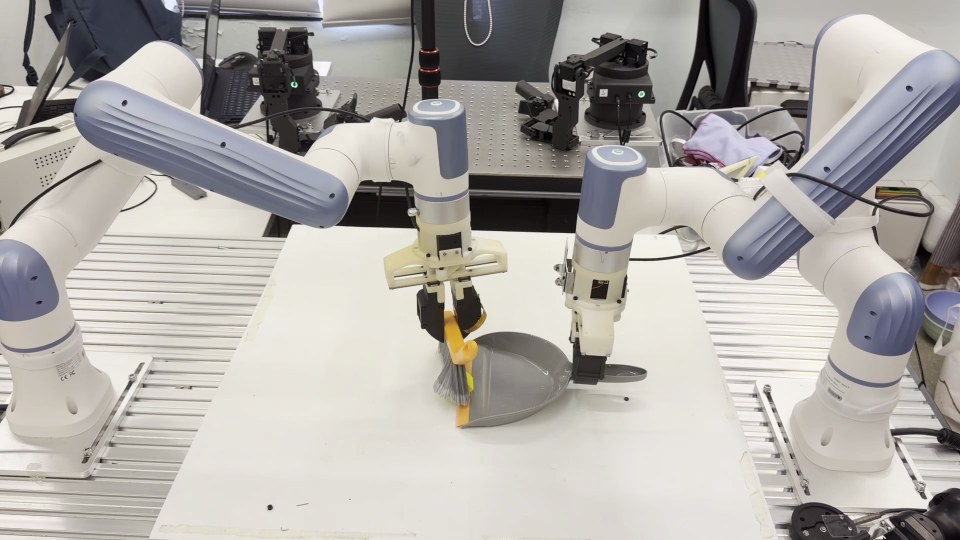} & \includegraphics[width=\rw]{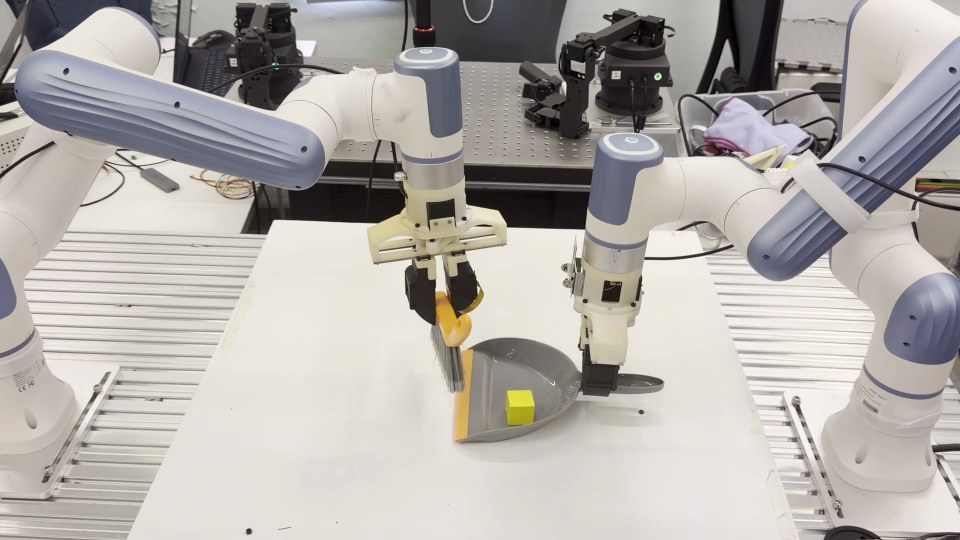} & & \rlab{Lift-Box} & \includegraphics[width=\rw]{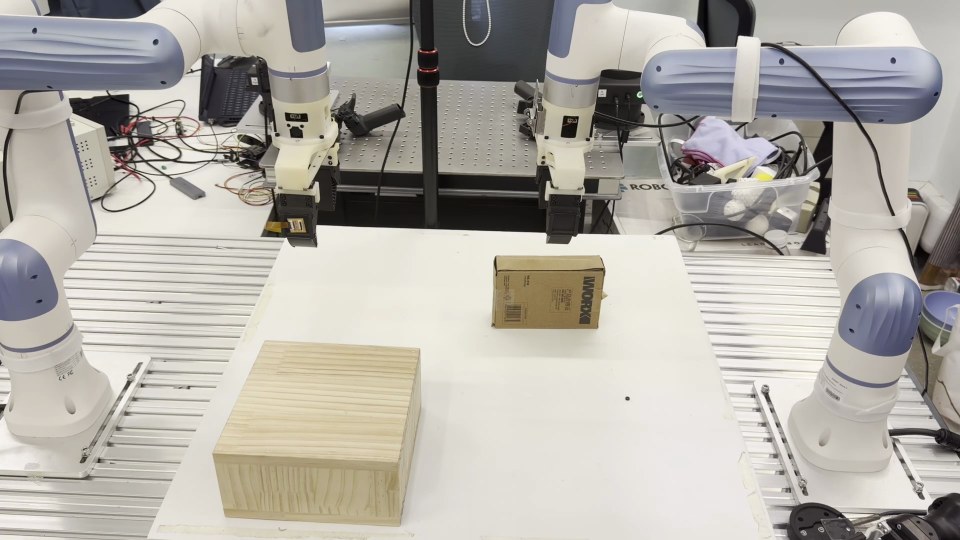} & \includegraphics[width=\rw]{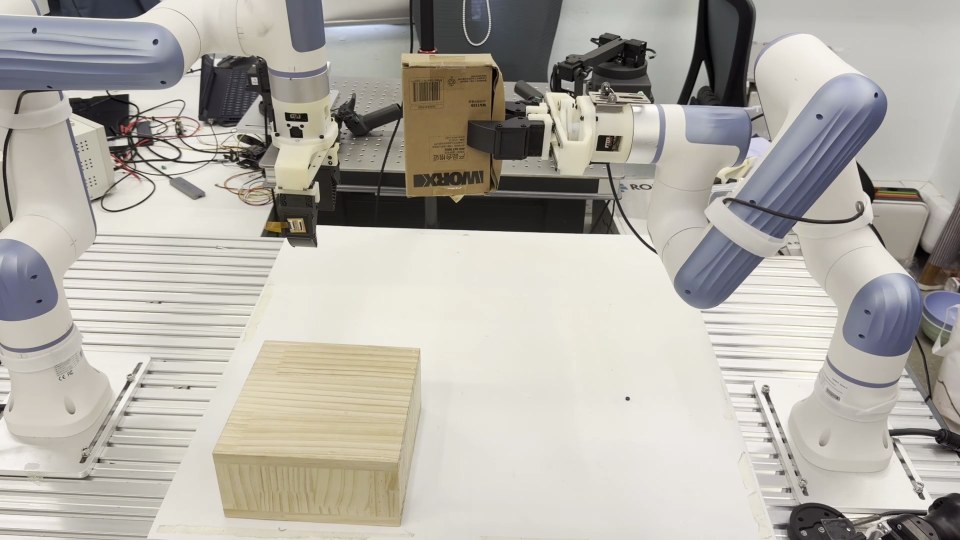} & \includegraphics[width=\rw]{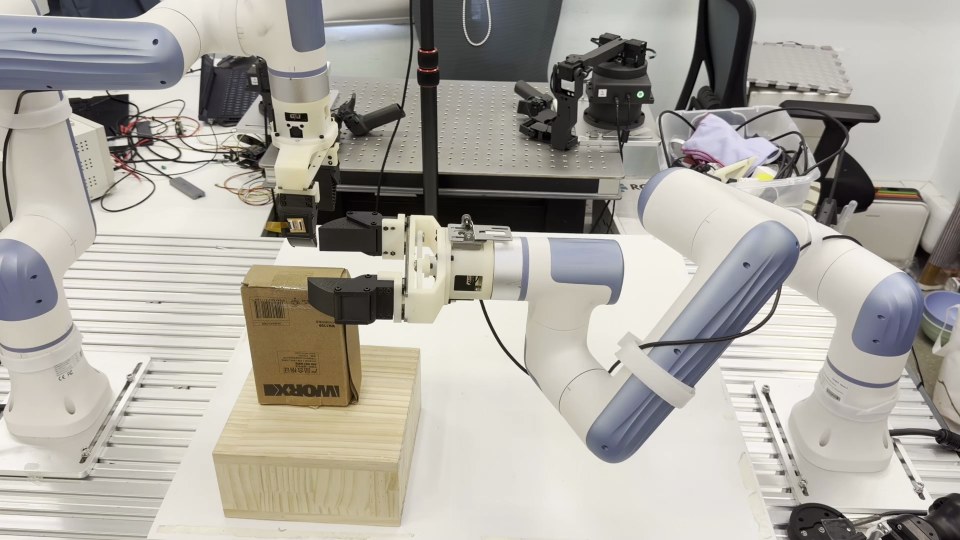} \\[1pt]
\end{tabular}
\caption{\textbf{Real-robot execution on xTrainer (source robot).} Start, key moment, and final state of one trial per task.}
\label{apx:fig:real-xtrainer}
\vspace{14pt}
\setcounter{figure}{5}\renewcommand{\thefigure}{S6b}\renewcommand{\theHfigure}{S6b}
\rtab
\begin{tabular}{@{}c@{\hspace{2pt}}ccc@{\hspace{6pt}}c@{\hspace{0pt}}c@{\hspace{2pt}}ccc@{}}
\rhead
\rlab{Apple} & \includegraphics[width=\rw]{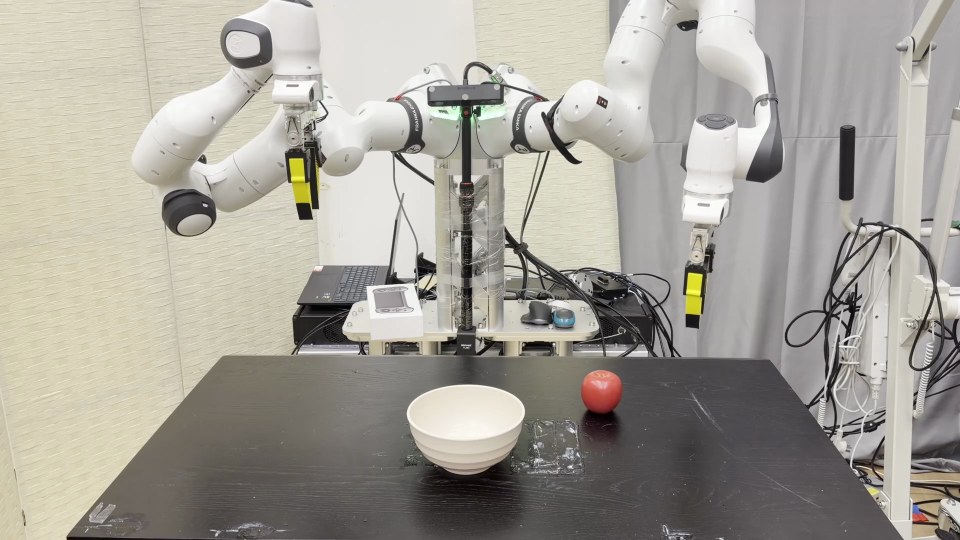} & \includegraphics[width=\rw]{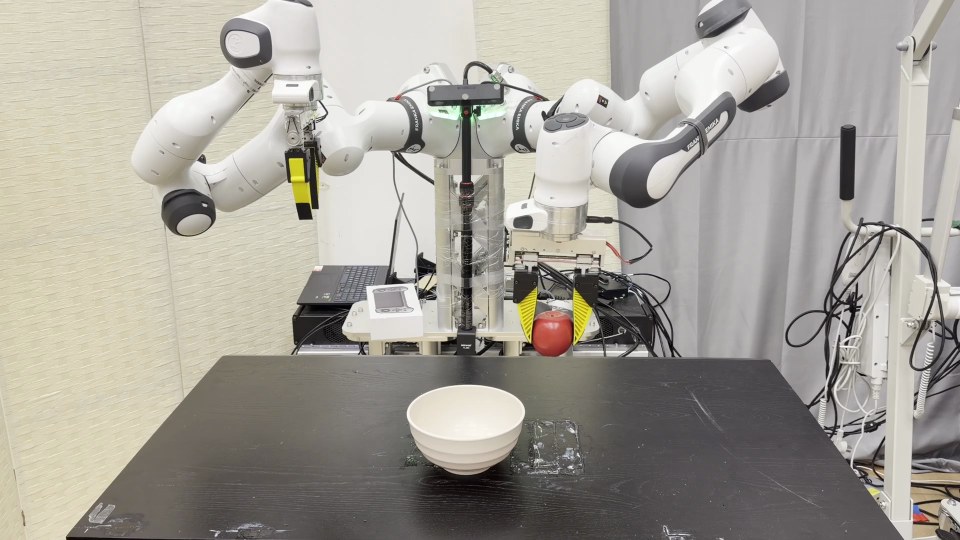} & \includegraphics[width=\rw]{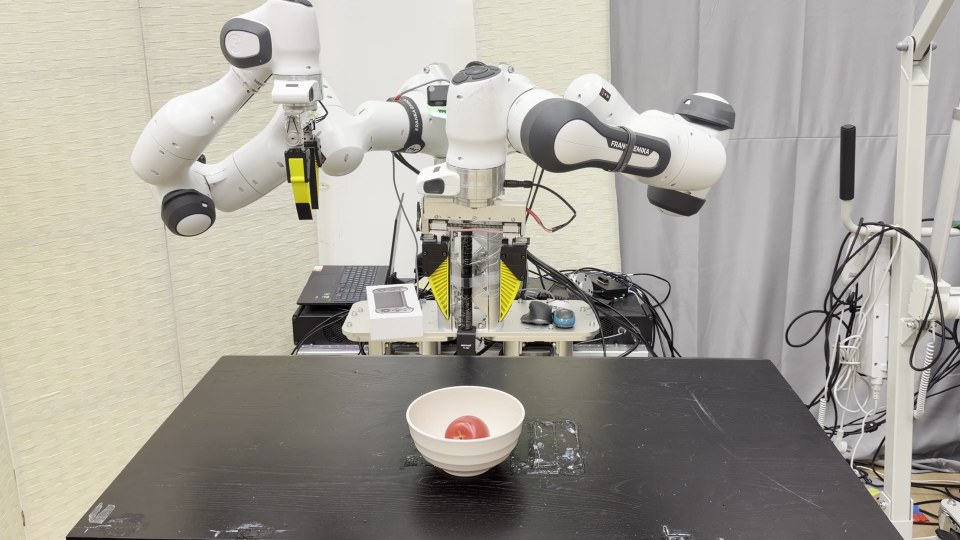} & & \rlab{Cubes} & \includegraphics[width=\rw]{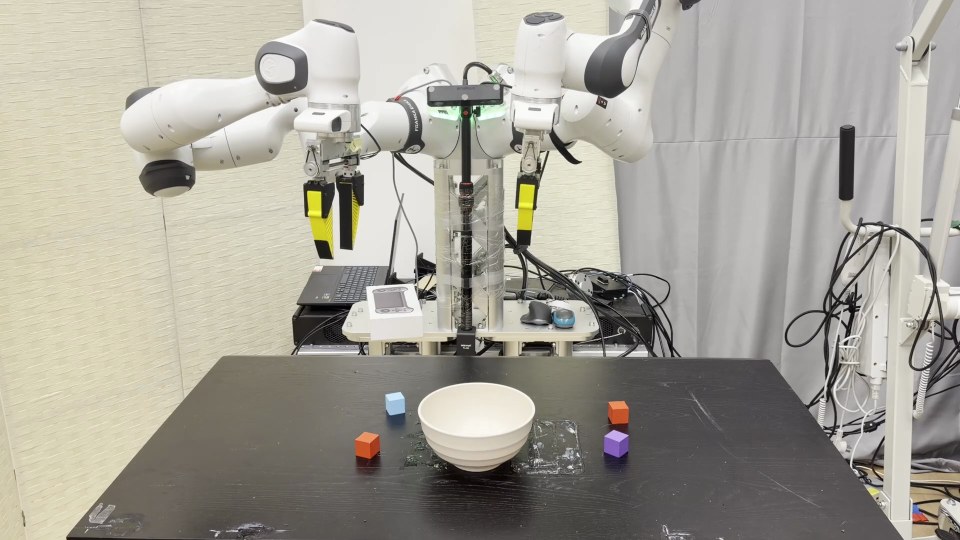} & \includegraphics[width=\rw]{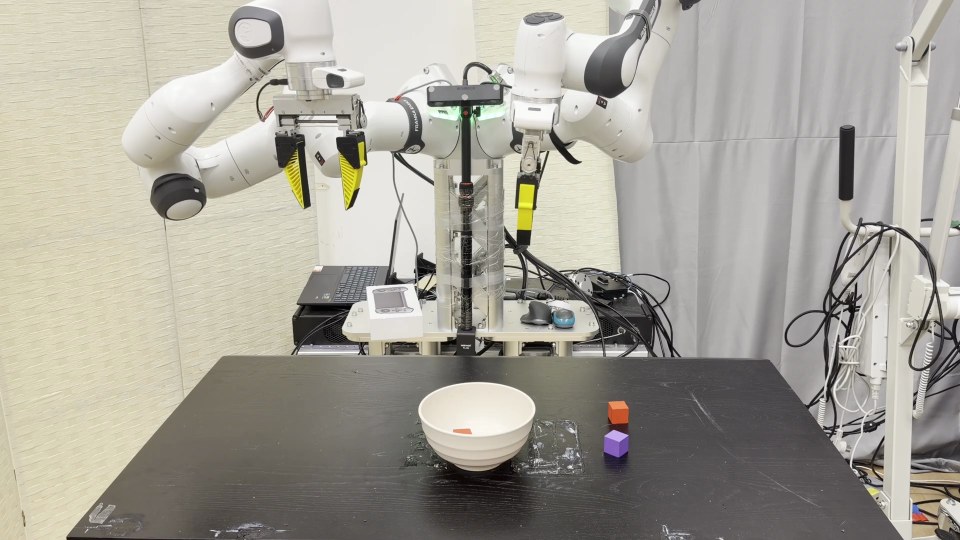} & \includegraphics[width=\rw]{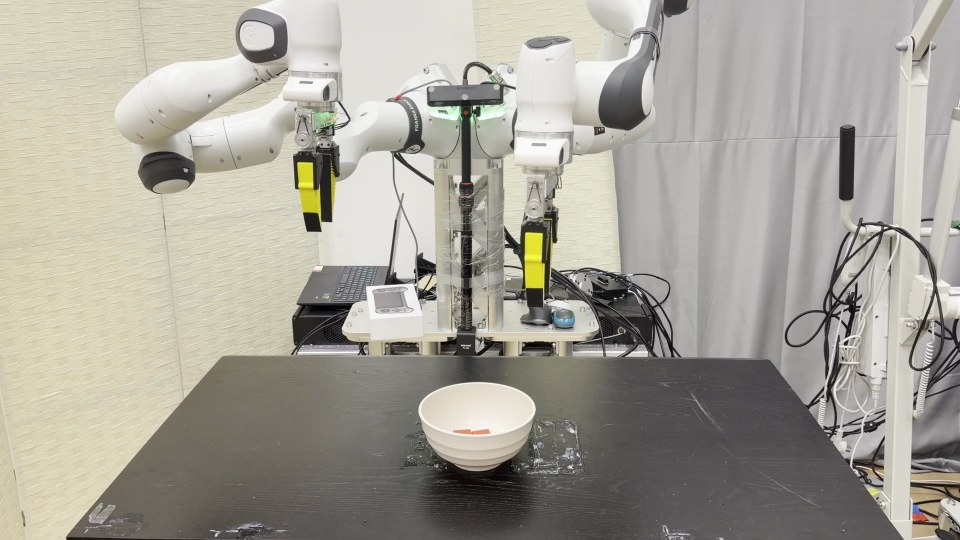} \\[1pt]
\rlab{Stack Cups} & \includegraphics[width=\rw]{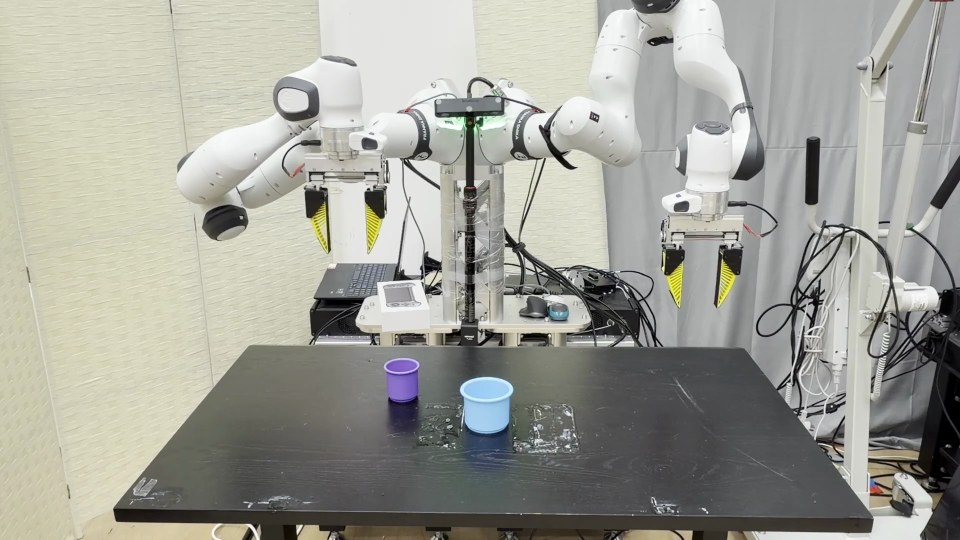} & \includegraphics[width=\rw]{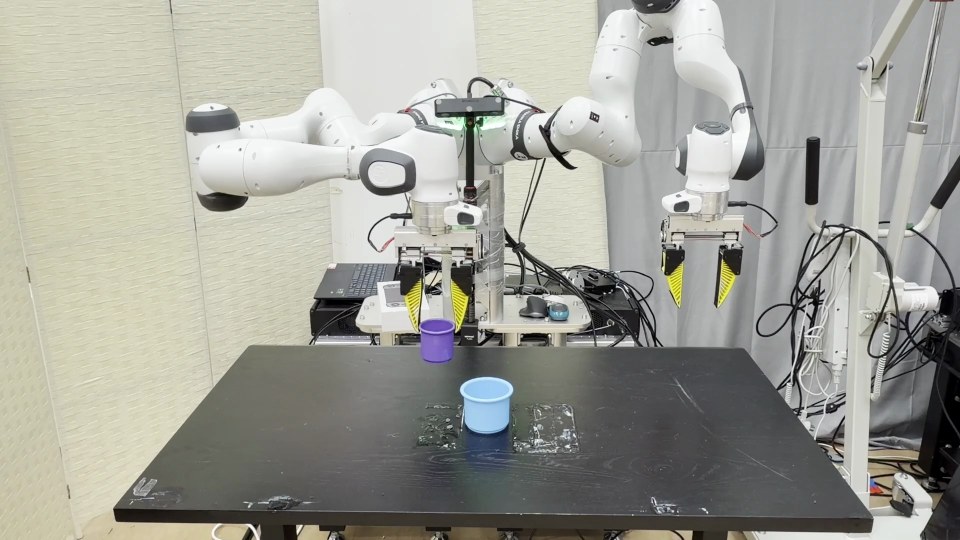} & \includegraphics[width=\rw]{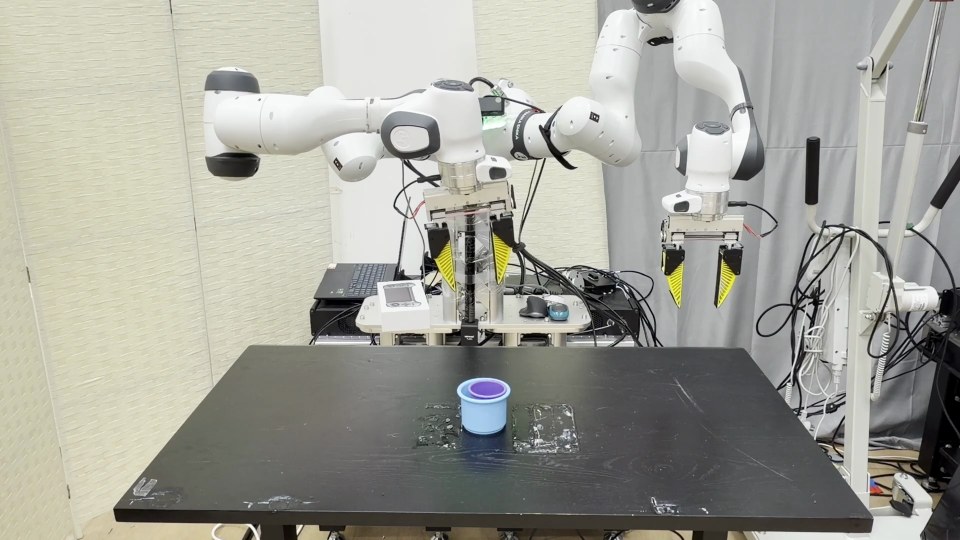} & & \rlab{Build Tower} & \includegraphics[width=\rw]{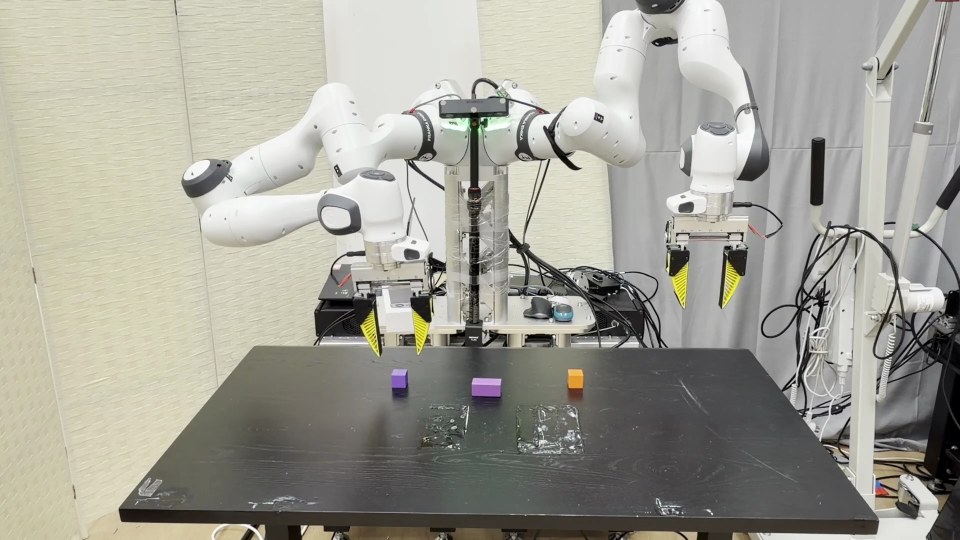} & \includegraphics[width=\rw]{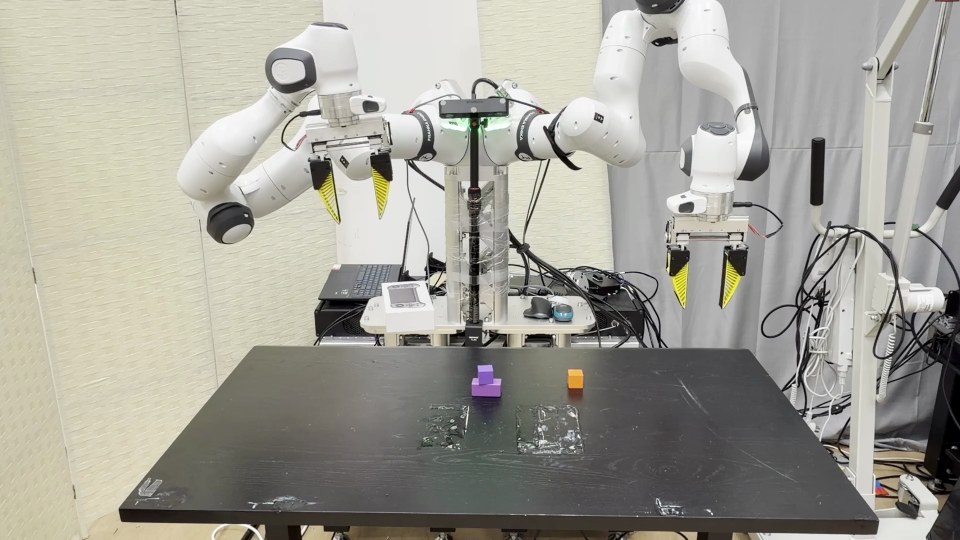} & \includegraphics[width=\rw]{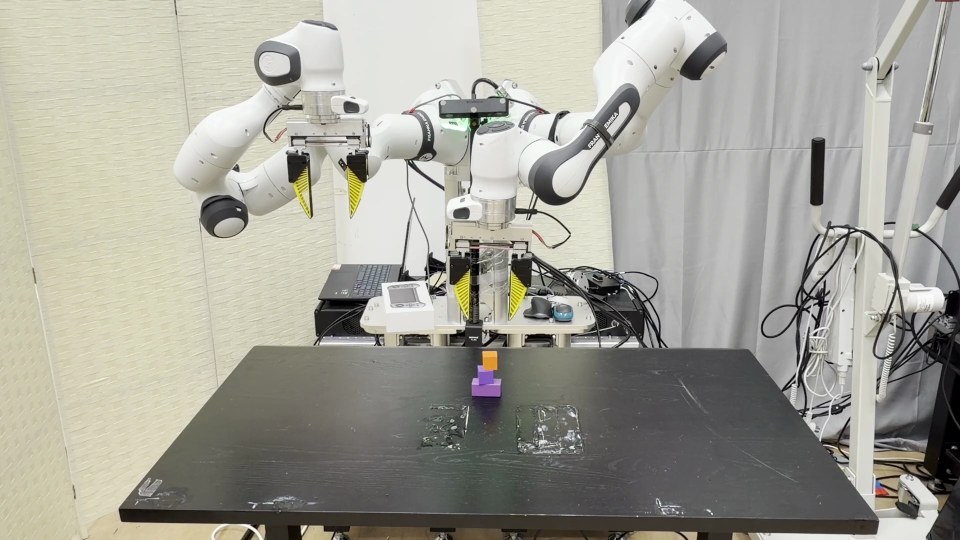} \\[1pt]
\rlab{Mug-Box} & \includegraphics[width=\rw]{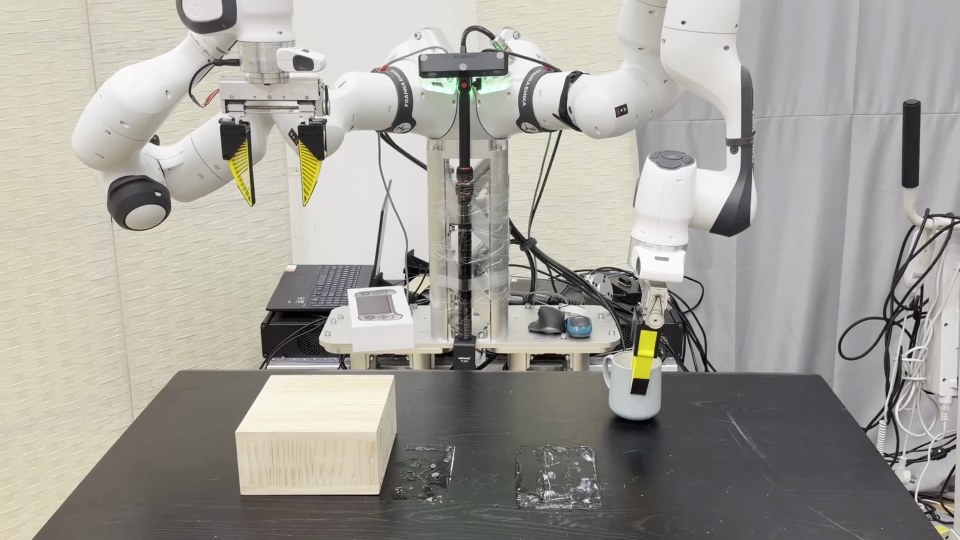} & \includegraphics[width=\rw]{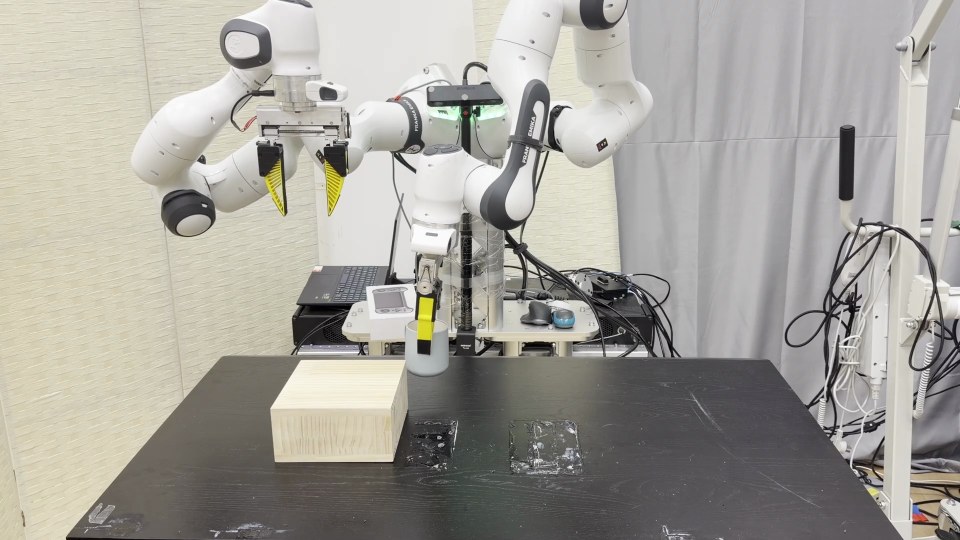} & \includegraphics[width=\rw]{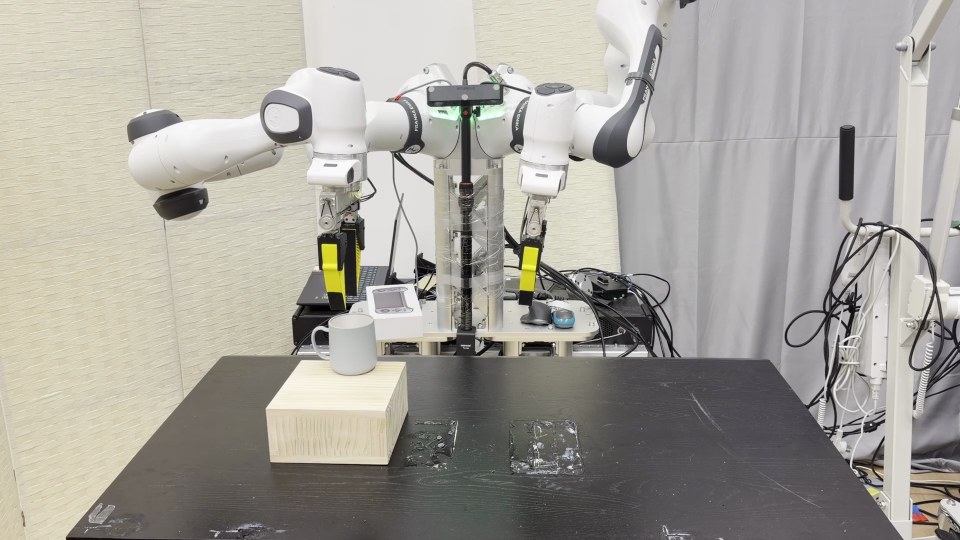} & & \rlab{Open Drawer} & \includegraphics[width=\rw]{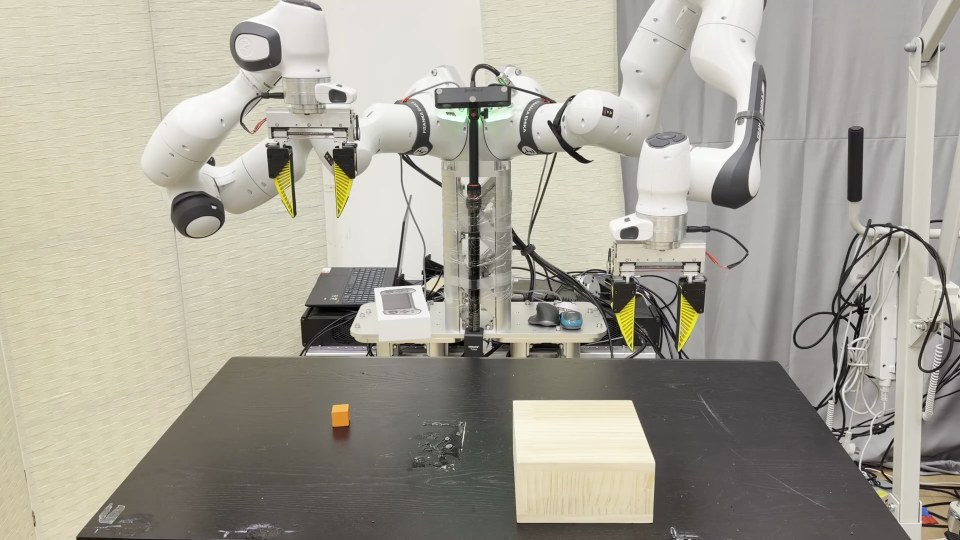} & \includegraphics[width=\rw]{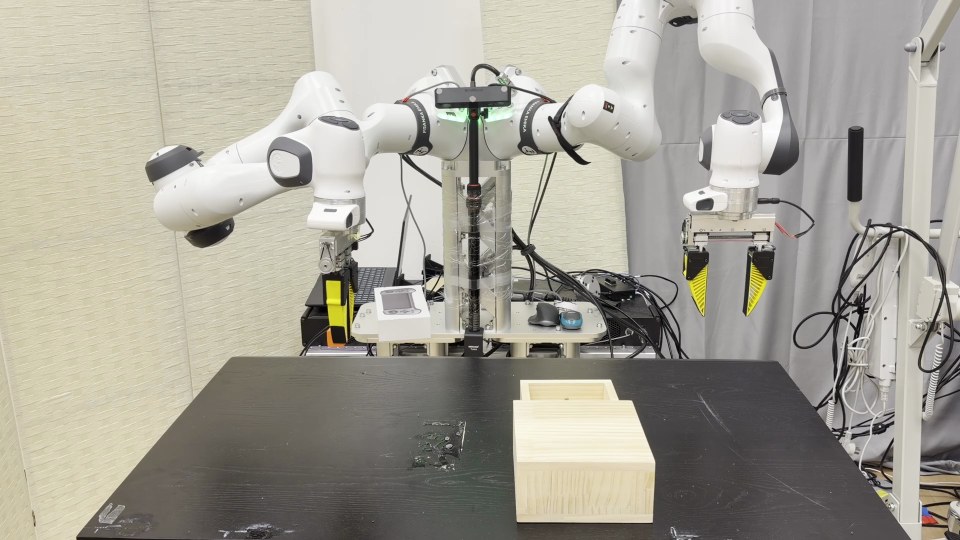} & \includegraphics[width=\rw]{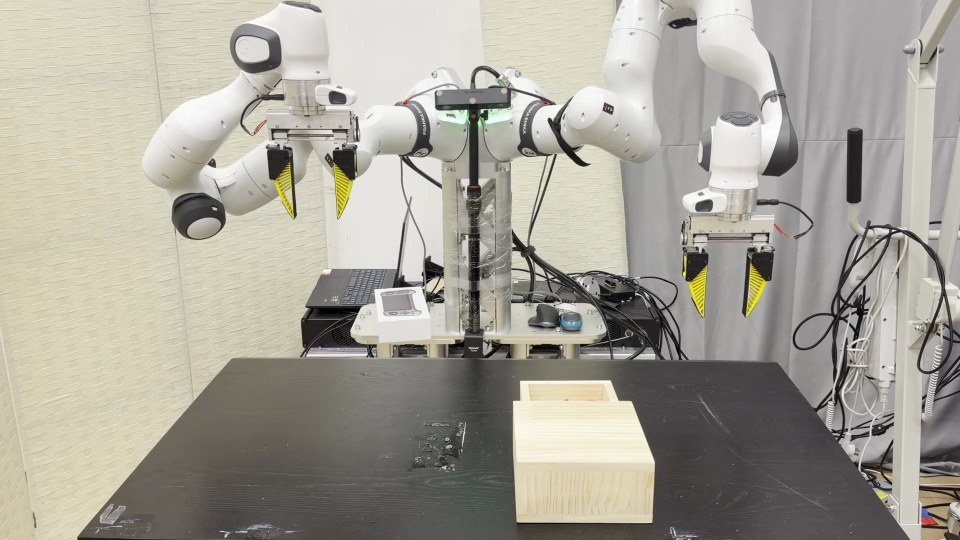} \\[1pt]
\rlab{Sweep} & \includegraphics[width=\rw]{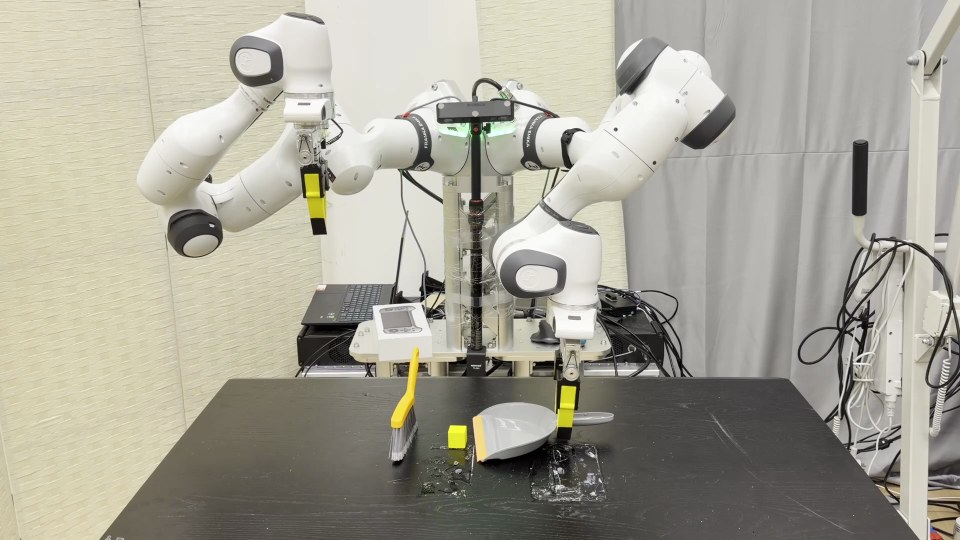} & \includegraphics[width=\rw]{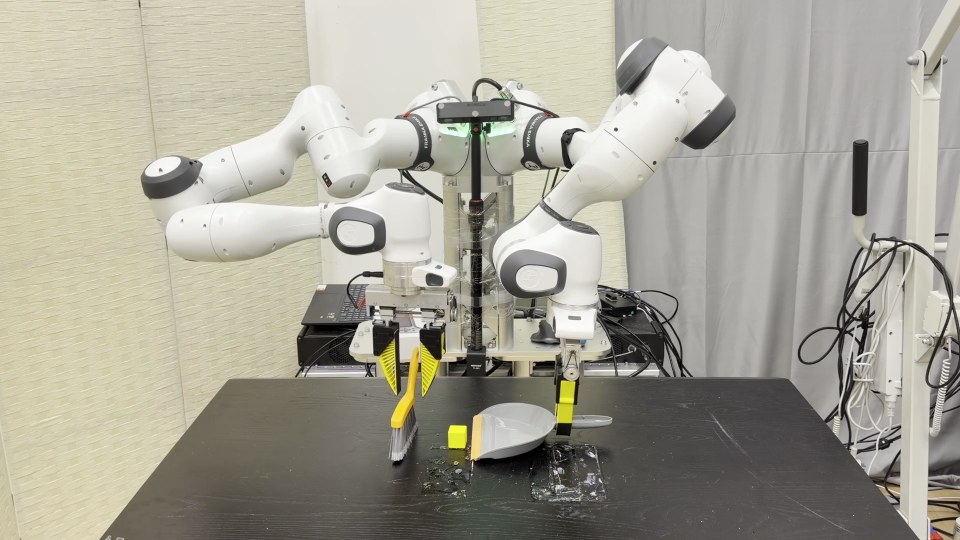} & \includegraphics[width=\rw]{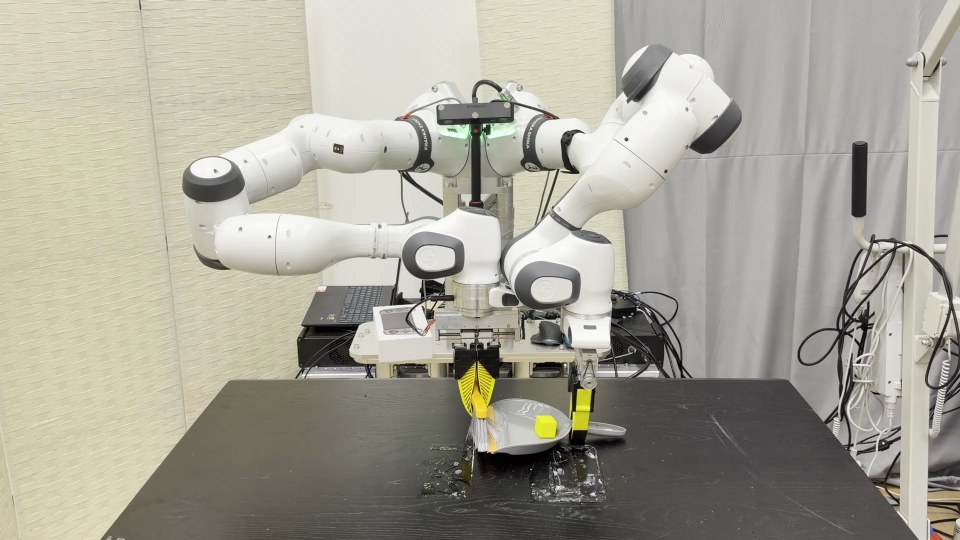} & & \rlab{Lift-Box} & \includegraphics[width=\rw]{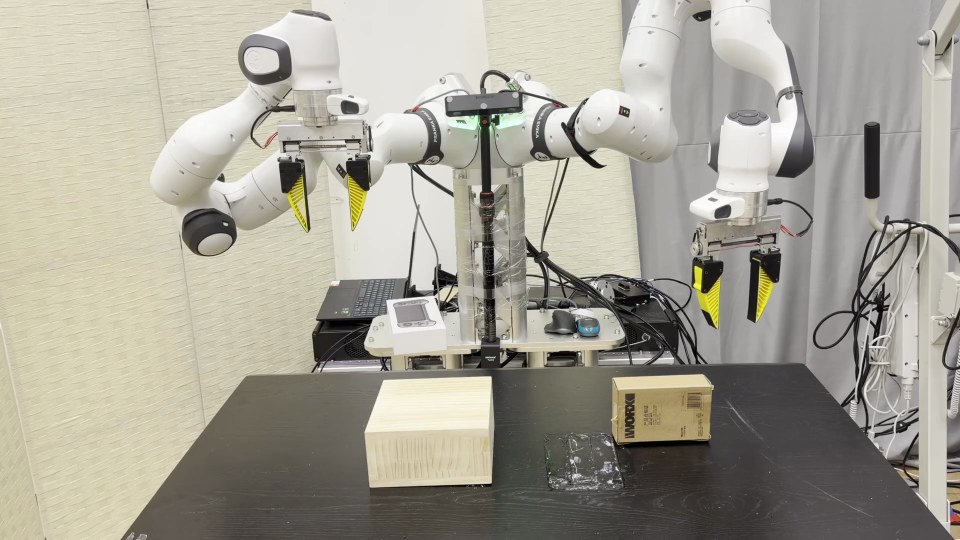} & \includegraphics[width=\rw]{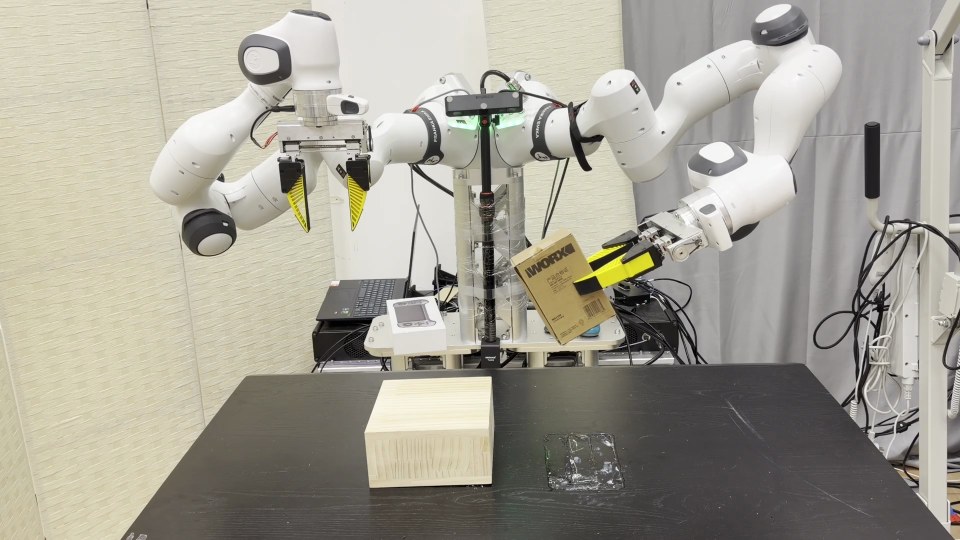} & \includegraphics[width=\rw]{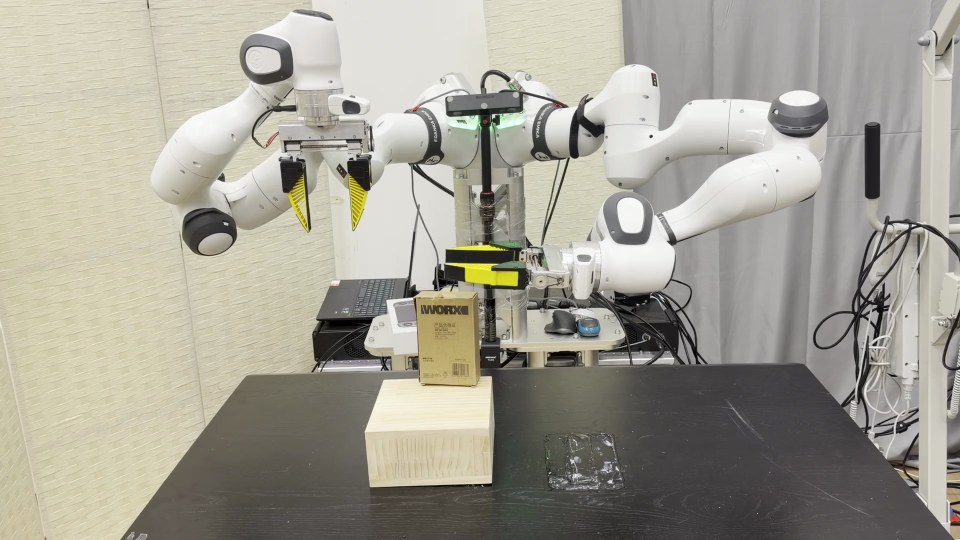} \\[1pt]
\end{tabular}
\caption{\textbf{Real-robot execution on Dual-arm Franka (no demonstration).} Start, key moment, and final state of one trial per task.}
\label{apx:fig:real-franka}
\end{figure*}

\begin{figure*}[p]
\centering
\setcounter{figure}{5}\renewcommand{\thefigure}{S6c}\renewcommand{\theHfigure}{S6c}
\rtab
\begin{tabular}{@{}c@{\hspace{2pt}}ccc@{\hspace{6pt}}c@{\hspace{0pt}}c@{\hspace{2pt}}ccc@{}}
\rhead
\rlab{Apple} & \includegraphics[width=\rw]{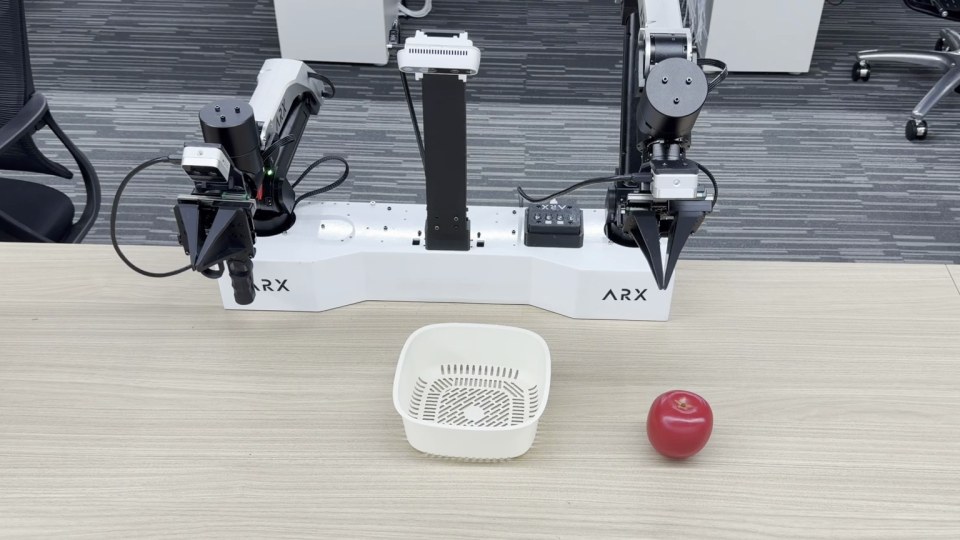} & \includegraphics[width=\rw]{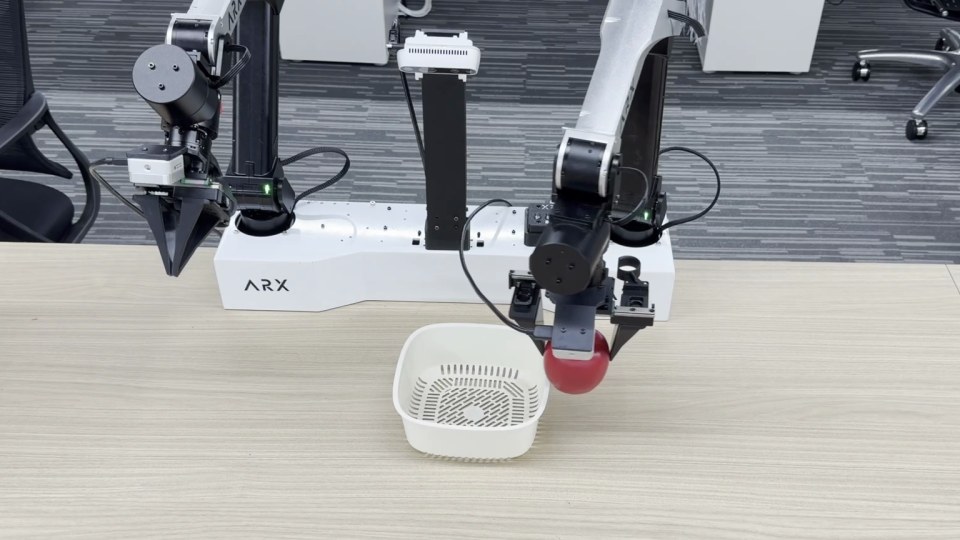} & \includegraphics[width=\rw]{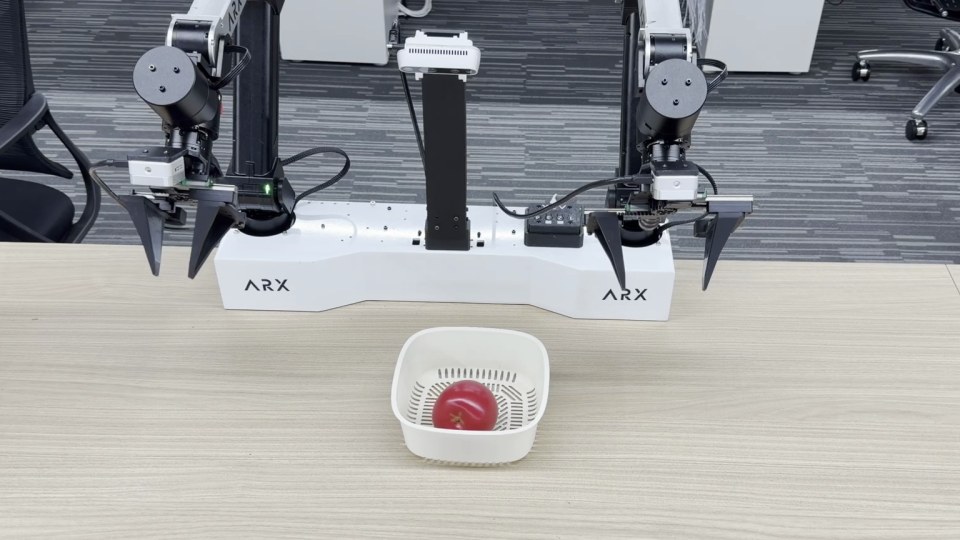} & & \rlab{Cubes} & \includegraphics[width=\rw]{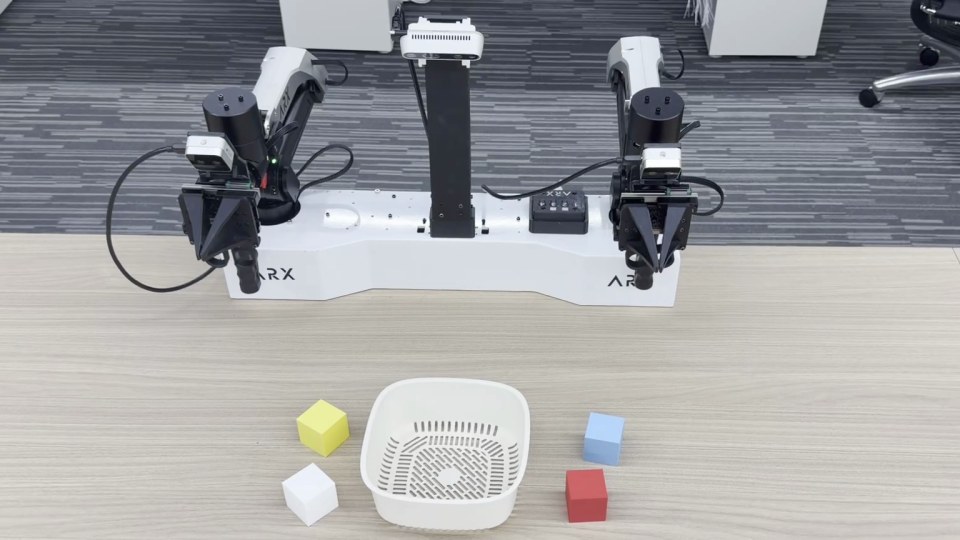} & \includegraphics[width=\rw]{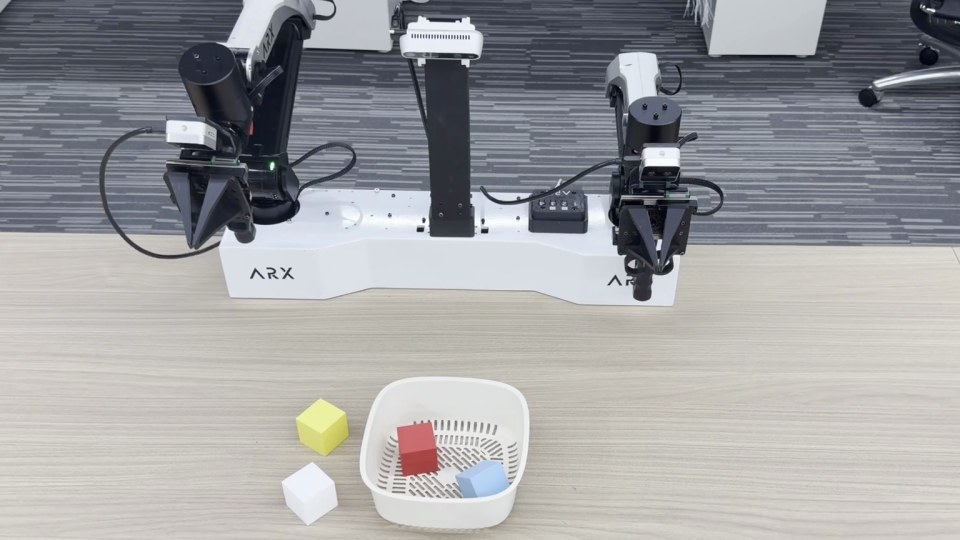} & \includegraphics[width=\rw]{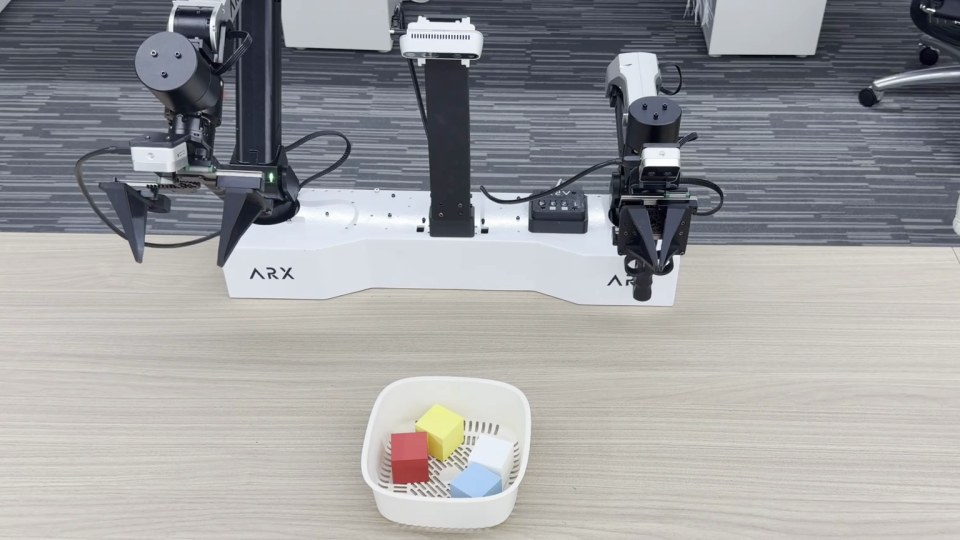} \\[1pt]
\rlab{Stack Cups} & \includegraphics[width=\rw]{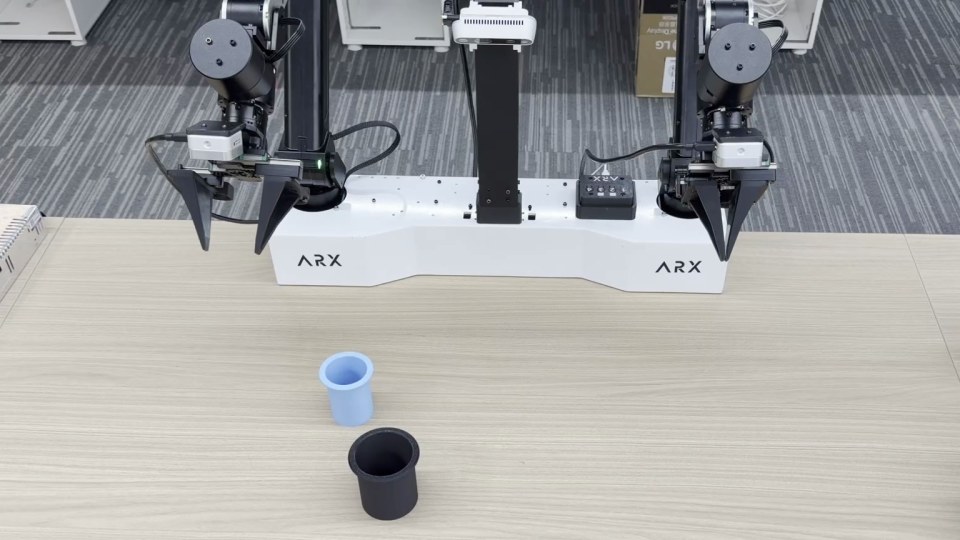} & \includegraphics[width=\rw]{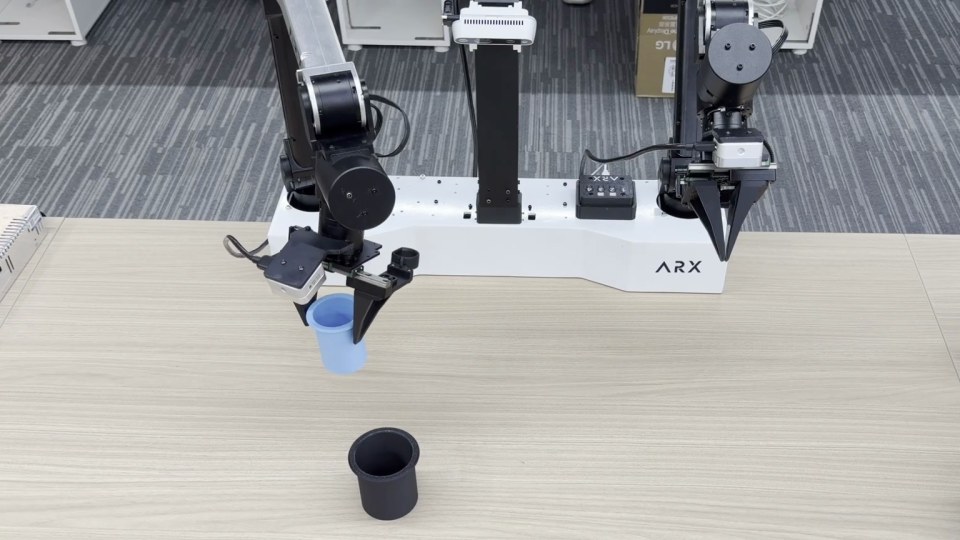} & \includegraphics[width=\rw]{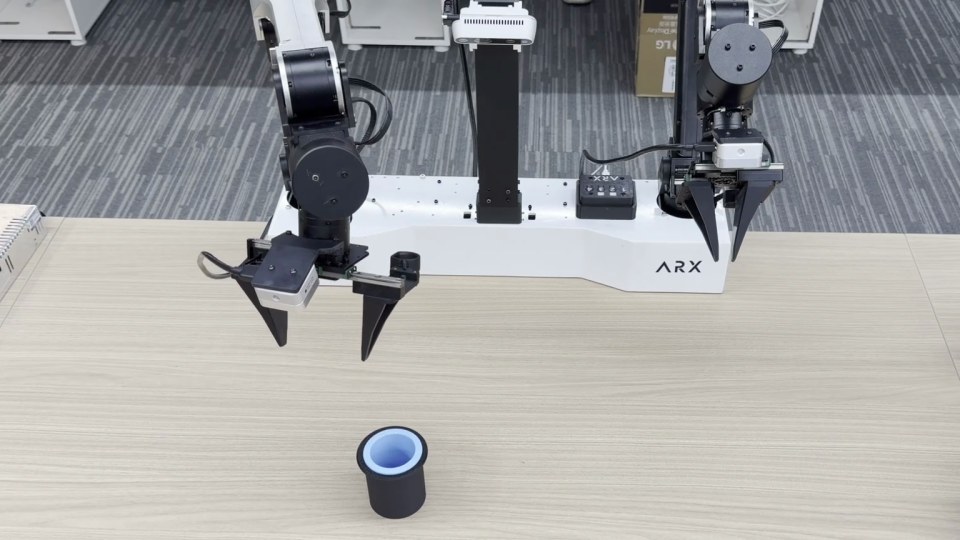} & & \rlab{Build Tower} & \includegraphics[width=\rw]{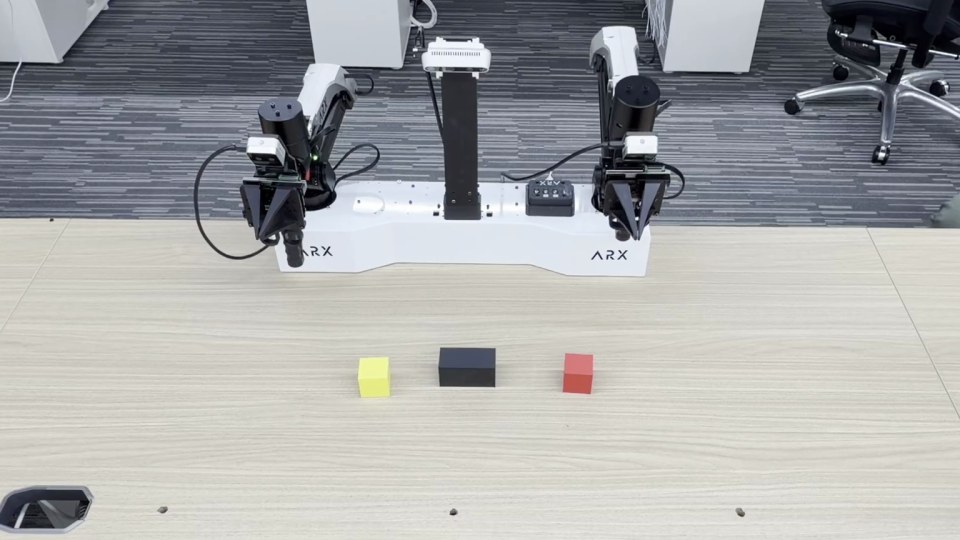} & \includegraphics[width=\rw]{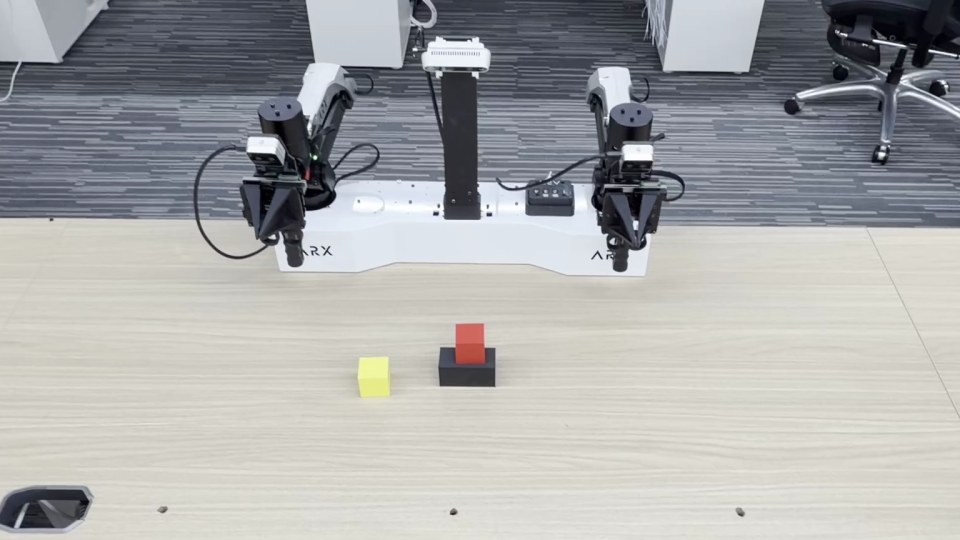} & \includegraphics[width=\rw]{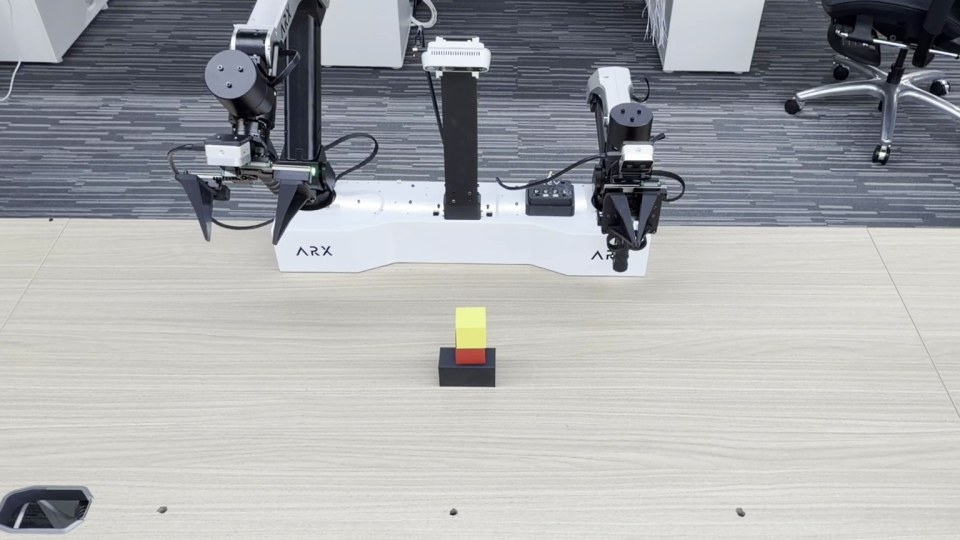} \\[1pt]
\rlab{Mug-Box} & \includegraphics[width=\rw]{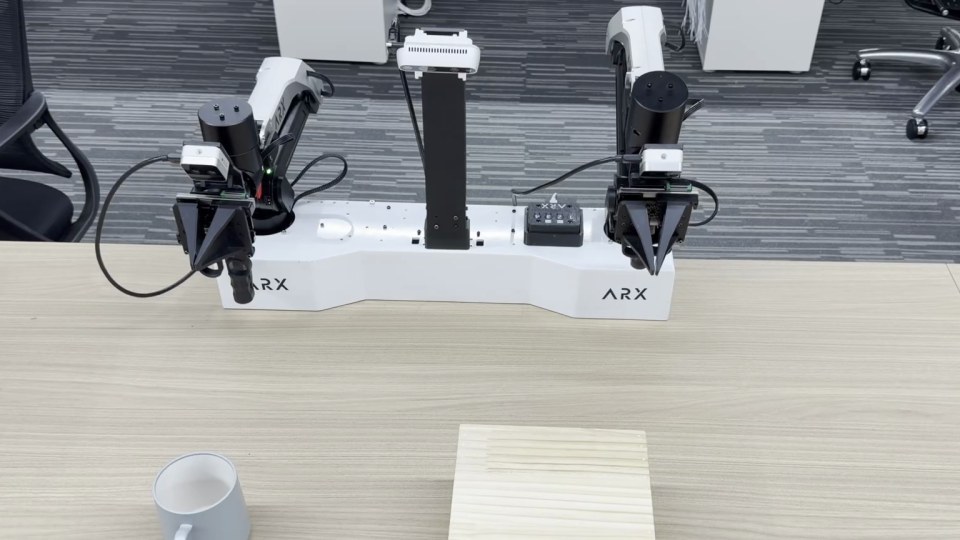} & \includegraphics[width=\rw]{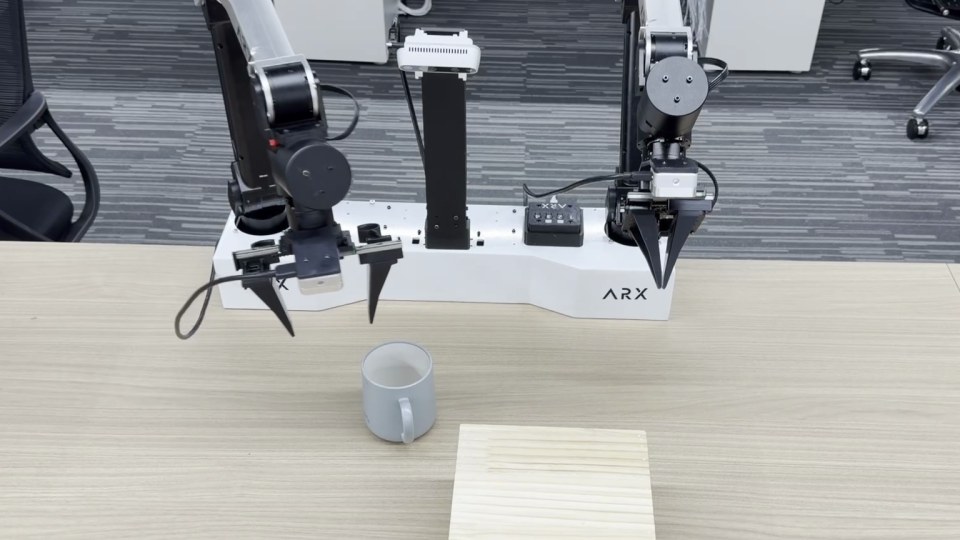} & \includegraphics[width=\rw]{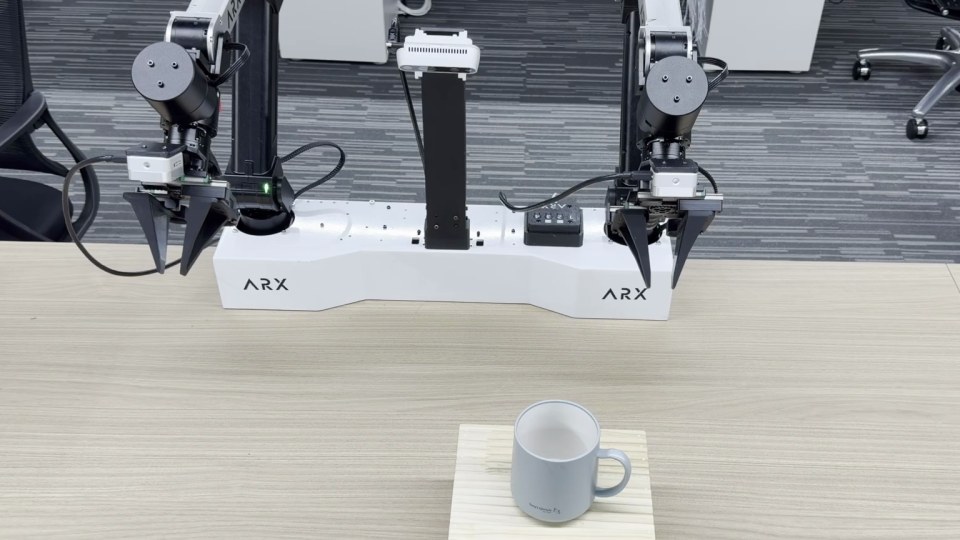} & & \rlab{Open Drawer} & \includegraphics[width=\rw]{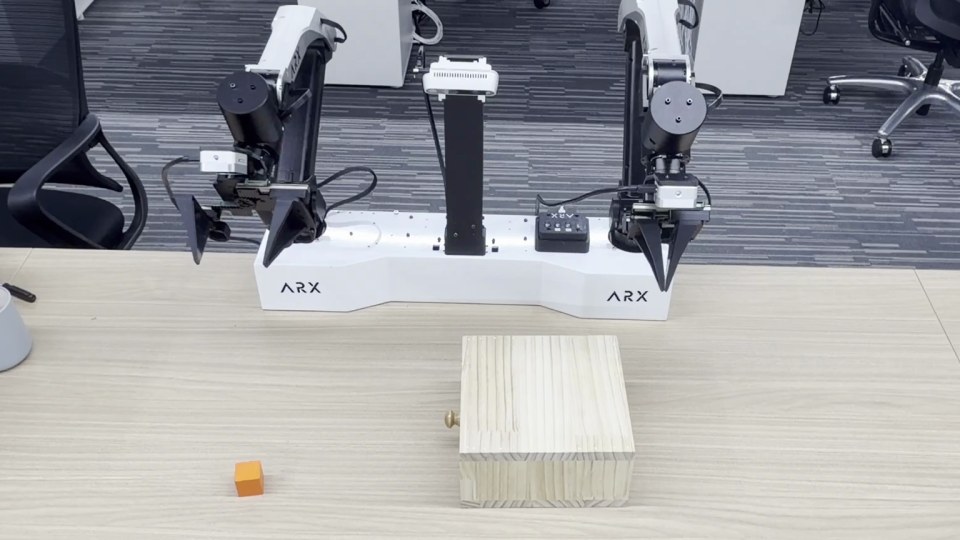} & \includegraphics[width=\rw]{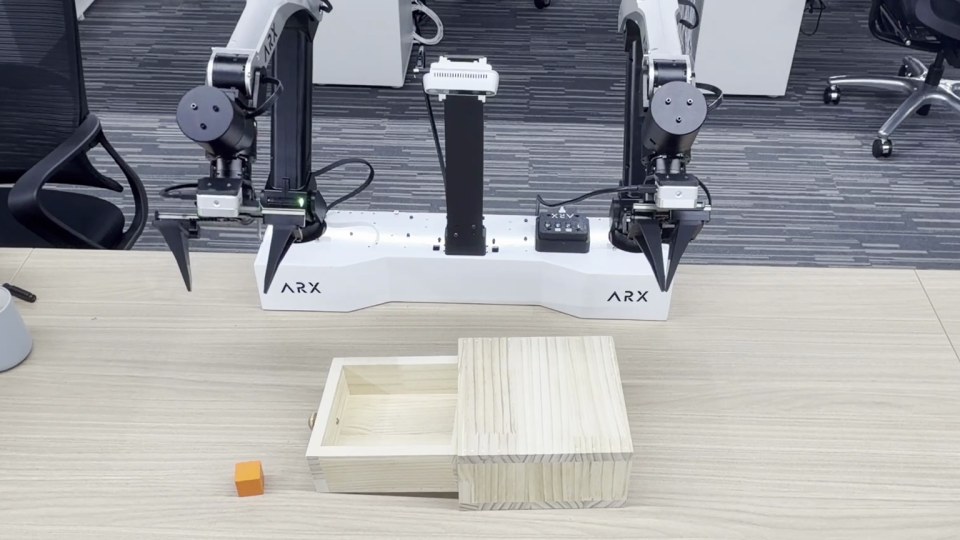} & \includegraphics[width=\rw]{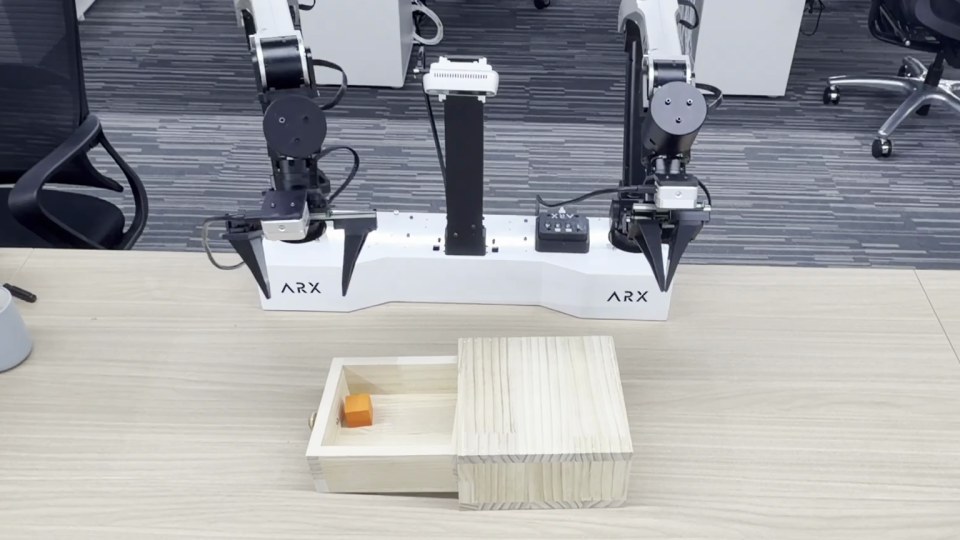} \\[1pt]
\rlab{Sweep} & \includegraphics[width=\rw]{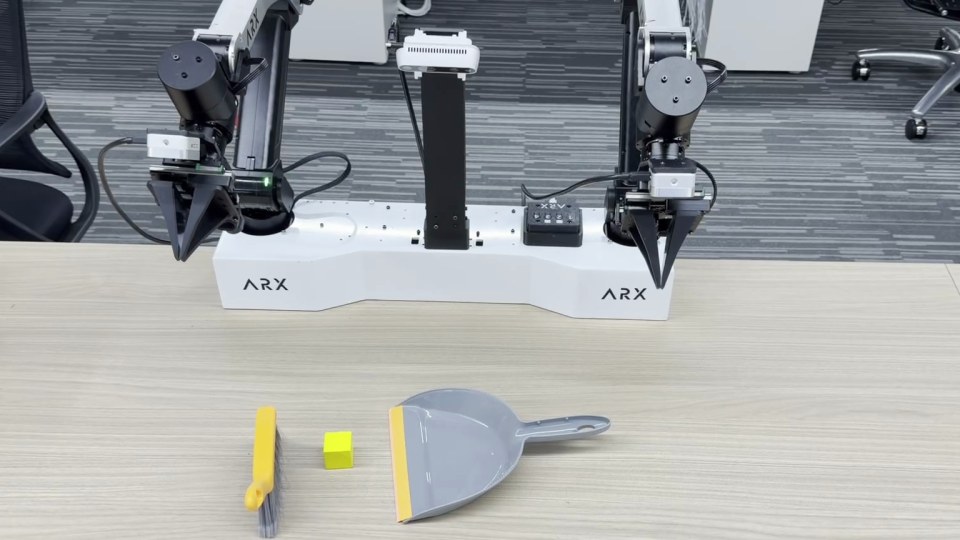} & \includegraphics[width=\rw]{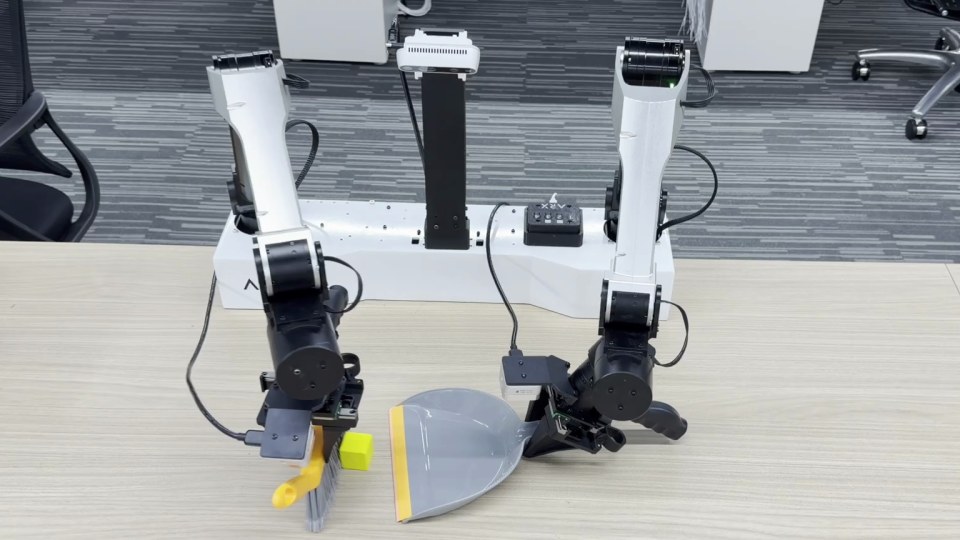} & \includegraphics[width=\rw]{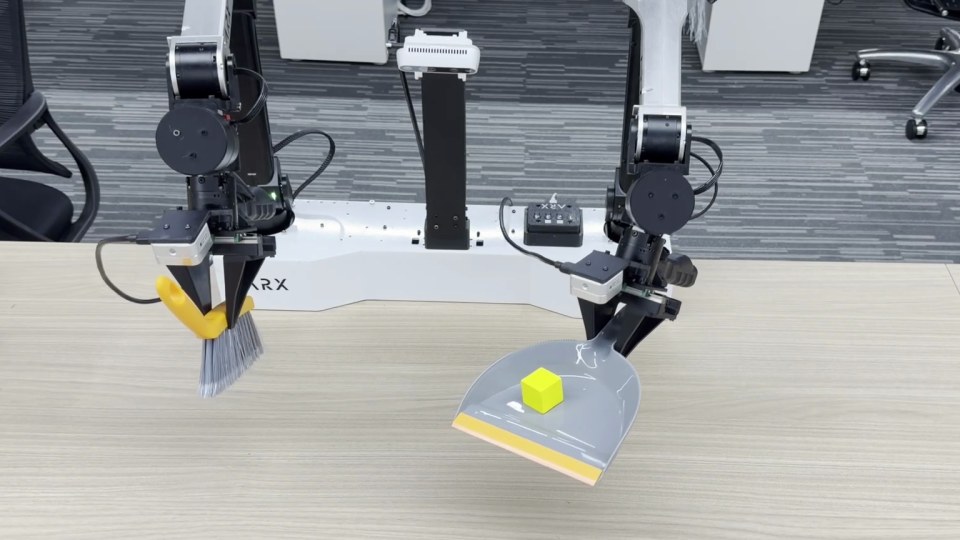} & & \rlab{Lift-Box} & \includegraphics[width=\rw]{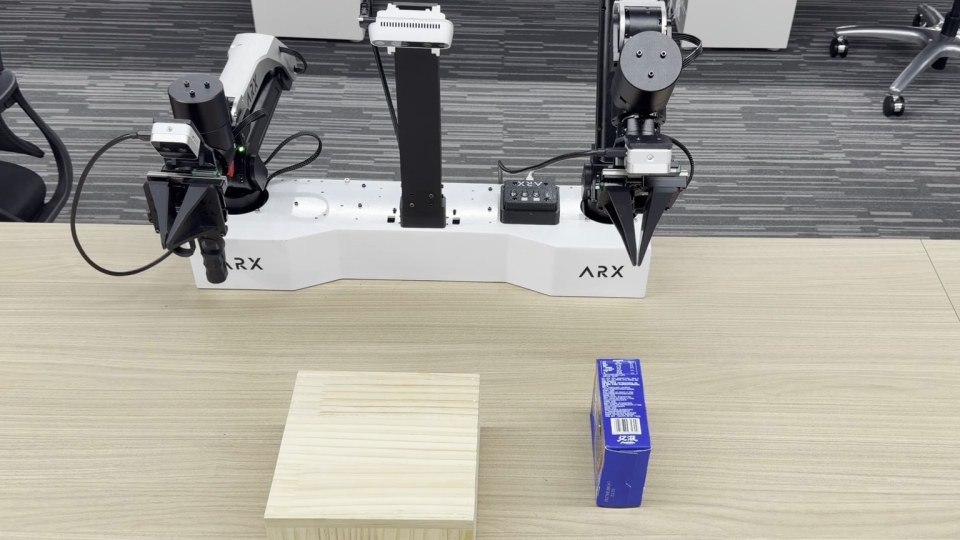} & \includegraphics[width=\rw]{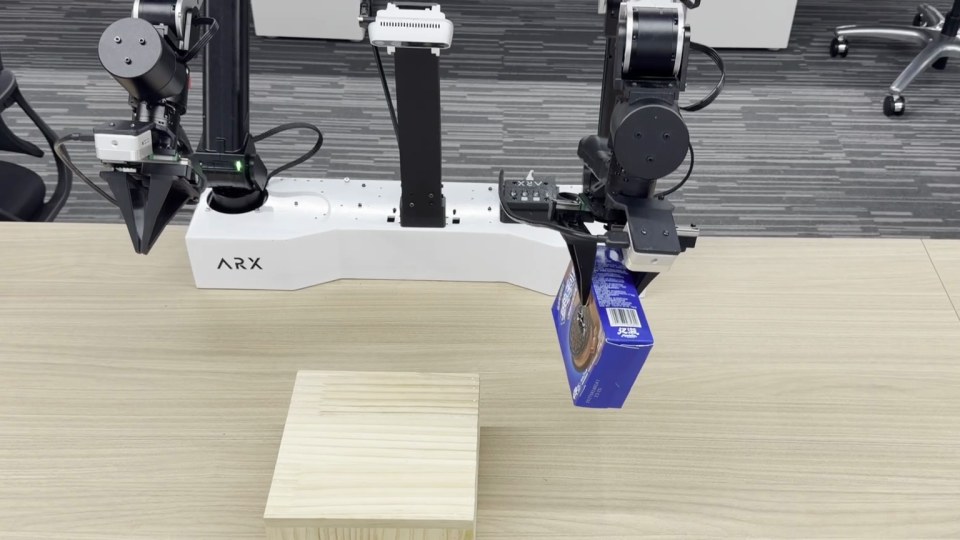} & \includegraphics[width=\rw]{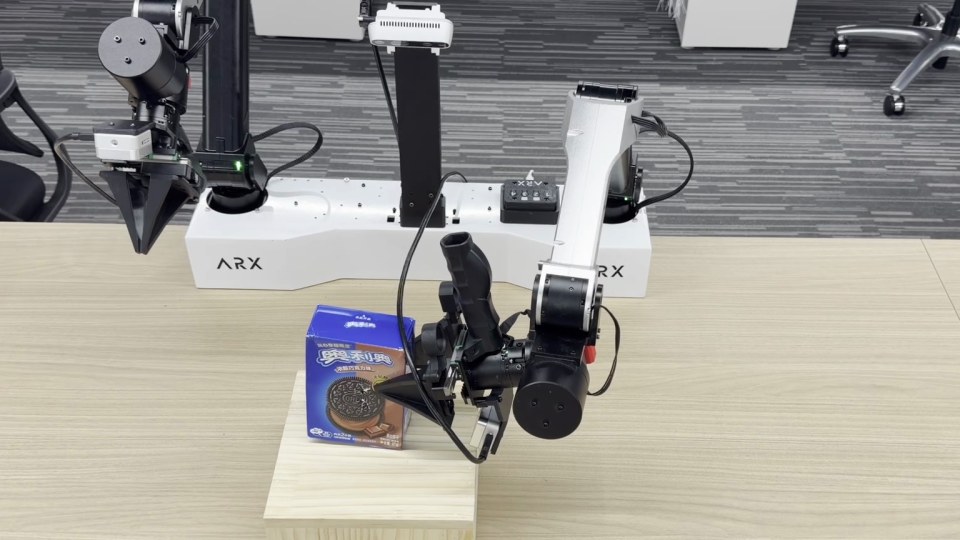} \\[1pt]
\end{tabular}
\caption{\textbf{Real-robot execution on ARX LIFT2 (no demonstration).} Start, key moment, and final state of one trial per task.}
\label{apx:fig:real-arx}
\vspace{14pt}
\setcounter{figure}{5}\renewcommand{\thefigure}{S6d}\renewcommand{\theHfigure}{S6d}
\rtab
\begin{tabular}{@{}c@{\hspace{2pt}}ccc@{\hspace{6pt}}c@{\hspace{0pt}}c@{\hspace{2pt}}ccc@{}}
\rhead
\rlab{Apple} & \includegraphics[width=\rw]{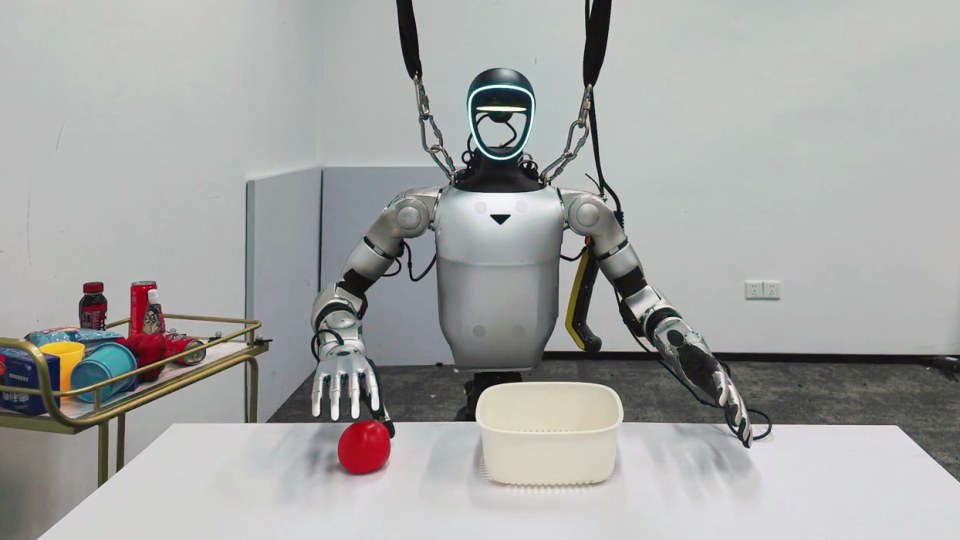} & \includegraphics[width=\rw]{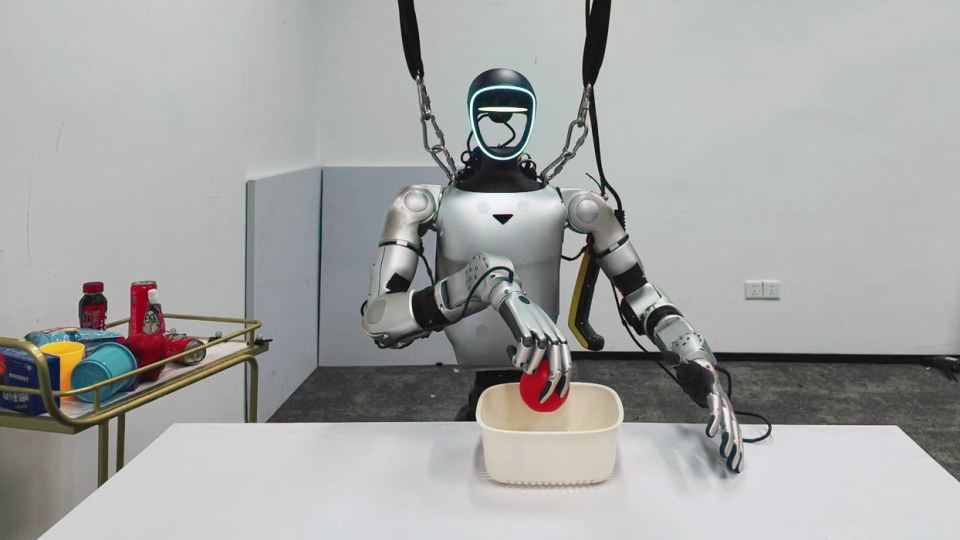} & \includegraphics[width=\rw]{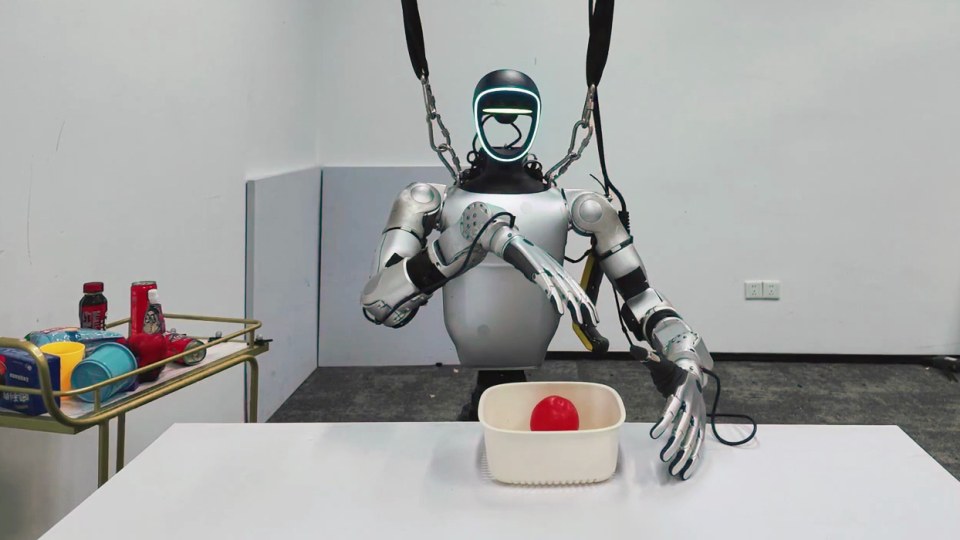} & & \rlab{Cubes} & \includegraphics[width=\rw]{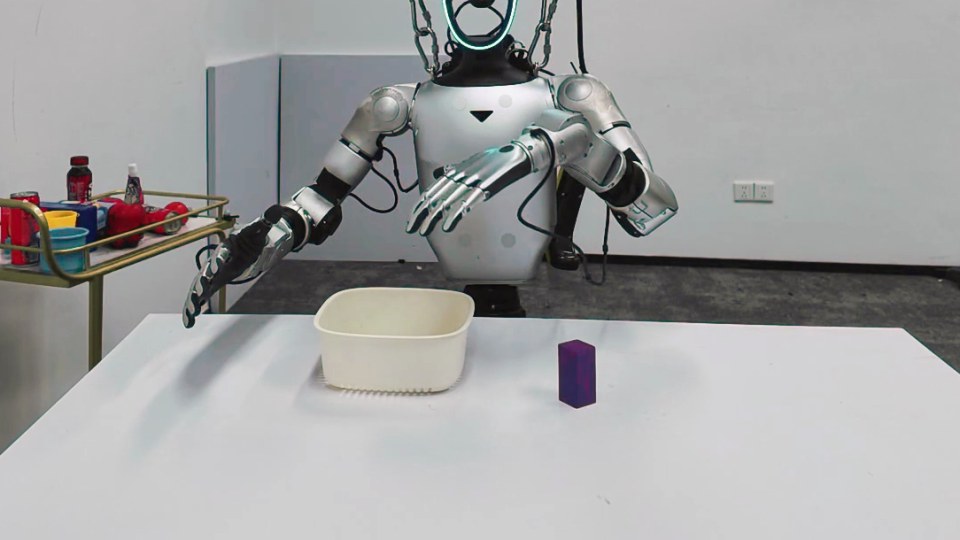} & \includegraphics[width=\rw]{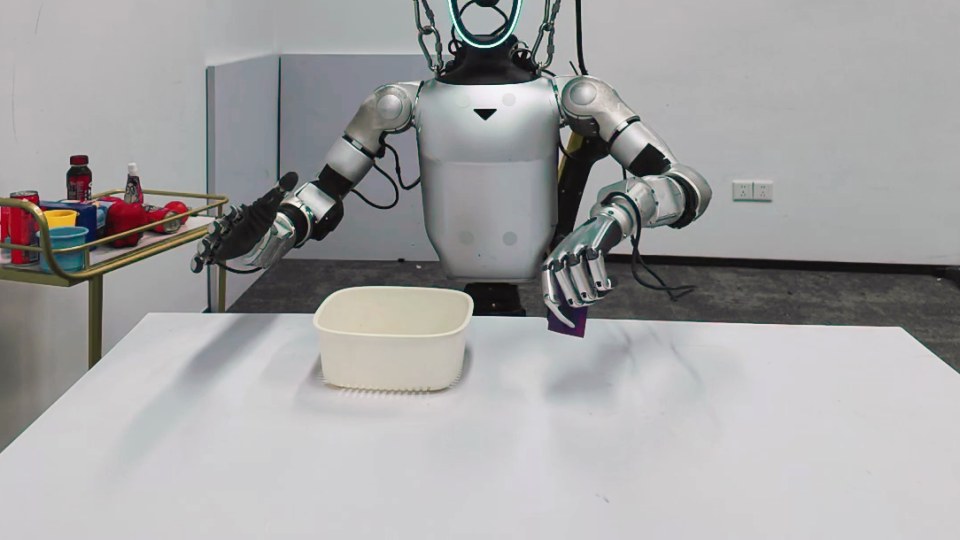} & \includegraphics[width=\rw]{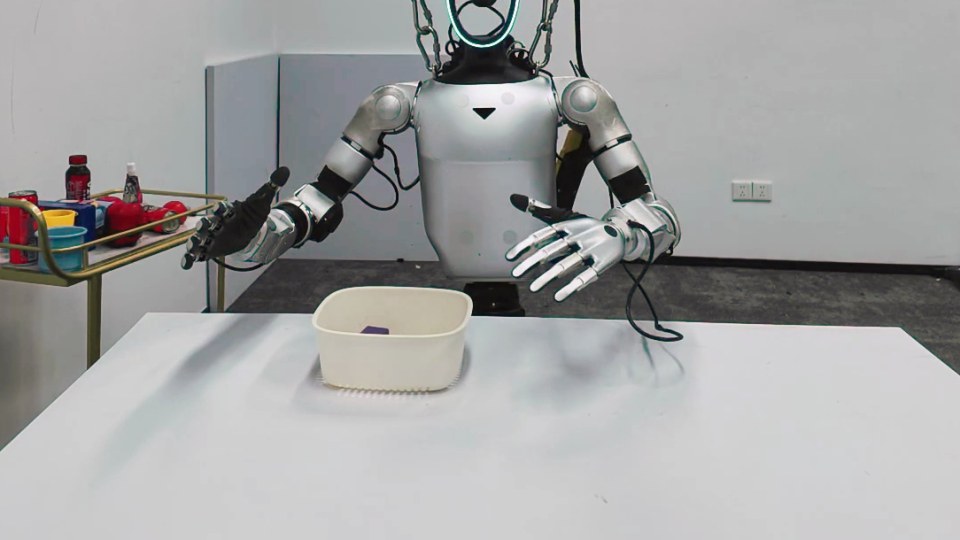} \\[1pt]
\rlab{Stack Cups} & \includegraphics[width=\rw]{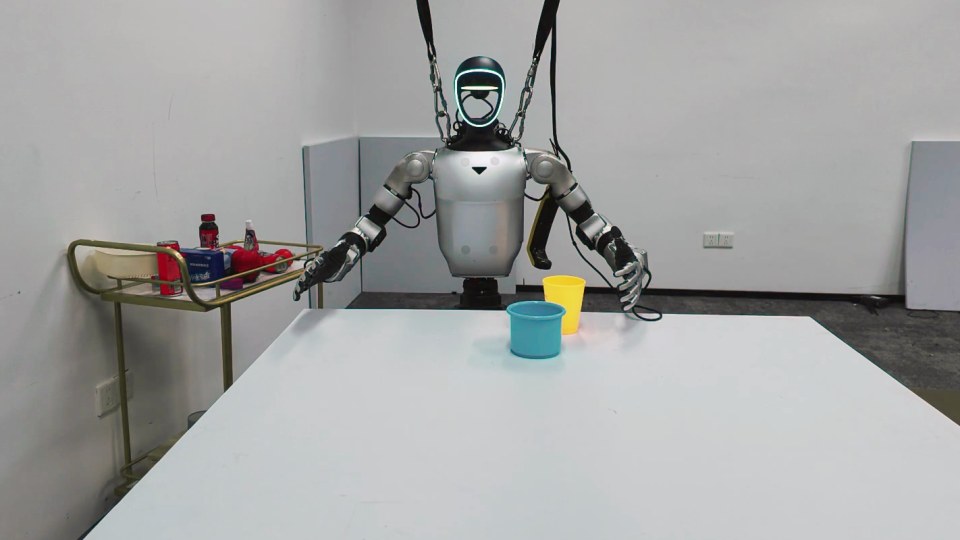} & \includegraphics[width=\rw]{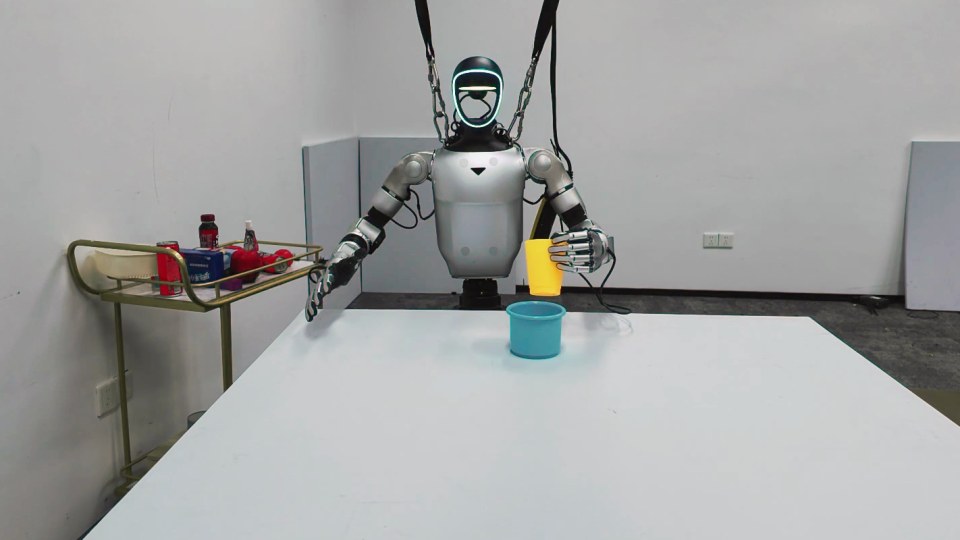} & \includegraphics[width=\rw]{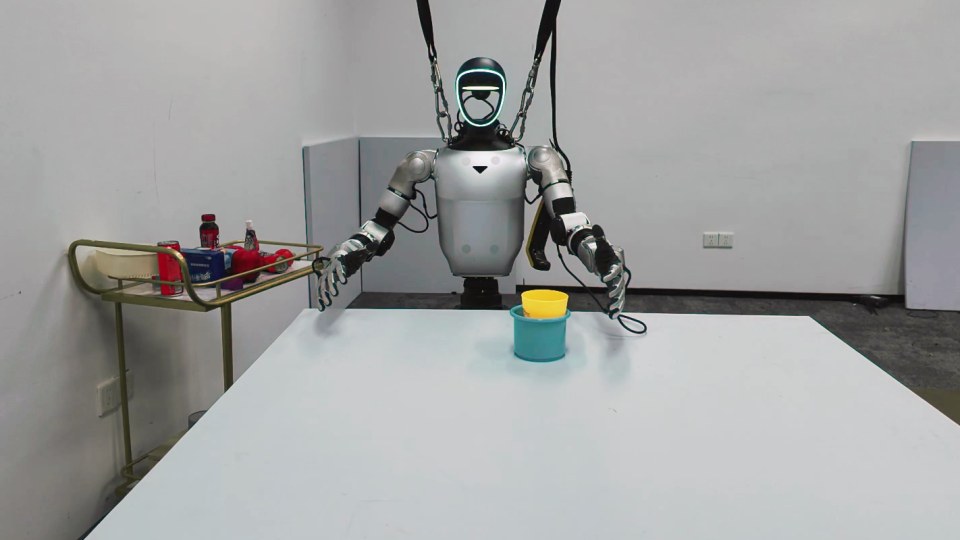} & & \rlab{Mug-Box} & \includegraphics[width=\rw]{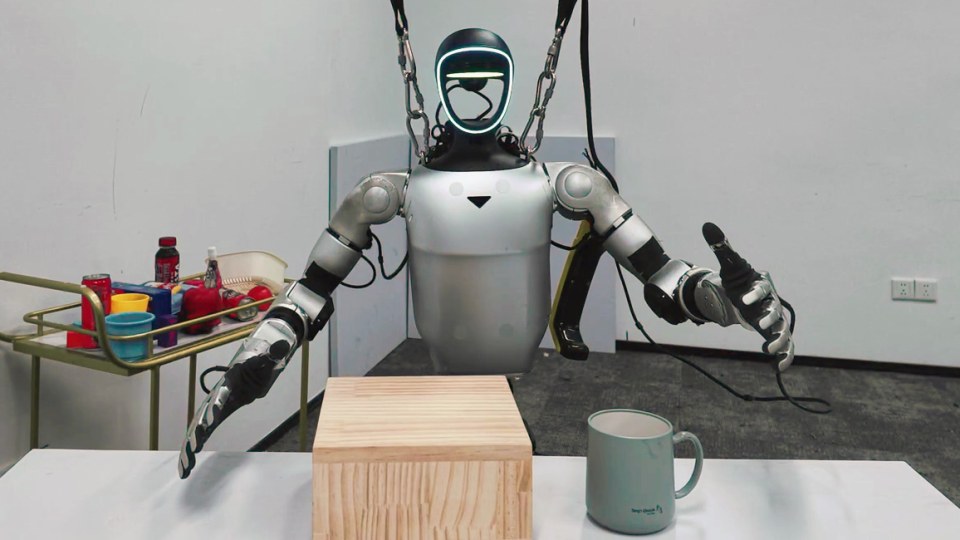} & \includegraphics[width=\rw]{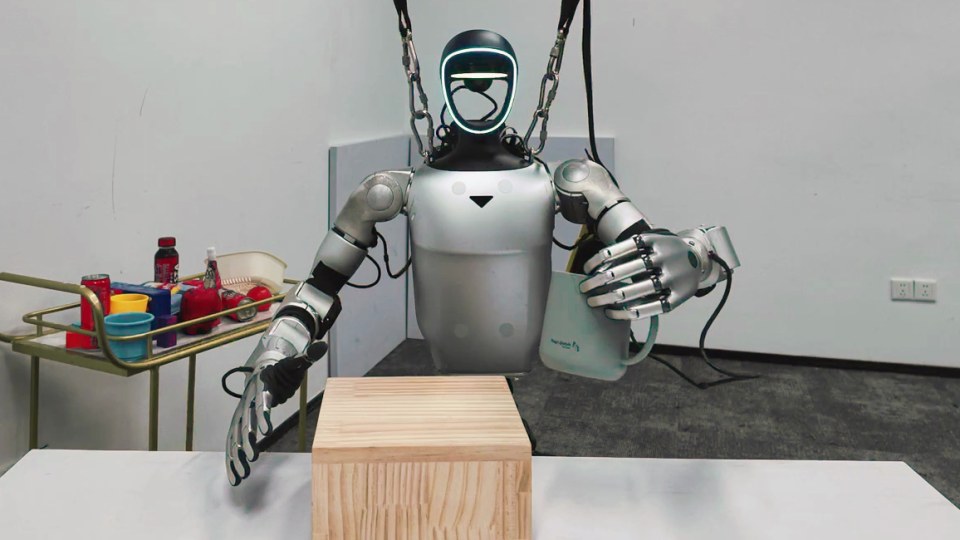} & \includegraphics[width=\rw]{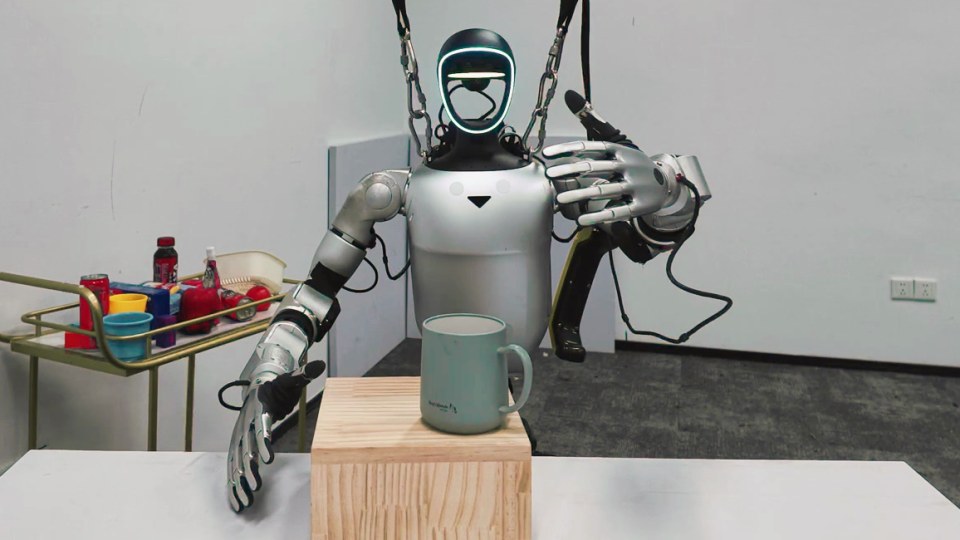} \\[1pt]
\end{tabular}
\caption{\textbf{Real-robot execution on Unitree G1 (no demonstration).} The four tasks it ran on the real robot; start, key moment, and final state of one trial per task.}
\label{apx:fig:real-g1}
\end{figure*}

\begin{figure*}[t]
\centering
\setlength{\tabcolsep}{1pt}
\newcommand{\lw}{0.157\textwidth}

\begin{tabular}{@{}c@{\hspace{3pt}}cc@{\hspace{5pt}}c@{\hspace{0pt}}cc@{\hspace{5pt}}c@{\hspace{0pt}}cc@{}}
 & \multicolumn{2}{c}{\footnotesize Layout 1} & & \multicolumn{2}{c}{\footnotesize Layout 2} & & \multicolumn{2}{c}{\footnotesize Layout 3} \\
 & \scriptsize Start & \scriptsize Final state & & \scriptsize Start & \scriptsize Final state & & \scriptsize Start & \scriptsize Final state \\[1pt]
\rl{0.50in}{Stack Cups} & \includegraphics[width=\lw]{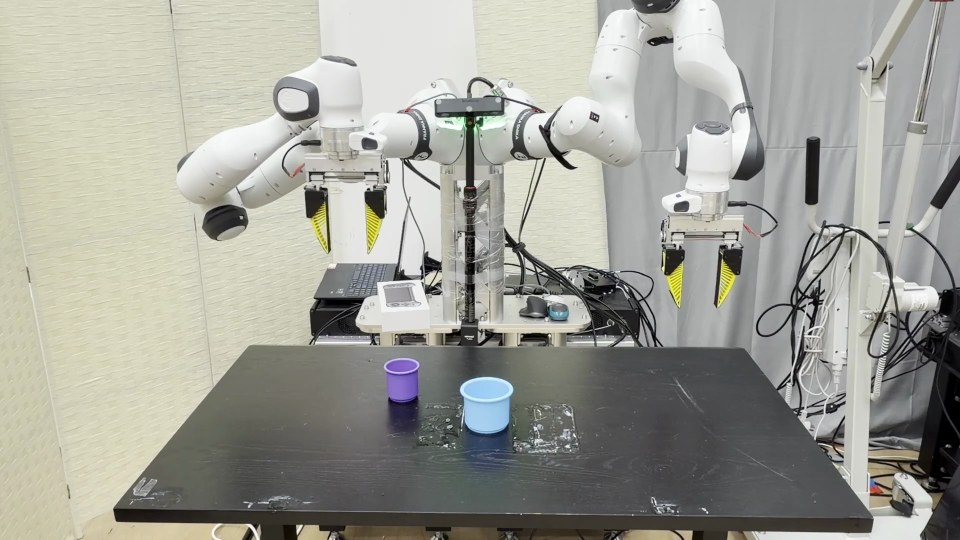} & \includegraphics[width=\lw]{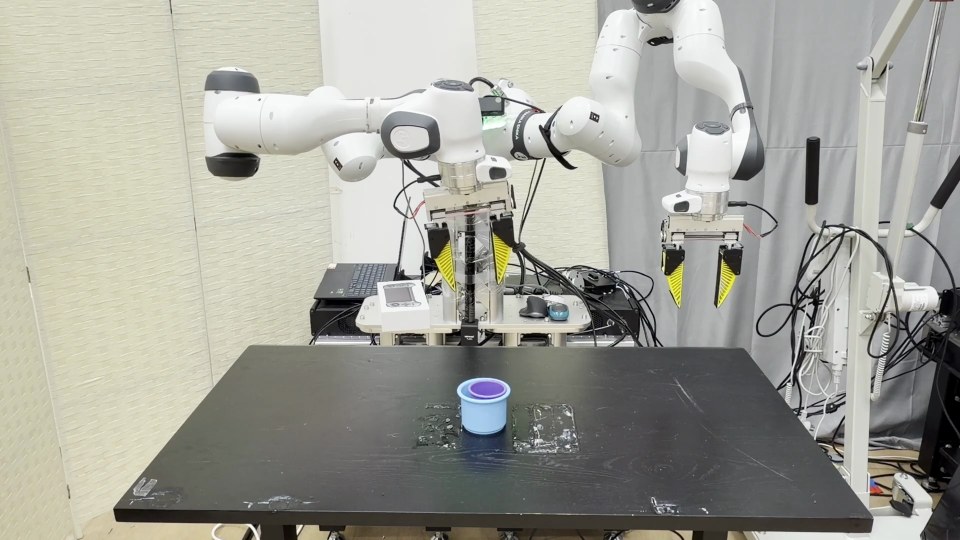} & & \includegraphics[width=\lw]{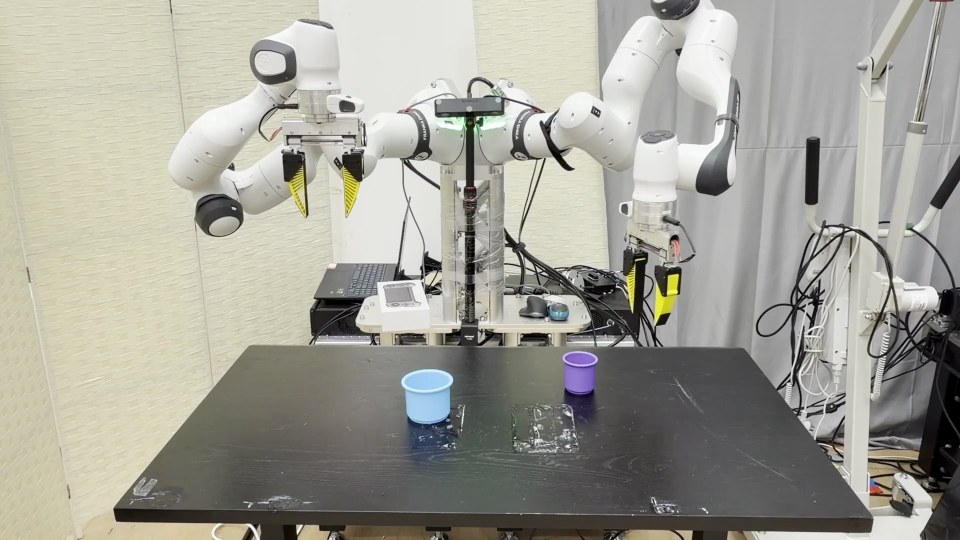} & \includegraphics[width=\lw]{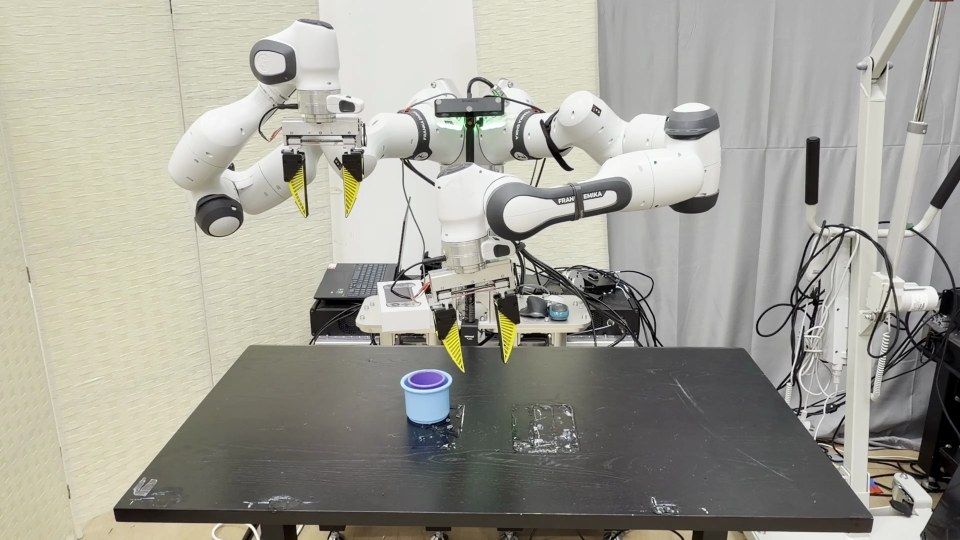} & & \includegraphics[width=\lw]{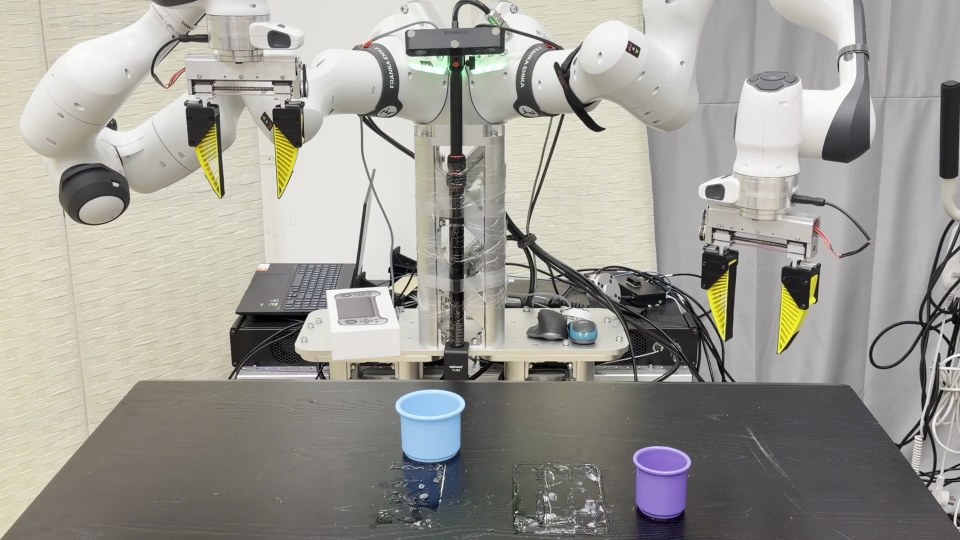} & \includegraphics[width=\lw]{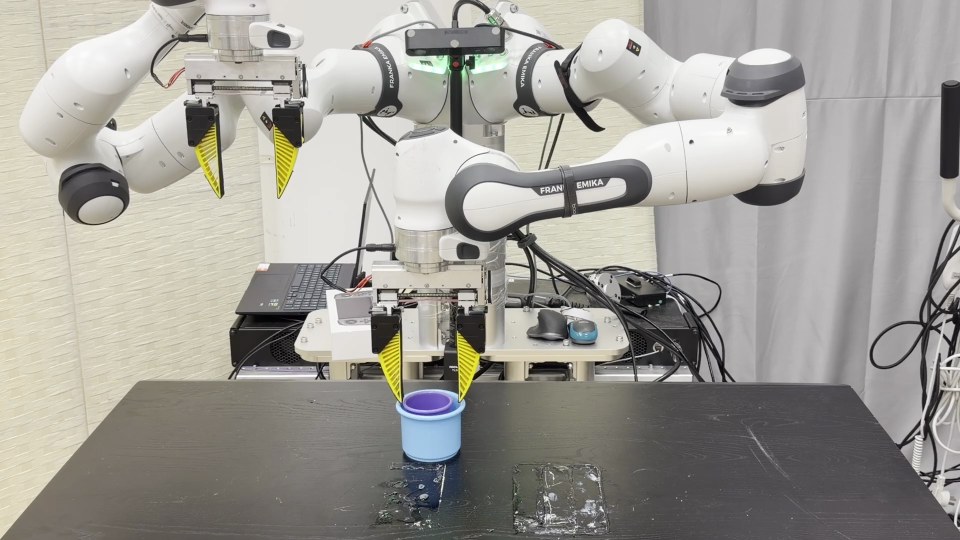} \\[2pt]
\rl{0.50in}{Build Tower} & \includegraphics[width=\lw]{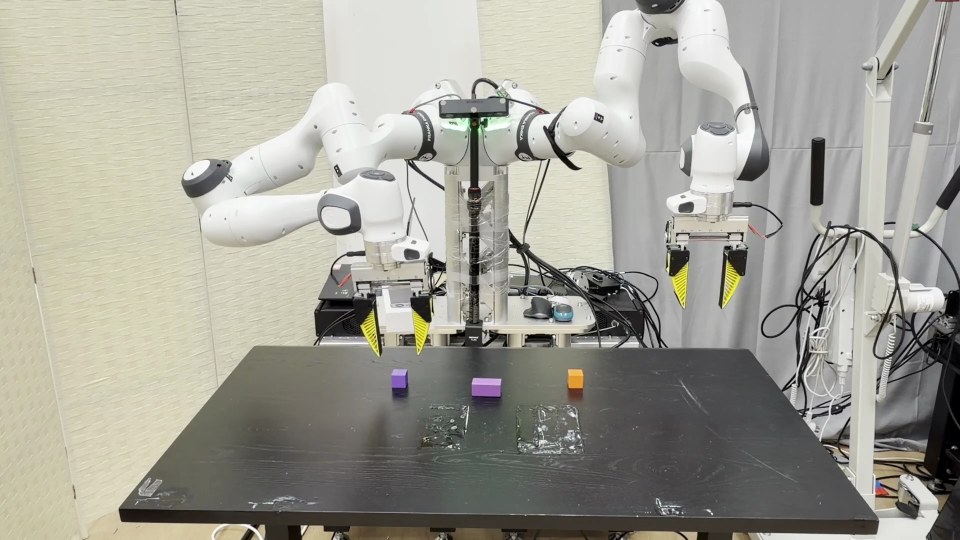} & \includegraphics[width=\lw]{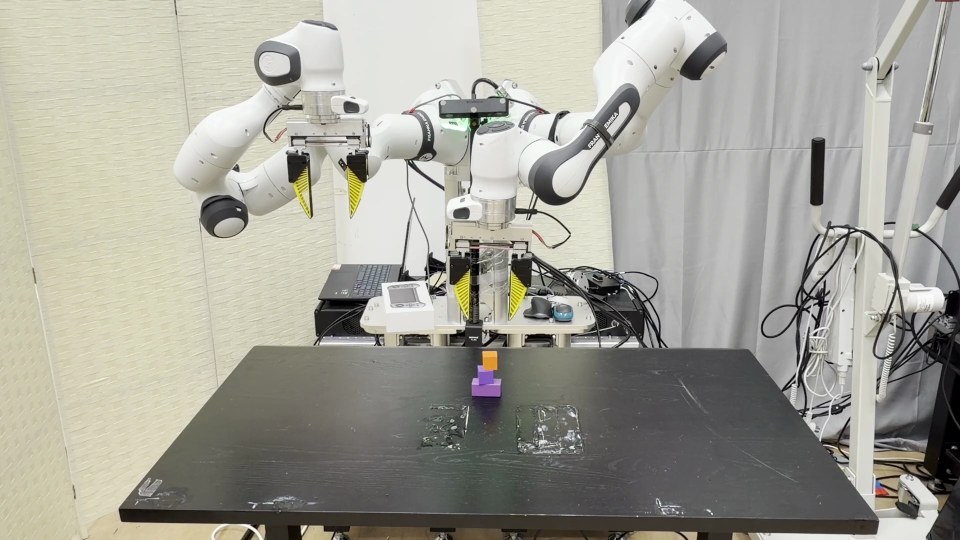} & & \includegraphics[width=\lw]{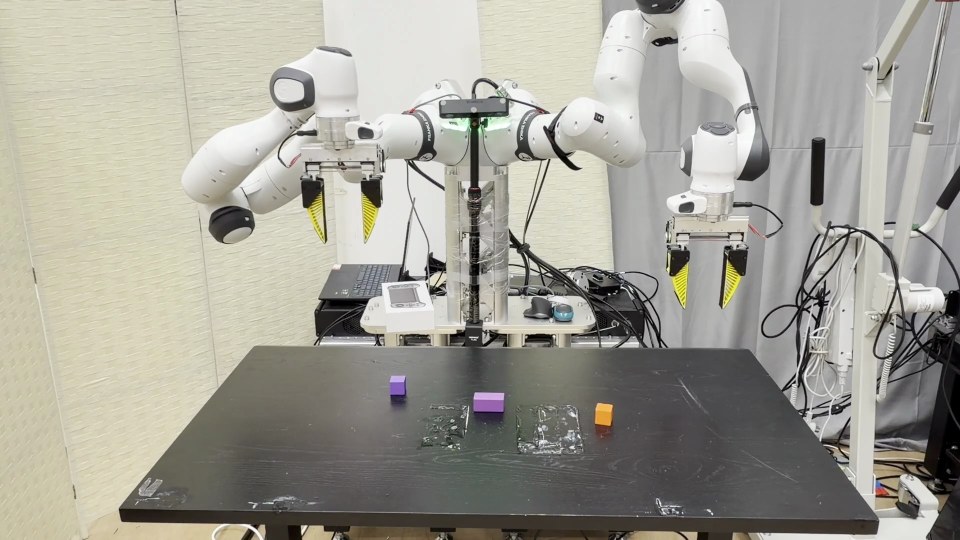} & \includegraphics[width=\lw]{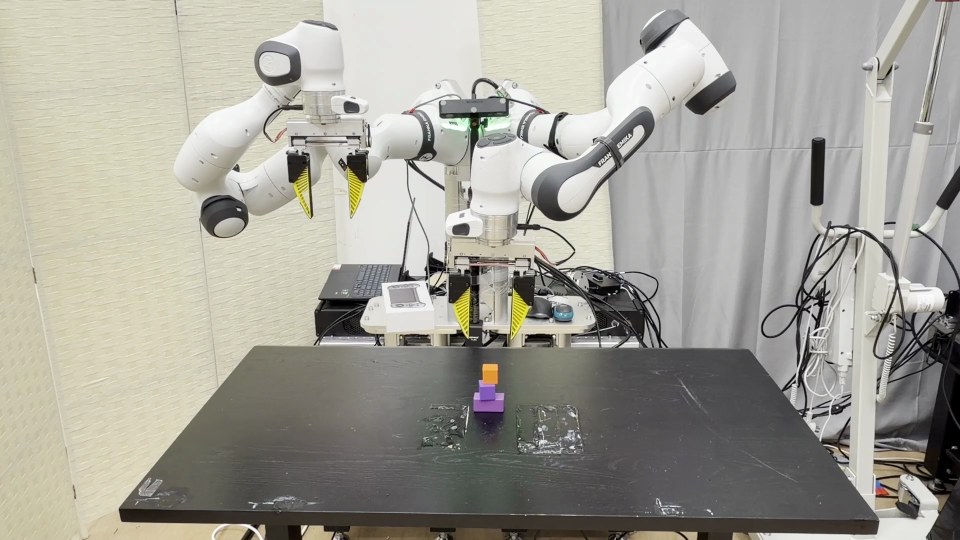} & & \includegraphics[width=\lw]{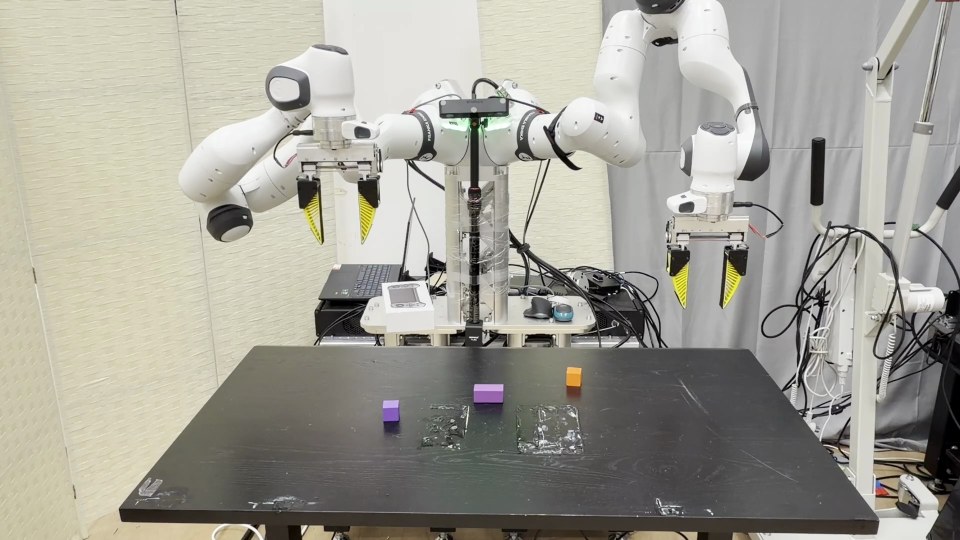} & \includegraphics[width=\lw]{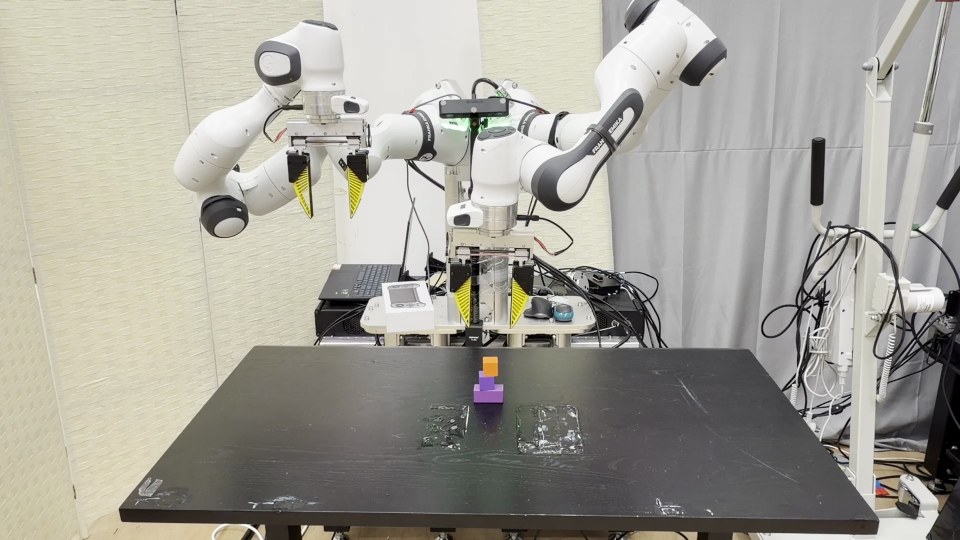} \\[2pt]
\rl{0.50in}{Lift-Box} & \includegraphics[width=\lw]{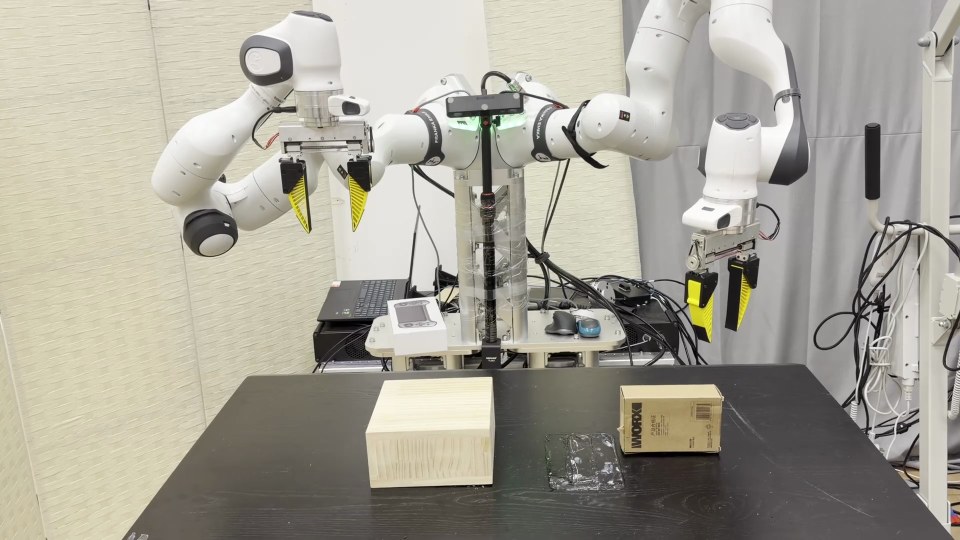} & \includegraphics[width=\lw]{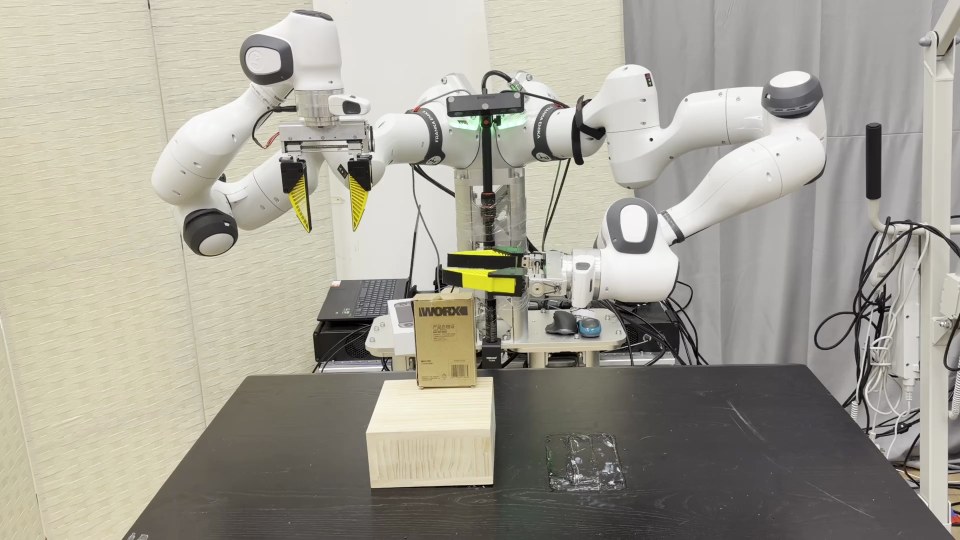} & & \includegraphics[width=\lw]{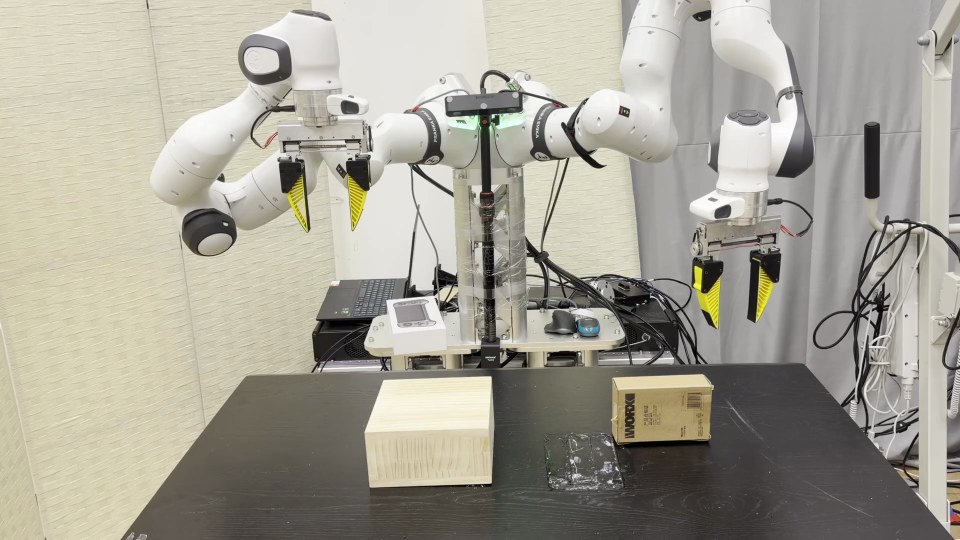} & \includegraphics[width=\lw]{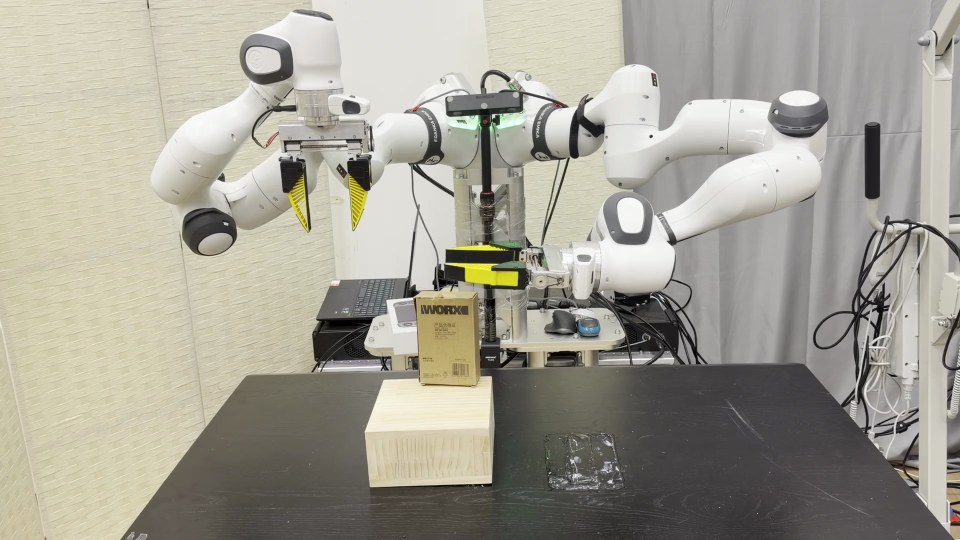} & & \includegraphics[width=\lw]{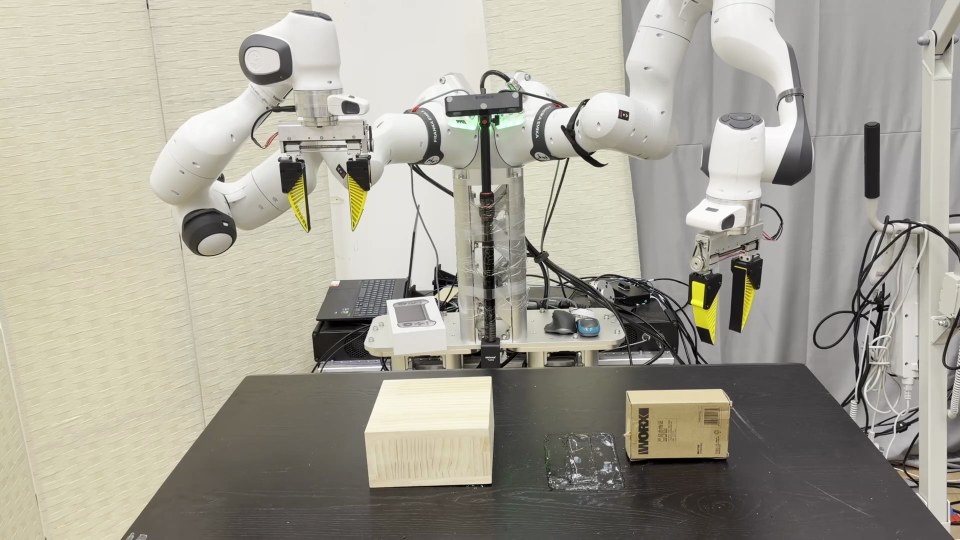} & \includegraphics[width=\lw]{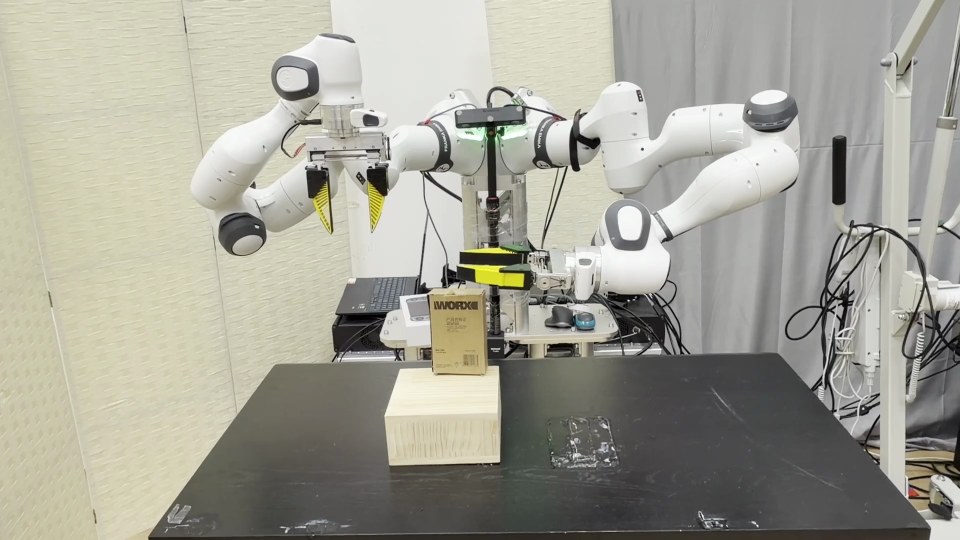} \\[2pt]
\end{tabular}
\caption{\textbf{Same program, different object layouts.} Dual-arm Franka on the real robot runs one program per task from three different object layouts, placed by hand; start and final frame of each trial.}
\label{apx:fig:layouts}
\end{figure*}

\section{Protocol and baselines}
\label{apx:sec:protocol}

\paragraph{Real-robot protocol.}
The final programs are frozen after practice in simulation and run on the real robots as ordinary code, with no code edits, no human intervention, and no LLM call at run time (Sec.~3.6). Each task is run for 10 trials, with the objects placed by hand at varied positions. xTrainer, Dual-arm Franka, and ARX LIFT2 run all 8 tasks; Unitree G1 runs 4 of them (Apple, Cubes, Stack Cups, and Mug-Box). Dual-arm Franka, ARX LIFT2, and Unitree G1 received no demonstration. Results per task and robot are in Tab.~5; examples are shown in Fig.~3(b), in Figs.~\ref{apx:fig:real-xtrainer}--\ref{apx:fig:real-g1} and~\ref{apx:fig:layouts}, and in the video.

\paragraph{Baselines.}
All baselines are evaluated in simulation on the source robot xTrainer, on the same 50 unseen test layouts per task (TEST 0--49) as RESET (Tab.~1).
\begin{itemize}[nosep,leftmargin=1.5em]
\item Open-loop Demo Replay replays the demonstration.
\item Zero-shot Code takes 3 independent samples per task, with the same interfaces and contact priors as RESET; Tab.~1 reports the mean and the best of the three.
\item $\pi_{0.5}$ is fine-tuned for 6,000 steps on 100 RESET trajectories from disjoint layouts, one model per task.
\item Diffusion Policy is trained from scratch on the same trajectories, one model per task.
\end{itemize}
Both policies use their default settings.

\section{Supplementary video}
\label{apx:sec:video}

The video on the project page presents, in order, the problem (why self-improving robot code needs a world to practice in), the idea (what one real demonstration provides), the method (the five steps of RESET: Rebuild, Examine, Self-improve, Execute, and Transfer), and the simulation results. It then shows simulation rollouts of all four robots on all eight tasks and real-robot rollouts of the final programs on the four robots. Rollouts are sped up.

\end{document}